# Wasserstein Distributionally Robust Optimization and Variation Regularization*

Rui Gao
Department of Information, Risk and Operations Management, University of Texas at Austin, rui.gao@mccombs.utexas.edu

Xi Chen
Stern School of Business, New York University, xchen3@stern.nyu.edu

Anton J. Kleywegt
H. Milton Stewart School of Industrial and Systems Engineering, Georgia Institute of Technology, anton@isye.gatech.edu

Wasserstein distributionally robust optimization (DRO) has recently achieved empirical success for various applications in operations research and machine learning, owing partly to its regularization effect. Although connection between Wasserstein DRO and regularization has been established in several settings, existing results often require restrictive assumptions, such as smoothness or convexity, that are not satisfied for many problems. In this paper, we develop a general theory on the *variation regularization* effect of the Wasserstein DRO – a new form of regularization that generalizes total-variation regularization, Lipschitz regularization and gradient regularization. Our results cover possibly non-convex and non-smooth losses and losses on non-Euclidean spaces. Examples include multi-item newsvendor, portfolio selection, linear prediction, neural networks, manifold learning, and intensity estimation for Poisson processes, etc. As an application of our theory of variation regularization, we derive new generalization guarantees for adversarial robust learning.

## 1. Introduction

Wasserstein distributionally robust optimization (DRO) [59, 24, 65, 11, 26] is an emerging framework for learning and decision-making under uncertainty in which the learner/decision-maker has limited knowledge on the data-generating mechanism. It studies a minimax robust optimization problem

$$\min_{f\in\mathcal{F}} \sup_{\mathbb{P}:W_p(\mathbb{P},\mathbb{P}_n)\le\rho} \mathbb{E}_{z\sim\mathbb{P}}[f(z)], \tag{P}$$

where $f\in\mathcal{F}$ represents the loss function dependent on the random data $z$; the inner supremum finds the worst-case expected loss among a ball of distributions with radius $\rho$, containing all distributions that are close to the empirical distribution $\mathbb{P}_n$ of sample size $n$ in $p$-Wasserstein distance $W_p$. In recent years, Wasserstein DRO has been successfully applied to numerous problems in machine learning, including (semi)-supervised learning [10, 18], adversarial learning [51, 53, 41, 36], reinforcement learning [1, 52, 20], transfer learning [55, 21, 35], etc. We refer to Kuhn et al. [32] for a recent survey.

The regularization effect of Wasserstein DRO contributes largely to its empirical success. In the literature, the connection between Wasserstein DRO and regularization has been established in various settings; see [24, 46, 12, 47, 18] for equivalence results when $p=1$, and see [12, 27, 55, 13, 5] for asymptotic equivalence results when $p\in(1,\infty)$. Unfortunately, however, all above-mentioned results have certain limitations making them hard to be applicable to important classes of problems in operations research and machine learning. More specifically, equivalence results on 1-Wasserstein DRO [24, 47] require unbounded support of distributions and convexity of loss functions, although many real-world problems have bounded support due to either the nature of the problem or data

* This manuscript is previously titled *Distributionally robust optimization and regularization in statistical learning*.

normalization [38], and most loss functions in deep learning are non-convex. Asymptotic equivalence results [12, 27, 55, 5] on $p$-Wasserstein DRO require smoothness on the loss functions, despite the fact that many common loss functions are merely piecewise smooth, including newsvendor cost, least absolute loss, and the ReLU neural network and its variants. From these aspects, the current theory on the regularization effect of Wasserstein DRO is not yet complete.

In this paper, we aim to close this gap by providing a general connection between Wasserstein DRO and regularization. To this end, we develop a new concept, termed as the **variation of loss** (see Definitions 1 and 2), denoted as $\mathcal{V}(f)$, that measures the magnitude of change on the expected loss when the data distribution is perturbed. It generalizes total variation for real-valued functions, and is reduced to the Lipschitz norm for Lipschitz continuous functions and to weighted empirical gradient norm for differentiable functions. Intuitively, when the variation of loss is controlled, small perturbations of random data would have little impact on the expected loss and thus would not deteriorate the solution quality much. We develop results showing that Wasserstein DRO (P) is closely related to a **variation regularization** problem:

$$\min_{f\in\mathcal{F}} \mathbb{E}_{z\sim\mathbb{P}_n}[f(z)]+\rho\mathcal{V}(f). \tag{V}$$

Our results illustrate the variation regularization effect intrinsically associated with the Wasserstein DRO in a general scope. More specifically,

(I) For $p$-Wasserstein DRO, $p\in(1,\infty]$, we show that for a broad class of loss functions, possibly non-convex and non-smooth, with high probability, the Wasserstein DRO (P) is asymptotically equivalent to variation regularization (V) up to a higher order $O(\rho^{2\wedge p})$ remainder (Theorems 1 and 4).
For $p<2$, the bound $\rho^p$ is tight (Example 1), indicating a qualitative disparity among different Wasserstein orders.
For $p=2$, one of the most popular choices of Wasserstein order, we demonstrate our results in the context of multi-item newsvendor (Example 5), portfolio selection with piecewise linear utility functions (Example 6), and gradient regularization for leaky ReLU networks (Example 9). Moreover, our results hold for general non-Euclidean metric spaces, illustrated by Laplacian regularization for manifold learning (Example 10), and score function regularization for intensity estimation of point processes (Example 11).

(II) For 1-Wasserstein DRO, the asymptotic equivalence between the Wasserstein DRO (P) and the variation regularization (V) may not hold in general (Example 3). Nevertheless, we prove a sandwich theorem (Theorem 2 and Corollary 1), in the sense that with high probability, (P) with radius $\rho$ is upper bounded by (V) with tunning parameter $\rho$ and lower bounded by (V) with tunning parameter $\eta\rho$, where $\eta\in(0,1]$. Thereby minimizing the Wasserstein robust loss also controls the variation of the loss function. As examples, we consider linear prediction with Lipschitz losses (Examples 7 and 8), and extend the existing equivalence results [24, 47] to a more general class of non-convex functions (Corollary 2).

(III) We apply our results on $\infty$-Wasserstein DRO to adversarial robust learning and establish its equivalence to empirical total-variation regularization (Example 12). Notably, we develop new generalization guarantees that quantify the gap between the empirical adversarial risk and population adversarial risk (Theorem 3). We show that in the adversarial setting, the generalization behavior of a machine learning model is affected not only by the complexity of the loss function class as in the classical empirical risk minimization, but also by the complexity of the slope of the loss function class.

### 1.1. Related work

In the literature, there is a general belief that robust optimization is related to regularization, dating back to the pioneer work of Xu et al. [60, 61]. They establish an equivalence between data-driven robust optimization and norm regularization for LASSO and support vector machine, which is then generalized to linear and matrix regression in [9] among others. Given the close relationship between Wasserstein DRO and data-driven robust optimization [26], it is expected that Wasserstein DRO would also exhibit certain regularization effect. Indeed, for 1-Wasserstein DRO, its equivalence to norm regularization has been established for piecewise-linear convex losses [24], logistic regression [46, 12], support vector machine [12], linear regression and classification and their kernelization [47, 18], etc. All these results require convexity and unboundedness of the data space. For $p$-Wasserstein DRO, $p \in (1, \infty]$, Blanchet et al. [12] establishes its connection to norm regularization under certain settings and studies the optimal selection of Wasserstein radius; the previous version of this work establishes an asymptotic equivalence between $p$-Wasserstein DRO and gradient regularization for smooth loss functions and is generalized by [5] under weaker assumptions; for 2-Wasserstein DRO, a finer analysis of the asymptotic equivalence is established in [13] and an asymptotic equivalence for its Lagrangian relaxation is developed in Volpi et al. [55]. All these results impose differentiability on the loss functions, with the exception of [5] that also considers weakly differentiable functions, but with a continuous nominal distribution that is not applicable to data-driven settings. For $p = \infty$, an equivalent form of Wasserstein DRO has been studied extensively in the context of adversarial robust learning (for example, [29, 39, 48]). The generalization bounds for adversarial robust learning have been studied recently in [63, 3, 4], and the generalization bounds for other finite $p$-Wasserstein DRO have been investigated in [51, 35, 47, 25].

In the DRO literature, besides Wasserstein DRO, other choices of distributional uncertainty sets have been explored [45, 64, 50, 15, 23, 43, 19, 28, 58, 8, 31, 6, 57]. In particular, for $\phi$-divergence DRO, its asymptotic equivalence to variance regularization has been established in [30, 33, 42], etc. Connection between regularization and various DRO formulations has been discussed in [62] under a new notion called worst-case sensitivity. We refer to Rahimian and Mehrotra [44] for a recent survey on distributionally robust optimization.

The rest of the paper proceeds as follows. In Section 2, we briefly review some results in Wasserstein DRO. We establish the connection between Wasserstein DRO and variation regularization in Section 3. To ease the exposition, we first state our results under the setting where data is supported on a subset of a Banach space on which the notion of differentiability is well-defined, and then we present the theory for general metric spaces in Section 3.4 and exemplify our results in Section 4. In Section 5, we study the adversarial robut learning and derive new generalization bounds. We conclude the paper in Section 6. Proofs of our results are for general metric spaces directly and are deferred to the Appendices.

## 2. Wasserstein Distributionally Robust Optimization

In this section, we introduce notations and provide some background on Wasserstein distributionally robust optimization.

Throughout this paper, we let $p \in [1, \infty]$ and denote by $q$ its Hölder conjugate, i.e., $\frac{1}{p} + \frac{1}{q} = 1$. Let $\mathcal{Z}$ be a metric space equipped with some metric $\mathrm{d}(\cdot, \cdot)$, measuring the similarity between samples. The diameter of a metric space $(\mathcal{Z}, \mathrm{d})$ is defined as $\operatorname{diam}(\mathcal{Z}) := \sup_{\tilde{z}, z \in \mathcal{Z}} \mathrm{d}(\tilde{z}, z)$. Denote by $\mathcal{P}(\mathcal{Z})$ the set of all Borel probability measures on $\mathcal{Z}$. The $\mathcal{L}^p(\mathbb{Q})$-norm of a $\mathbb{Q}$-measurable function $h$ is denoted by $\|h\|_{\mathbb{Q},p} := (\int_{\mathcal{Z}} h^p d\mathbb{Q})^{\frac{1}{p}}$ for $p \in [1, \infty)$, and $\|h\|_{\mathbb{Q},\infty} = \mathbb{Q}\text{-}\operatorname*{ess\,sup}_{z \in \mathcal{Z}} h(z)$. The Wasserstein distance of order $p$ between distributions $\mathbb{P}, \mathbb{Q} \in \mathcal{P}(\mathcal{Z})$ is defined as

$$W_p(\mathbb{P}, \mathbb{Q}) := \inf_{\pi \in \Gamma(\mathbb{P}, \mathbb{Q})} \|\mathrm{d}\|_{\pi,p},$$

where the minimization of $\pi$ is over the set $\Gamma(\mathbb{P}, \mathbb{Q})$ of all Borel probability distributions on $\mathcal{Z} \times \mathcal{Z}$ with marginal distributions $\mathbb{P}$ and $\mathbb{Q}$. For $p \in [1, \infty)$, we denote by $\mathcal{P}_p(\mathcal{Z}) := \{\mathbb{Q} \in \mathcal{P}(\mathcal{Z}) : \mathbb{E}_{z \sim \mathbb{Q}}[\mathrm{d}(z, z_0)^p] <$ $\infty\}$ the subset of $\mathcal{Z}$ with finite $p$-th moment, where $z_0 \in \mathcal{Z}$ is any fixed point in $\mathcal{Z}$. To ease notation, we adopt the convention $\mathcal{P}_\infty(\mathcal{Z}) := \mathcal{P}(\mathcal{Z})$.

Given a family $\mathcal{F}$ of loss functions, a nominal distribution $\mathbb{Q} \in \mathcal{P}_p(\mathcal{Z})$ and a radius $\rho \geq 0$, the corresponding Wasserstein DRO problem is

$$\min_{f\in\mathcal{F}} \sup_{\mathbb{P}\in\mathcal{P}(\mathcal{Z})} \left\{\mathbb{E}_{z\sim\mathbb{P}}[f(z)] : W_p(\mathbb{P},\mathbb{Q}) \leq \rho\right\}.$$

Quite often, we will consider a data-driven problem, in which the nominal distribution chosen as the empirical distribution $\mathbb{P}_n = \frac{1}{n}\sum_{i=1}^n \boldsymbol{\delta}_{z_i^n}$ independently sampled from the underlying true distribution $\mathbb{P}_{\text{true}}$, where $\boldsymbol{\delta}_z$ denotes the Dirac point mass on $z$. The dual problem of the inner supremum in (P) is defined as

$$\begin{cases} \min_{\lambda\geq 0}\left\{\lambda\rho^p + \mathbb{E}_{z\sim\mathbb{Q}}\left[\sup_{\tilde{z}\in\mathcal{Z}}\{f(\tilde{z}) - \lambda \mathsf{d}(\tilde{z},z)^p\}\right]\right\}, & p\in[1,\infty), \\ \mathbb{E}_{z\sim\mathbb{Q}}\left[\sup_{\tilde{z}\in\mathcal{Z}}\{f(\tilde{z}) : \mathsf{d}(\tilde{z},z)\leq\rho\}\right], & p=\infty. \end{cases} \tag{D}$$

An important result in Wasserstein distributionally robust optimization is that the strong duality holds in rather general conditions and particularly, for $\mathbb{Q} = \mathbb{P}_n$ and the setup described above; see Lemma EC.1 in Appendix EC.1 for more details.

We define the *Wasserstein regularizer* $\mathcal{R}$ as the difference between Wasserstein robust loss and nominal loss

$$\mathcal{R}_{\mathbb{Q},p}(\rho; f) := \sup_{\mathbb{P}\in\mathcal{P}(\mathcal{Z})} \left\{\mathbb{E}_{z\sim\mathbb{P}}[f(z)] : W_p(\mathbb{P},\mathbb{Q}) \leq \rho\right\} - \mathbb{E}_{\mathbb{Q}}[f].$$

By definition, when $\mathbb{Q} = \mathbb{P}_n$, the Wasserstein regularizer $\mathcal{R}_{\mathbb{P}_n,p}(\rho; f)$ can be viewed as a data-dependent regularizer of the loss $f$. Proposition 1 below is a consistency-type result under a growth condition, which shows that $\mathcal{R}_{\mathbb{P}_n,p}(\rho; f)$ converges to zero as the radius shrinks.

Proposition 1 **(Consistency).** *Let $p \in [1,\infty]$. Assume $f$ is upper semi-continuous for all $f \in \mathcal{F}$ and satisfies the following growth condition when $p \in [1,\infty)$:*

$$\exists z_0 \in \mathcal{Z},\ s.t.,\ \sup_{f\in\mathcal{F}} \limsup_{\mathsf{d}(\tilde{z},z_0)\to\infty} \frac{\left(f(\tilde{z}) - f(z_0)\right)_+}{\mathsf{d}(\tilde{z},z_0)^p} < \infty, \tag{G}$$

*where we use the convention that the ratio is zero if* $\operatorname{diam}(\mathcal{Z}) < \infty$. *Then*

$$\lim_{\rho\to 0} \mathcal{R}_{\mathbb{P}_n,p}(\rho; f) = 0 = \mathcal{R}_{\mathbb{P}_n,p}(0; f).$$

The assumptions in Proposition 1 are minimal in the following sense. The upper semi-continuity of $f$ is necessary to ensure $\mathcal{R}_{\mathbb{P}_n,p}(0; f) = 0$, and the growth condition is necessary to ensure the finiteness of the Wasserstein robust loss [26]. Proposition 1 generalizes Theorem 3.6 in [24] by relaxing the convergence condition on the radius $\rho$. The main goal of this paper is to study the *convergence rate* of the Wasserstein regularizer $\mathcal{R}_{\mathbb{P}_n,p}(\rho; f)$ as $\rho \to 0$.

We close this section by introducing further notations used throughout the paper.

**Notation.** We denote by $\|\cdot\|_*$ the dual norm of a norm $\|\cdot\|$ when $\mathcal{Z}$ is a normed space, and use $\langle\cdot,\cdot\rangle$ to represent the associated bilinear form. The sup-norm of a function $h$ is denoted by $\|h\|_\infty$, and the Lipschitz norm of a Lipschitz continuous function $h : \mathcal{Z} \to \mathbb{R}$ is denoted by $\|h\|_{\text{Lip}} := \sup_{\tilde{z},z\in\mathcal{Z}} \frac{f(\tilde{z})-f(z)}{\mathsf{d}(\tilde{z},z)}$. For a function $h$ defined on a metric space, $\limsup_{\tilde{z}\to z} h(\tilde{z}) := \limsup_{\delta\downarrow 0}\{h(\tilde{z}) : 0 < \mathsf{d}(\tilde{z},z) < \delta\}$. The distance between a point $z \in \mathcal{Z}$ and a set $\mathcal{D} \subset \mathcal{Z}$ is defined as $\mathsf{d}(z,\mathcal{D}) := \inf_{\tilde{z}\in\mathcal{D}} \mathsf{d}(\tilde{z},z)$. The interior and closure of a set $A$ are denoted by $\operatorname{int}(A)$ and $\operatorname{cl}(A)$ respectively. We denote $a \wedge b = \min(a,b)$, $a \vee b = \max(a,b)$ and $a_+ = \max(a,0)$. The support of a distribution is denoted by $\operatorname{supp}\mathbb{Q}$. We use $O$ and $O_p$ to represent the big O and the big O in probability notations respectively and use $\tilde{O}$ when we omit the polylog term. We use $\mathbb{P}_\otimes$ or $\mathbb{E}_\otimes$ to indicate that the probability or expectation is evaluated with respect to the sampling distribution, namely the $n$-fold product distribution $\otimes_{i=1}^n \mathbb{P}_{\text{true}}$ over $\mathcal{Z}^n$.

# 3. Variation Regularization Effect of Wasserstein DRO

In this section we study the regularization effect of Wasserstein DRO. In Section 3.1, we introduce a new concept, the variation of a function. Then in Sections 3.2 and 3.3, we show that the variation is a natural quantity characterizing the convergence rate of the Wasserstein regularizer $\mathcal{R}_{\mathbb{P}_n,p}(f;\rho)$ as $\rho \to 0$. Quite often, we focus on a radius selection rule $\rho_n = \rho_0/\sqrt{n}$ for some $\rho_0 > 0$, which has been empirically used in practice and also theoretically investigated in [12, 47, 13, 25]. For ease of exposition, from Section 3.1 to Section 3.3, we present our results when $\mathcal{Z}$ is a subset of a Banach space. Extensions to general metric spaces are presented in Section 3.4.

## 3.1. Variation

We introduce a new concept, the variation of a function, inspired from the total variation of a real-valued function.

Definition 1 (Variation). Let $q \in [1,\infty]$ and $f$ be a continuous function on $\mathcal{Z}$. When $q \in [1,\infty)$, assume $\nabla f$ exists $\mathbb{Q}$-almost everywhere. The *variation* of $f$ with respect to $\mathbb{Q}$ is defined as

$$\mathcal{V}_{\mathbb{Q},q}(f) := \begin{cases} \| \|\nabla f\|_* \|_{\mathbb{Q},q}, & q \in [1,\infty), \\ \mathbb{Q}\text{-ess}\sup_{z\in\mathcal{Z}} \sup_{\tilde{z}\neq z} \dfrac{(f(\tilde{z}) - f(z))_+}{\|\tilde{z}-z\|}, & q = \infty. \end{cases}$$

◇

In essence, the variation $\mathcal{V}_{\mathbb{Q},q}(f)$ is a weighted average of gradient norms over all data $z \in \operatorname{supp}\mathbb{Q}$. In particular, when $f$ is univariate and differentiable, $\mathcal{V}_{\mathbb{Q},1}(f)$ reduces to the usual representation of total variation of a function $\int_{\mathbb{R}} |f'(z)|dz$. For Lipschitz continuous function $f$, we have $\mathcal{V}_{\mathbb{Q},\infty}(f) \le \|f\|_{\mathrm{Lip}}$ and $\mathcal{V}_{\mathbb{Q},\infty}(f) = \|f\|_{\mathrm{Lip}}$ if $\operatorname{supp}\mathbb{Q} = \mathcal{Z}$.

As promised, we will bound $\mathcal{R}_{\mathbb{Q},p}(\rho;f)$ using $\mathcal{V}_{\mathbb{Q},q}(f)$. In particular, developing a lower bound is crucial to understanding the variation regularization effect of Wasserstein DRO. Indeed, once the lower bound holds, minimizing the Wasserstein robust loss controls the variation of the loss. In the sequel, we separate the cases of $p > 1$ and $p = 1$.

## 3.2. $p$-Wasserstein DRO ($p > 1$)

In this subsection we consider $p \in (1,\infty]$. We start by establishing upper and lower bounds for smooth loss functions in Section 3.2.1. Since we aim to provide statistical guarantees for loss functions that are possibly non-smooth, additional assumptions on the loss function class $\mathcal{F}$ as well as the underlying true distribution $\mathbb{P}_{\mathrm{true}}$ are needed. In Section 3.2.2, we study a simple example of piecewise linear loss to motivate proper assumptions, and the result for general non-smooth functions is developed in Theorem 1 and is further generalized in Section 3.4.

Throughout this subsection, we are mostly interested in piecewise smooth losses that cover many real-world applications. Specifically, we impose the following two assumptions.

Assumption 1 **(Piecewise smoothness)**. *For every $f \in \mathcal{F}$, there exists a partition $\mathcal{Z} = \bigcup_{1\le k\le K_f} \mathcal{Z}_{f,k}$, where $\mathcal{Z}_{f,j} \cap \mathcal{Z}_{f,k} = \emptyset$ for all $j \neq k$ such that $f$ is differentiable on $\operatorname{int}(\mathcal{Z}_{f,k})$, $1 \le k \le K_f$. Moreover, there exists $H \in \mathcal{L}^{\frac{p}{p-2}}(\mathbb{P}_{\mathrm{true}})$ when $p \in (2,\infty]$ and $H \in \mathcal{L}^\infty(\mathcal{Z})$ when $p \in (1,2]$ such that for any $\epsilon > 0$, there exists $\delta > 0$ such that for all $1 \le k \le K_f$ and $\tilde{z}, z \in \operatorname{int}(\mathcal{Z}_{f,k})$ with $\|\tilde{z}-z\| \le \delta$,*

$$\frac{\|\nabla f(\tilde{z}) - \nabla f(z)\|_*}{\|\tilde{z}-z\|} \le H(z) + \epsilon. \qquad \textbf{(S)}$$

*We denote by*

$$\mathcal{D}_f := \bigcup_{1\le k\neq j\le K_f} \mathrm{cl}(\mathcal{Z}_{f,j})\cap \mathrm{cl}(\mathcal{Z}_{f,k})$$

*the union set of intersections of pieces, which is assumed to be a $\mathbb{P}_{\mathrm{true}}$-null set for every $f\in\mathcal{F}$.*

By definition, all non-differentiable points of $f$ are contained in $\mathcal{D}_f$.

ASSUMPTION 2 **(Growth and jump of gradient)**.

(I) *When $p\in(1,\infty)$, assume there exists $M,L\ge 0$ such that for every $f\in\mathcal{F}$ and $\tilde{z},z\in\mathcal{Z}\setminus\mathcal{D}_f$,*

$$\|\nabla f(\tilde{z})-\nabla f(z)\|_*\le M+L\|\tilde{z}-z\|^{p-1}.$$

(II) *When $p=\infty$, assume there exists $M\ge 0$ and $\delta_0>0$ such that for every $f\in\mathcal{F}$ and $\tilde{z},z\in\mathcal{Z}\setminus\mathcal{D}_f$ with $\|\tilde{z}-z\|<\delta_0$,*

$$\|\nabla f(\tilde{z})-\nabla f(z)\|_*\le M.$$

Assumption 2 imposes a growth condition on the gradient norm when $p\in(1,\infty)$, consistent with the growth condition (G) on the loss; as well as a bounded jump condition on $f$, namely, the gap of gradient norms around a non-differentiable point is at most $M$.

**3.2.1. Smooth Losses** We first establish a result demonstrating the gradient regularization effect for smooth losses, whose detailed proof is given in Appendix EC.2.2.

LEMMA 1. *Let $p\in(1,\infty]$ and $\rho_n=\rho_0/\sqrt{n}$. Assume Assumption 1 holds with $K_f=1$ for all $f\in\mathcal{F}$ and Assumption 2(I) holds. Then there exists $\bar{\rho},C>0$ such that for all $\rho_0<\bar{\rho}$ and $f\in\mathcal{F}$,*

$$\left|\mathcal{R}_{\mathbb{P}_n,p}(\rho_n;f)-\rho_n\mathcal{V}_{\mathbb{P}_n,q}(f)\right|\le \rho_n^{2\wedge p}\big(C+\|H\|_{\mathbb{P}_n,\frac{p}{p-2}}1\{p>2\}\big).$$

We remark that under mild conditions, $(\|H\|_{\mathbb{P}_n,\frac{p}{p-2}}-\|H\|_{\mathbb{P}_{\mathrm{true}},\frac{p}{p-2}})_+$ is of the order $O_p(n^{-1/2})$. As a result, when $\rho_n=O(n^{-1/2})$, Lemma 1 gives a first-order Taylor expansion for the Wasserstein regularizer $\mathcal{R}_{\mathbb{P}_n,p}(\rho;f)$ of smooth losses for $p\in[2,\infty]$ with a remainder $O_p(n^{-1})$ uniformly for all $f\in\mathcal{F}$. The cases for $p=2$ and $p\in(2,\infty)$ in normed vector spaces have been developed in [55, Section 3.2] and [5, Remark 8] respectively. When $p\in(1,2)$, the order of the remainder $O(n^{-p/2})$ cannot be improved in general, as can be seen from the following example.

EXAMPLE 1. Consider $f(z)=z^p$, where $p\in(1,2)$ and $z\in\mathcal{Z}=(\mathbb{R}_+,|\cdot|)$. Suppose $\mathbb{P}_{\mathrm{true}}=\boldsymbol{\delta}_0$, which implies $\mathbb{P}_n=\boldsymbol{\delta}_0$ almost surely. Observe that

$$\sup_{\tilde{z}\in\mathbb{R}_+}\left\{f(\tilde{z})-f(0)-\lambda|\tilde{z}-0|^p\right\}=\begin{cases}+\infty, & \forall\lambda<1,\\ 0, & \forall\lambda\ge 1.\end{cases}$$

Thus by (D) we have

$$\mathcal{R}_{\mathbb{P}_n,p}(\rho;f)=\min_{\lambda\ge 0}\left\{\lambda\rho^p+\sup_{\tilde{z}\in\mathbb{R}_+}\left\{f(\tilde{z})-f(0)-\lambda|\tilde{z}-0|^p\right\}\right\}=\rho^p.$$

On the other hand, $\mathcal{V}_{\mathbb{P}_n,q}(f)=f'(0)=0$ almost surely. Hence $\mathcal{R}_{\mathbb{P}_n,p}(\rho;f)-\rho\mathcal{V}_{\mathbb{P}_n,q}(f)=\rho^p$ for all $\rho\ge 0$. ♣

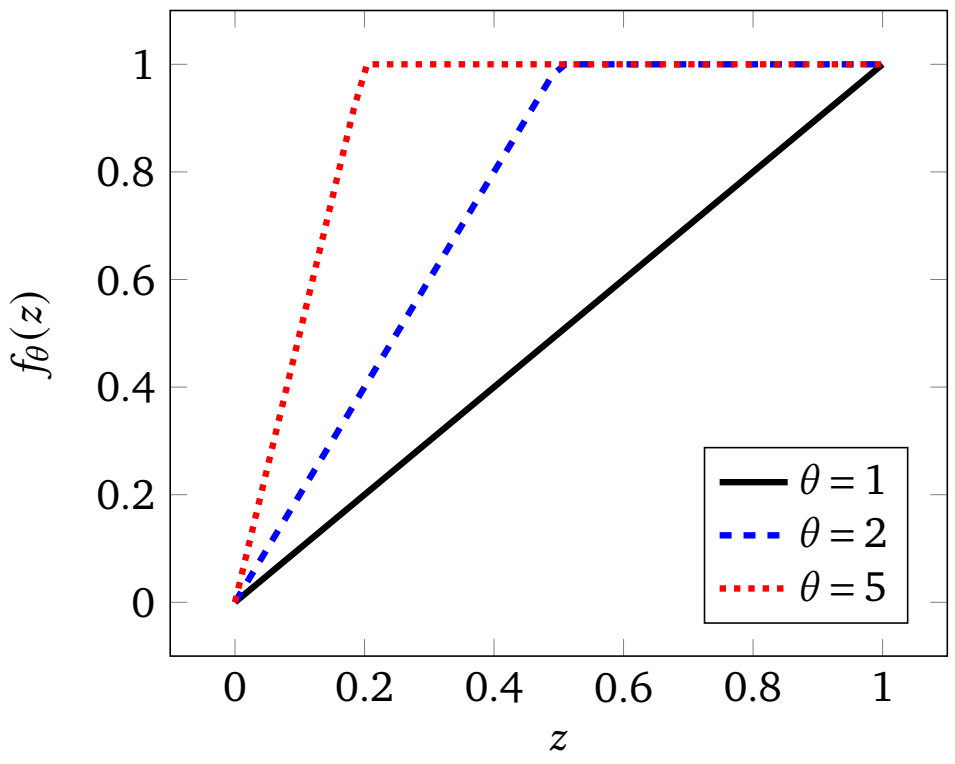

**Figure 1** $f_\theta(z) = \theta z \wedge 1,\ z \in [0,1]$

**3.2.2. Non-smooth Losses** Next, we consider non-smooth losses. Proofs of the results in this subsection can be found in Appendix EC.2.3. We start with an illustrating example of piecewise linear functions.

EXAMPLE 2. Let $\mathcal{Z} = [0,1] \subset (\mathbb{R}, |\cdot|)$. Consider

$$f_\theta(z) = \theta z \wedge 1, \text{ where } \theta \geq 0,$$

which is illustrated in Figure 1. Then $f'_\theta(z) = \theta$ when $z \in [0, 1/\theta)$ and is 0 when $z \in (1/\theta, 1]$. Thus, whenever $1/\theta \notin \operatorname{supp} \mathbb{P}_n$ which holds almost surely, we have

$$\mathcal{V}_{\mathbb{P}_n,1}(f_\theta) = \theta \mathbb{E}_{\mathbb{P}_n}[1\{z < 1/\theta\}].$$

Suppose $\rho_n = \rho_0/\sqrt{n}$ for some $\rho_0 > 0$. Using the dual form (D),

$$\begin{aligned}\mathcal{R}_{\mathbb{P}_n,\infty}(\rho_n; f_\theta) &= \mathbb{E}_{\mathbb{P}_n}\Big[\sup_{0\leq \tilde{z} \leq 1}\big\{f_\theta(\tilde{z}) - f_\theta(z) : |\tilde{z} - z| \leq \rho_n\big\}\Big] \\ &= \rho_n \theta \mathbb{E}_{\mathbb{P}_n}[1\{z < 1/\theta\}] - \rho_n \theta \mathbb{E}_{\mathbb{P}_n}[(1/\theta - z)1\{1/\theta - \rho_n < z < 1/\theta\}],\end{aligned}$$

where the second term indicates that for a point $z$ close to the non-differentiable point $1/\theta$ of $f_\theta(z)$, perturbing $z$ leads to a change of loss by at most $\rho_n(1 - \theta z)$ rather than $\rho_n \theta$. It then follows that

$$\rho_n \mathcal{V}_{\mathbb{P}_n,\infty}(f_\theta) - \mathcal{R}_{\mathbb{P}_n,\infty}(\rho_n; f_\theta) = \rho_n \theta \mathbb{E}_{\mathbb{P}_n}[(1/\theta - z)1\{1/\theta - \rho_n < z < 1/\theta\}].$$

Therefore, if

$$\mathbb{E}_{\mathbb{P}_n}[1\{1/\theta - \delta < z < 1/\theta\}] = O_p(\delta),$$

then the gap would be of the order $O_p(1/n)$. ♣

Motivated by Example 2, we impose the following continuity assumption around non-smooth points for the underlying data-generating distribution. Recall from Assumption 1 that $\mathcal{D}_f$ is the union of intersections of pieces of $f$.

ASSUMPTION 3 **(Bounded density)**.

$$\limsup_{\delta \downarrow 0} \sup_{f \in \mathcal{F}: \mathcal{D}_f \neq \varnothing} \frac{\mathbb{P}_{\text{true}}\{z : 0 < \mathrm{d}(z, \mathcal{D}_f) < \delta\}}{\delta} < \infty.$$

When $p \in (1,\infty)$, we impose an additional assumption on the normalized gradient norm $\frac{\|\nabla f(z)\|_*}{\|\,\|\nabla f\|_*\|_{\mathbb{P}_{\text{true}},q}}$.

Assumption 4 **(Growth of normalized gradient norm)**. *Let $p \in (1,\infty)$. For $z \in \mathcal{Z} \setminus \mathcal{D}_f$, define $w_f(z) := \left(\frac{\|\nabla f(z)\|_*}{\|\|\nabla f\|_*\|_{\mathbb{P}_{\text{true}},q}}\right)^{\frac{1}{p-1}}$. Assume there exists $c_1, c_2, c_3 > 0$ such that for all $f \in \mathcal{F}$ with $\|\|\nabla f\|_*\|_{\mathbb{P}_{\text{true}},q} > 0$ and $\mathcal{D}_f \neq \varnothing$ and for all $z \in \mathcal{Z} \setminus \mathcal{D}_f$,*

$$c_3 \leq w_f(z)^{p-1} \leq c_1 + c_2 \mathsf{d}(z, \mathcal{D}_f)^{p-1}.$$

It specifies the growth and jump condition deviated from non-differentiable points for the normalized gradient norm $w_f$. Practical examples satisfying this condition will be provided in Section 4. In Appendix A we also provide an alternative result when $w_f$ does not have a positive uniform lower bound $c_3$.

We define classes of functions

$$\mathcal{I}_\rho := \left\{z \mapsto 1\{\mathsf{d}(z, \mathcal{D}_f) < \rho\} : f \in \mathcal{F}, \mathcal{D}_f \neq \varnothing\right\}. \tag{1}$$

$$\mathcal{E} := \left\{\mathsf{d}(\cdot, \mathcal{D}_f) : f \in \mathcal{F} \text{ with } \mathcal{D}_f \neq \varnothing\right\}. \tag{2}$$

Recall the *Rademacher complexity* of a function class $\mathcal{H}$ with respect to a sample $\{z_i^n\}_{i=1}^n$ is defined as $\mathfrak{R}_n(\mathcal{H}) := \mathbb{E}_\sigma\left[\sup_{h\in\mathcal{H}} \frac{1}{n}\sum_{i=1}^n \sigma_i h(z_i^n)\right]$, where $\sigma_i$'s are i.i.d. Rademacher random variables with $\mathbb{P}\{\sigma_i = \pm 1\} = \frac{1}{2}$. The Rademacher complexity of the function class $\mathcal{H}$ with respect to $\mathbb{P}_{\text{true}}$ for sample size $n$ is defined as $\mathbb{E}_\otimes[\mathfrak{R}_n(\mathcal{H})]$. Recall also that the *covering number* $\mathcal{N}(\epsilon; \mathcal{H}, \mathsf{d}_\mathcal{H})$ of a function class $\mathcal{H}$ with respect to a metric $\mathsf{d}_\mathcal{H}$ is defined as the smallest cardinality of an $\epsilon$-cover of $\mathcal{F}$; here $\mathcal{F}_\epsilon$ is an *$\epsilon$-cover* of $\mathcal{F}$ if for each $f \in \mathcal{F}$, there exists $\tilde{f} \in \mathcal{F}_\epsilon$ such that $\mathsf{d}_\mathcal{F}(\tilde{f}, f) \leq \epsilon$.

Now we are ready to state the main result in this subsection.

Theorem 1 **($p$-Wasserstein DRO)**.

(I) *Let $p = \infty$. Assume Assumptions 1, 2 are in force. Then there exists $\bar{\rho} > 0$ such that for all $\rho < \bar{\rho}$ and $f \in \mathcal{F}$,*

$$\left|\mathcal{R}_{\mathbb{P}_n,\infty}(\rho; f) - \rho\|\|\nabla f\|_*\|_{\mathbb{P}_n,1}\right| \leq \rho^2 \|H\|_{\mathbb{P}_n,1} + M\mathbb{E}_{\mathbb{P}_n}[(\rho - \mathsf{d}(z, \mathcal{D}_f))_+].$$

(II) *Let $p \in (1,\infty)$ and $\rho_n = \rho_0/\sqrt{n}$. Assume Assumptions 1, 2, 4 are in force. Then there exists $\bar{\rho}, C_1, C_2 > 0$ such that for all $\rho_0 < \bar{\rho}$ and $f \in \mathcal{F}$,*

$$\left|\mathcal{R}_{\mathbb{P}_n,p}(\rho_n; f) - \rho_n\|\|\nabla f\|_*\|_{\mathbb{P}_n,q}\right| \leq \rho_n^{2\wedge p}(\|H\|_{\mathbb{P}_n,\frac{p}{p-2}} 1\{p > 2\} + C_1) + M\mathbb{E}_{\mathbb{P}_n}\left[\left(C_2\rho_n - \mathsf{d}(z, \mathcal{D}_f)\right)_+\right].$$

(III) *Assume Assumption 3 holds. Let $t > 0$. Then there exists $\bar{\rho}, C > 0$ such that for all $\rho < \bar{\rho}$, with probability at least $1 - e^{-t}$, for every $f \in \mathcal{F}$,*

$$\mathbb{E}_{\mathbb{P}_n}[(\rho - \mathsf{d}(z, \mathcal{D}_f))_+] \leq C\rho^2 + 2\rho\mathbb{E}_\otimes[\mathfrak{R}_n(\mathcal{I}_\rho)] + \rho\sqrt{\frac{t}{2n}},$$

*and with probability at least $1 - e^{-t}$, for every $f \in \mathcal{F}$,*

$$\mathbb{E}_{\mathbb{P}_n}[(\rho - \mathsf{d}(z, \mathcal{D}_f))_+] \leq 2C\rho^2 + \frac{48\rho}{\sqrt{n}}\int_0^1 \sqrt{\log \mathcal{N}(\epsilon\rho; \mathcal{E}, \|\cdot\|_{\mathbb{P}_n,2})}d\epsilon + \rho\sqrt{\frac{t}{2n}}.$$

For many important applications, it can be shown that when $\rho_n = O(1/\sqrt{n})$, $\mathbb{E}_{\mathbb{P}_n}[(\rho - \mathsf{d}(z, \mathcal{D}_f))_+] = O(1/n)$. Thereby, an immediate consequence of Theorem 1 is that for $p \in (1,\infty]$, the $p$-Wasserstein DRO (P) is asymptotically equivalent to the empirical variation regularization problem (V):

$$\min_{f\in\mathcal{F}} \sup_{\mathbb{P}: W_p(\mathbb{P},\mathbb{P}_n) \leq \rho_n} \mathbb{E}_{z\sim\mathbb{P}}[f(z)] = \min_{f\in\mathcal{F}} \left\{\mathbb{E}_{z\sim\mathbb{P}_n}[f(z)] + \rho_n \mathcal{V}_{\mathbb{P}_n,q}(f)\right\} + O_p(1/n).$$

Comparing to the smooth case (Lemma 1), the major difference is that the statement becomes probabilistic. In Section 4, we illustrate the case $p \in (1,\infty)$ in multi-item newsvendor (Example 5) and leaky ReLU neural networks (Example 9). In Section 5, we demonstrate the case $p = \infty$ for adversarial learning (Example 12).

### 3.3. 1-Wasserstein DRO

The case of $p = 1$ turns out to be qualitatively different from $p > 1$, as hinted from the remainder term of Lemma 1 when $p \in (1, 2)$. We study a simple example in Section 3.3.1 that motivates our main result in Section 3.3.2.

#### 3.3.1. A Motivating Example

EXAMPLE 3. Let $\mathcal{Z} = [0, 1]$ with $\mathrm{d}(\tilde{z}, z) = |\tilde{z} - z|$ and $\mathbb{Q} = \mathrm{Uniform}([0, 1])$. Consider

$$f_\theta(z) = (\theta z - \theta + 1)_+^2, \quad \theta \geq 1,$$

illustrated in Figure 3. Then $f_\theta(\cdot)$ is differentiable and $|\partial f_\theta|(z) = |f_\theta'(z)| = 2\theta(\theta z - \theta + 1)_+$. By Definition

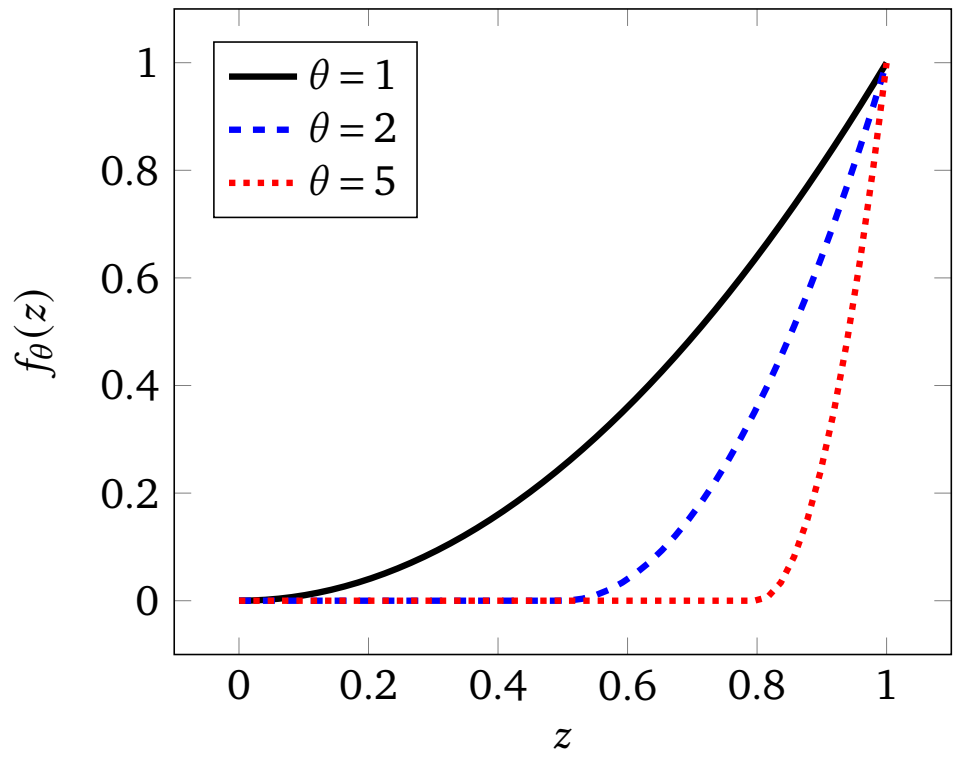

**Figure 2** $f_\theta(z) = (\theta z - \theta + 1)_+^2$, $z \in [0, 1]$

2, we have $\mathcal{V}_{\mathbb{Q},\infty}(f_\theta) = \|\mathrm{I}_{f_\theta}\|_{\mathbb{Q},\infty} = |f_\theta'(1)| = 2\theta$. Let $\rho \in (0, 1/2)$. We claim that the worst-case distribution $\mathbb{P}_*$ has the form

$$\mathbb{P}_* = \mathbb{Q}_{|\{z<z_*\}} + \mathbb{Q}\{z \geq z_*\} \cdot \boldsymbol{\delta}_1, \quad z_* \in [0, 1],$$

where $\mathbb{Q}_{|\{z<z_*\}}$ is the restriction of $\mathbb{Q}$ on $\{z < z_*\}$. To see this, observe from the convexity of $f$ that, for any $\lambda \geq 0$, $\sup_{\tilde{z}\in\mathcal{Z}}\{f_\theta(\tilde{z}) - f_\theta(z) - \lambda|\tilde{z} - z|\}$ attains its maximum at $\tilde{z} = 1$ and there exists $z_* \in [0, 1]$ such that

$$\sup_{\tilde{z}\in\mathcal{Z}}\{f_\theta(\tilde{z}) - f_\theta(z) - \lambda|\tilde{z} - z|\} \begin{cases} > 0, & \forall z > z_*, \\ = 0, & \forall z = z_*. \end{cases}$$

Consequently, using the dual formulation (D) and the structure of the worst-case distribution [26], we prove the claim. Solving for $\rho = \mathbb{E}_\mathbb{Q}[(1 - z)1\{z > z_*\}]$ yields $z_* = 1 - \sqrt{2\rho}$. It follows that

$$\mathcal{R}_{\mathbb{Q},1}(\rho; f_\theta) = \mathbb{E}_\mathbb{Q}\left[\left(1 - (\theta z - \theta + 1)_+^2\right)1\{z > z_*\}\right] = \begin{cases} 2\rho\theta - (2\rho)^{\frac{3}{2}}\theta^2/3, & \theta \leq 1/\sqrt{2\rho}, \\ \frac{2}{3\theta} + (\sqrt{2\rho} - 1/\theta)_+, & \theta > 1/\sqrt{2\rho}. \end{cases}$$

Therefore,

$$\rho\mathcal{V}_{\mathbb{Q},\infty}(f_\theta) - \mathcal{R}_{\mathbb{Q},1}(\rho; f_\theta) = \begin{cases} (2\rho)^{\frac{3}{2}}\theta^2/3, & \theta \leq 1/\sqrt{2\rho}, \\ 2\rho\theta - \frac{2}{3\theta} - (\sqrt{2\rho} - 1/\theta)_+, & \theta > 1/\sqrt{2\rho}. \end{cases}$$

This shows that the remainder $\rho\mathcal{V}_{\mathbb{Q},\infty}(f_\theta) - \mathcal{R}_{\mathbb{Q},1}(\rho; f_\theta)$ may not be of the desired order $O(\rho^2)$, and can even be *linear* in $\rho$ for $\theta > 1/\sqrt{2\rho}$. ♣

In Example 3, the worst-case distribution perturbs points in $[z_*, 1]$ – which have large slopes – to the boundary point 1 so as to maximize the loss. Recall from Definition 1 that the maximum rate of change of loss by perturbing a point $z$ equals $\sup_{\tilde{z}\neq z}(f(\tilde{z})-f(z))_+/\|\tilde{z}-z\|$; and that $\mathcal{V}_{\mathbb{Q},\infty}(f)$ measures the largest possible change of loss by perturbation among all $z \in \operatorname{supp}\mathbb{Q}$. The gap between these two quantities can lead to a remainder with an undesired order. Specifically, in Example 3 we have

$$\frac{\sup_{\tilde{z}\neq z_*}(f(\tilde{z})-f(z_*))_+/\|\tilde{z}-z_*\|}{\mathcal{V}_{\mathbb{Q},\infty}(f)} = \frac{2\theta(\theta(1-\sqrt{2\rho})-\theta+1)}{2\theta} = 1-\theta\sqrt{2\rho}.$$

However, in order to have a $O(\rho^2)$ remainder, every perturbed point $z \in \mathcal{Z}$ should satisfy

$$\frac{\sup_{\tilde{z}\neq z}(f(\tilde{z})-f(z))_+/\|\tilde{z}-z\|}{\mathcal{V}_{\mathbb{Q},\infty}(f)} \geq 1-O(\rho). \tag{3}$$

As a result, perturbing points close to $z_*$ does not provide sufficient change of the loss compared to $\mathcal{V}_{\mathbb{Q},\infty}(f)$, leading to a large remainder. In high dimensions, the condition (3) can be difficult to satisfy. For example, consider $\mathcal{Z} \subset \mathbb{R}^d$, $\rho < 1$, $f$ is twice differentiable and $z_{\max}$ is the unique point satisfying $\sup_{\tilde{z}\neq z}(f(\tilde{z})-f(z_{\max}))_+/\|\tilde{z}-z_{\max}\| = \mathcal{V}_{\mathbb{Q},\infty}(f)$. It may happen that only points in the neighborhood $\{z : \|z-z_{\max}\| \leq c\rho\}$ satisfy the condition (3). Suppose $\mathbb{Q}$ is continuous with bounded density, then this set has $\mathbb{Q}$-measure only $O(\rho^d)$, where $d$ is the dimension of $\mathcal{Z}$. Hence, most points being perturbed by the worst-case distribution would not satisfy (3).

The discussion above suggests that for $p = 1$, the remainder $\rho\mathcal{V}_{\mathbb{Q},\infty}(f) - \mathcal{R}_{\mathbb{Q},1}(\rho; f)$ cannot be of the desired order $O(\rho^2)$ in general. Fortunately, as will be formalized in the next subsection, one can show that $\mathcal{R}_{\mathbb{Q},1}(\rho; f)$ achieves a fraction of $\rho\mathcal{V}_{\mathbb{Q},\infty}(f)$ uniformly for all $f \in \mathcal{F}$ under mild conditions. Thereby, the variation of loss is still under control by minimizing the Wasserstein robust loss.

**3.3.2. A Sandwich Theorem** Motivated by the discussion in the previous section, particularly condition (3), we develop the following Theorem 2, which are instantiated under two important situations (Corollaries 1 and 2). The proofs are given in Appendix EC.2.5.

Define

$$\mathrm{d}_{\mathcal{F}}(f,\tilde{f}) := \max\left(\|f-\tilde{f}\|_\infty,\ \left|\|f\|_{\mathrm{Lip}} - \|\tilde{f}\|_{\mathrm{Lip}}\right|\right).$$

Theorem 2 **(1-Wasserstein DRO)**. *Let $p = 1$. Assume every $f \in \mathcal{F}$ is Lipschitz continuous. Assume further that there exists $\varepsilon > 0$, $\delta_n, \eta \in (0,1]$ such that for every $f \in \mathcal{F}$, there exists $\mathcal{Z}_f \subset \mathcal{Z}$ and $\mathfrak{T}_f : \mathcal{Z}_f \to \mathcal{Z}$ such that with probability at least $1-\delta_n$,*

$$\begin{aligned} & f(\mathfrak{T}_f(z)) - f(z) \geq \eta(\|f\|_{\mathrm{Lip}} - \varepsilon)\|\mathfrak{T}_f(z) - z\|, \quad \forall z \in \mathcal{Z}_f, \\ & \mathbb{E}_{\mathbb{P}_n}\left[\|\mathfrak{T}_f(z) - z\| 1\{z \in \mathcal{Z}_f\}\right] > 0. \end{aligned} \tag{T}$$

*Suppose $\rho_n \leq \min_{f\in\mathcal{F}} \mathbb{E}_{\mathbb{P}_n}[\|\mathfrak{T}_f(z)-z\|1\{z \in \mathcal{Z}_f\}]$. Then with probability at least $1 - \mathcal{N}(\frac{1}{n}; \mathcal{F}, \mathrm{d}_{\mathcal{F}}) \cdot \delta_n$,*

$$\eta\rho_n\mathcal{V}_{\mathbb{P}_n,\infty}(f) - \rho_n\varepsilon - (1 \vee \rho_n)/n \leq \mathcal{R}_{\mathbb{P}_n,1}(\rho_n; f) \leq \rho_n\mathcal{V}_{\mathbb{P}_n,\infty}(f).$$

This theorem shows that the Wasserstein regularizer $\mathcal{R}_{\mathbb{P}_n,1}(\rho; f)$ is sandwiched by $\rho\mathcal{V}_{\mathbb{P}_n,\infty}(f)$ and its *fraction* $\eta\rho\mathcal{V}_{\mathbb{P}_n,\infty}(f)$ which, according to Example 3, is generally the best one can hope for. Assumption (**T**) means that every point $z \in \mathcal{Z}_f$ can be perturbed to some point $\mathfrak{T}_f(z)$ resulting in an increment no less than a fixed fraction of $\mathcal{V}_{\mathbb{P}_n,\infty}(f)$; and that the total perturbations $\mathbb{E}_{\mathbb{P}_n}[\mathrm{d}(\mathfrak{T}_f(z), z)1\{z \in \mathcal{Z}_f\}]$ has a positive lower bound uniformly for all $f \in \mathcal{F}$. There is a tradeoff between $\eta$ and $\delta_n$ – one can increase $\eta$ at the cost of a smaller $\delta_n$. For loss functions that are spurious, i.e., with large probability $\|\nabla f(z)\|_*$ is much smaller than $\mathcal{V}_{\mathbb{P}_n,\infty}(f)$, we would like to reduce the fraction $\eta$ in order to increase the probability bound $\delta$ so that it has a mild dependence on the dimension of $\mathcal{Z}$. Below we provide two important situations where the condition (**T**) can be satisfied either probabilistically or deterministically.

Denote by $H(a\|b) := a\log\frac{a}{b} + (1-a)\log\frac{1-a}{1-b}$.

Corollary 1 **(Data-driven 1-Wasserstein DRO)**. *Assume every* $f \in \mathcal{F}$ *is Lipschitz continuous. Suppose* $\rho_n = \rho_0/\sqrt{n}$ *for some* $\rho_0 > 0$. *Assume every* $f \in \mathcal{F}$ *is* $\hbar$*-semi-convex, i.e., there exists* $\hbar \in \mathbb{R}$ *such that*

$$f(\tilde{z}) - f(z) \geq g^\top(\tilde{z} - z) - \hbar\|\tilde{z} - z\|^2, \quad \forall z, \tilde{z} \in \mathcal{Z},$$

*where* $g$ *is any element in the subdifferential* $\partial f(z)$. *Assume further that there exists* $\eta \in (0,1]$ *such that*

$$\alpha := \inf_{f \in \mathcal{F}} \mathbb{P}_{\text{true}}\Big\{z : \sup_{g \in \partial f(z)} \|g\|_* \geq \eta\|f\|_{\text{Lip}}\Big\} \in (0,1).$$

*Let* $c < \frac{\alpha}{1-\alpha} \wedge \frac{1-\alpha}{\alpha}$. *Then the condition* (**T**) *is satisfied by setting*

$$\varepsilon = c\hbar\rho_n^2, \quad \mathcal{Z}_f = \Big\{z : \sup_{g \in \partial f(z)} \|g\|_* \geq \eta\|f\|_{\text{Lip}}\Big\}, \quad \delta_n = \exp\big(-nH(c\|\tfrac{\alpha}{1-\alpha} \wedge \tfrac{1-\alpha}{\alpha})\big).$$

*In addition, with probability at least* $1 - \exp\big(-nH(c\|\frac{\alpha}{1-\alpha} \wedge \frac{1-\alpha}{\alpha}) + \log\mathcal{N}(\frac{1}{n}; \mathcal{F}, \mathsf{d}_{\mathcal{F}})\big)$,

$$\eta\rho_n\mathcal{V}_{\mathbb{P}_n,\infty}(f) - c\hbar\rho_n^2 - (1 \vee \rho_n)/n \leq \mathcal{R}_{\mathbb{P}_n,1}(\rho_n; f) \leq \rho_n\mathcal{V}_{\mathbb{P}_n,\infty}(f).$$

An illustration of this result for linear prediction with Lipschitz loss on a bounded domain is given in Example 7 in Section 4.

Corollary 2 **(Lipschitz regularization)**. *Assume every* $f \in \mathcal{F}$ *is Lipschitz continuous. Suppose* $\text{diam}(\mathcal{Z}) = \infty$ *and there exists* $z_0 \in \mathcal{Z}$ *such that*

$$\limsup_{\|\tilde{z} - z_0\| \to \infty} \frac{f(\tilde{z}) - f(z_0)}{\|\tilde{z} - z_0\|} = \|f\|_{\text{Lip}}, \quad \forall f \in \mathcal{F}, \tag{L}$$

*then* (**T**) *is satisfied for any* $\varepsilon > 0$ *with* $\eta = 1$ *and* $\delta = 0$. *In addition, for all* $\rho \geq 0$ *and* $f \in \mathcal{F}$,

$$\mathcal{R}_{\mathbb{P}_n,1}(\rho; f) = \rho\mathcal{V}_{\mathbb{P}_n,\infty}(f) = \rho\|f\|_{\text{Lip}}.$$

This provides a situation of exact equivalence between Wasserstein DRO and regularization. As detailed in the proof, if (**L**) holds for some $z_0 \in \mathcal{Z}$ then it holds for every $z \in \mathcal{Z}$. Hence, condition (**L**) means that the Lipschitz norm is attained approximately between $z \in \text{supp}\,\mathbb{P}_n$ and some distant point $\tilde{z}$: for any $\epsilon > 0$ and $r > 0$ there exists $\tilde{z} =: \mathfrak{T}_f(z)$ such that $\|\mathfrak{T}_f(z) - z\| > r$ and $f(\mathfrak{T}_f(z)) - f(z) \geq (\|f\|_{\text{Lip}} - \epsilon)\|\mathfrak{T}_f(z) - z\|$. The (approximately) worst-case distribution perturbs some point $z_{i_0} \in \text{supp}\,\mathbb{P}_n$ to $\mathfrak{T}_f(z_{i_0})$ with tiny probability $\delta/n$ where $\delta \in (0,1)$, and therefore has the form

$$\frac{1}{n}\sum_{i \neq i_0} \boldsymbol{\delta}_{z_i} + \frac{1-\delta}{n}\boldsymbol{\delta}_{z_{i_0}} + \frac{\delta}{n}\boldsymbol{\delta}_{\mathfrak{T}_f(z_{i_0})}.$$

Condition (**L**) can be satisfied when $f$ is convex and Lipschitz, which has been considered in [24] and in [47]. It also holds for non-convex losses; a one-dimensional example is the inverse $S$-shaped curve plotted in Figure 3. We illustrate this corollary in Example 8 in Section 4 for linear prediction with Lipschitz loss on an unbounded domain.

### 3.4. Theory for General Metric Spaces

In previous subsections, we primarily focus on losses on a Banach space. In this subsection, we extend the results to a general metric space without isolated point. To this end, we first introduce the following notion of *slope* adopted from [17, 2], which measures the modulus of continuity of a function on a metric space without isolated points.

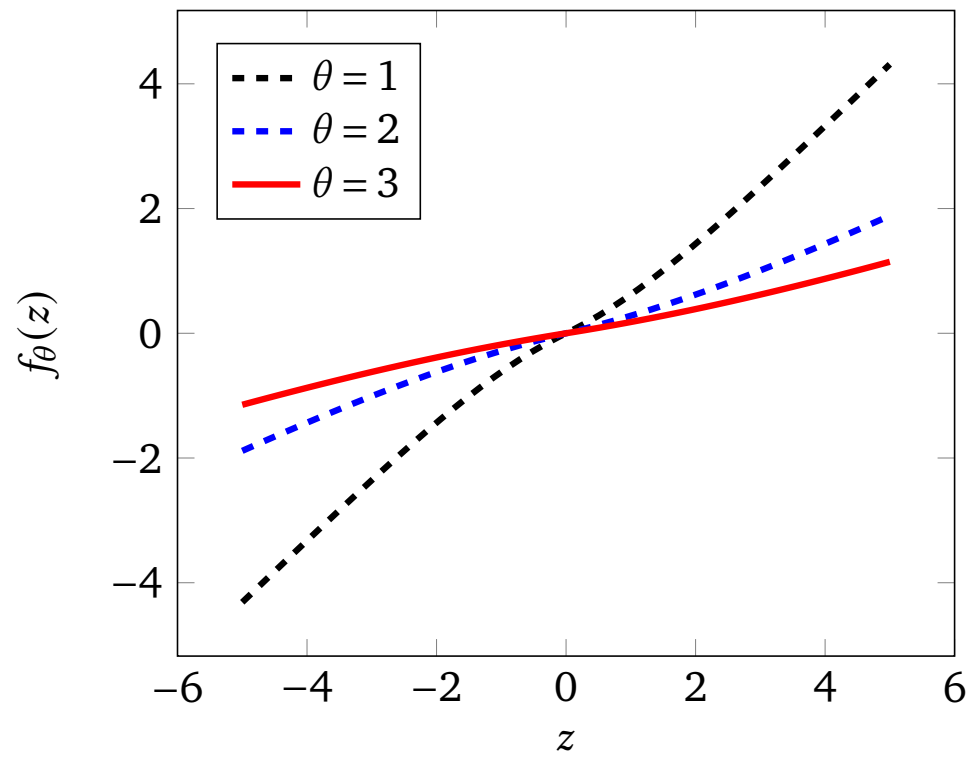

**Figure 3** $f_\theta(z) = \mathrm{sgn}(z)\ln((1+\exp(z/\theta))/2),\ z \in \mathbb{R}$

Definition 2 (Slopes and variation). *The local slope $|\partial f|(z)$ and global slope $\mathfrak{l}_f(z)$ of a function $f : \mathcal{Z} \to \mathbb{R}$ at $z \in \mathcal{Z}$ is defined as*

$$|\partial f|(z) := \limsup_{\tilde{z}\to z} \frac{(f(\tilde{z}) - f(z))_+}{\mathrm{d}(\tilde{z}, z)},$$
$$\mathfrak{l}_f(z) := \sup_{\tilde{z}\neq z} \frac{(f(\tilde{z}) - f(z))_+}{\mathrm{d}(\tilde{z}, z)}.$$

*The variation of a function $f$ with respect to a distribution $\mathbb{Q}$ is defined as*

$$\mathcal{V}_{\mathbb{Q},q}(f) := \begin{cases} \| \, |\partial f| \, \|_{\mathbb{Q},q}, & q \in [1,\infty), \\ \|\mathfrak{l}_f\|_{\mathbb{Q},\infty}, & q = \infty. \end{cases}$$

◇

Our definition of slope generalizes the slope for univariate functions. The local slope $|\partial f|(z)$ measures the magnitude of the change of the loss when perturbing $z$ locally, while the global slope $\mathfrak{l}_f(z)$ measures the largest magnitude of the loss when perturbing $z$ to any point in $\mathcal{Z}$. Obviously we have $|\partial f| \le \mathfrak{l}_f$. Slopes are well-defined for a very broad class of continuous but *not necessarily differentiable* loss functions on any metric space without isolated points. The next example demonstrates our definition of variation reduces to Definition 1 when $\mathcal{Z}$ is a Banach space.

Example 4. Suppose $\mathcal{Z}$ is an open subset of a Banach space $(\mathcal{B}, \|\cdot\|)$ and $f : \mathcal{B} \to \mathbb{R}$.

(I) When $f$ is differentiable, by definition and Cauchy-Schwarz inequality, we have

$$|\partial f|(z) = \limsup_{\tilde{z}\to z} \frac{f(\tilde{z}) - f(z)}{\|\tilde{z} - z\|} = \limsup_{\tilde{z}\to z} \frac{\langle \nabla f(z), \tilde{z} - z\rangle}{\|\tilde{z} - z\|} = \|\nabla f(z)\|_*.$$

Thus $\| \, |\partial f| \, \|_{\mathbb{Q},q} = \| \, \|\nabla f\|_* \, \|_{\mathbb{Q},q}$.

(II) When $f$ is Lipschitz, by definition $|\partial f|(z) \le \mathfrak{l}_f(z) \le \|f\|_{\mathrm{Lip}}$. Thus

$$\|\mathfrak{l}_f\|_{\mathbb{Q},\infty} = \mathbb{Q}\text{-}\operatorname*{ess\,sup}_{z\in\mathcal{Z}} \mathfrak{l}_f(z) = \mathbb{Q}\text{-}\operatorname*{ess\,sup}_{z\in\mathcal{Z}} \sup_{\tilde{z}\neq z} \frac{(f(\tilde{z}) - f(z))_+}{\mathrm{d}(\tilde{z}, z)}.$$

♣

Define

$$G_f(\delta, z) := \sup_{\tilde{z}\in\mathcal{Z}:\, \mathrm{d}(\tilde{z},z)\le\delta} f(\tilde{z}) - f(z), \qquad \delta \ge 0,\ z \in \mathcal{Z}. \tag{4}$$

In Lemma EC.8 in Appendix EC.2.2 we will show that Assumptions 1 and 2 imply Assumption 5.

Assumption 5 **(Growth, continuity and jump)**.

(I) *When $p=\infty$, assume there exists $\delta_0, M \geq 0$, and $H \in \mathcal{L}^1(\mathbb{P}_{\text{true}})$ such that for all $f \in \mathcal{F}$ and all $z \in \mathcal{Z}$,*

$$\left|G_f(\delta,z) - |\partial f|(z)\delta\right| \leq H(z)\delta^2 + M(\delta - \mathsf{d}(z,\mathcal{D}_f))_+, \quad \forall \delta < \delta_0.$$

(II) *When $p \in (2,\infty)$, assume there exists $\delta_0, L, M \geq 0$, and $H \in \mathcal{L}^{\frac{p}{p-2}}(\mathbb{P}_{\text{true}})$ such that for all $f \in \mathcal{F}$ and all $z \in \mathcal{Z}$,*

$$G_f(\delta,z) - |\partial f|(z)\delta \leq H(z)\delta^2 + L\delta^p + M(\delta - \mathsf{d}(z,\mathcal{D}_f))_+, \quad \forall \delta \geq 0,$$
$$G_f(\delta,z) - |\partial f|(z)\delta \geq -H(z)\delta^2 - M(\delta - \mathsf{d}(z,\mathcal{D}_f))_+, \qquad \forall \delta \leq \delta_0.$$

(III) *When $p \in (1,2]$, assume there exists $\hbar, M \geq 0$ such that for all $f \in \mathcal{F}$ and all $z \in \mathcal{Z}$,*

$$-\hbar\delta^2 - M(\delta - \mathsf{d}(z,\mathcal{D}_f))_+ \leq G_f(\delta,z) - |\partial f|(z)\delta \leq L\delta^p + M(\delta - \mathsf{d}(z,\mathcal{D}_f))_+, \quad \forall \delta \geq 0.$$

The case of $M=0$ corresponds to smooth losses. By replacing Assumptions 1 and 2 with Assumption 5 and using the substitutions:

$$\|\tilde{z}-z\| \mapsto \mathsf{d}(\tilde{z},z), \quad \|\nabla f(z)\|_* \mapsto |\partial f|(z), \quad \sup_{\tilde{z}\neq z} \frac{(f(\tilde{z})-f(z))_+}{\|\tilde{z}-z\|} \mapsto \mathfrak{l}_f(z),$$

Theorems 1 and 2 remain to hold; see Appendix B for a complete statement of these results. In Section 4, we illustrate the results on general metric spaces for manifold regularization (Example 10) and intensity estimation of point processes (Example 11).

Remark 1 (Comparison among Wasserstein orders $p$). Comparing Theorems 1 and 2, we observe that as the Wasserstein order $p$ decreases, stronger assumptions are needed to obtain the asymptotic equivalence between the Wasserstein DRO and variation regularization, which sheds light on the modeling choice of Wasserstein order $p$. Specifically, when $p=\infty$, only local assumptions on the continuity and jump are needed (Assumption 5 (I)); when $p \in [1,\infty)$, global growth condition is required (Assumptions 5 (II)(III)), and $p \geq 2$, $p < 2$ have different orders of remainder $O(\rho_n^{p\wedge 2})$; when $p=1$, the lower and upper bound in Theorem 2 may not match but only a sandwich inequality is available. As can be seen from the proof, the worst-case distribution has a qualitative difference between $p<2$ and $p \geq 2$. For $p \geq 2$, the largest distance of perturbation is bounded for all empirical points with high probability; whereas for $p<2$, the worst-case distribution tends to perturb the empirical points with a large distance – the most extreme case being $p=1$; see the comment after Corollary 2.

# 4. Applications

In this section, we instantiate our results using various examples. For each example, we discuss why the assumptions for the corresponding result to hold and provide more detailed verification of the assumptions in Appendix EC.3.

## 4.1. Multi-item Newsvendor

We start with the classicial newsvendor problem with piecewise linear objective that makes use of Theoerem 1.

Example 5 (Mult-item Newsvendor). Consider a newsvendor problem in which the decision maker needs to decide the ordering quantities $\theta \in \mathbb{R}^d$ for $d$ products, before their random demands $z$ are realized. Let $h=(h_1,\ldots,h_d)$ and $b=(b_1,\ldots,b_d)$ be respectively the holding cost vector and backorder cost vector. The overall cost is given by

$$f_\theta(z) = \sum_{j=1}^d h_j(\theta_j - z_j)_+ + b_j(z_j - \theta_j)_+.$$

Suppose $\mathcal{Z} \subset (\mathbb{R}^d_+, \|\cdot\|_2)$ and $\Theta \subset \{\theta \in \mathbb{R}^d : \|\theta\|_\infty \le B\}$. Assume each marginal distribution of $\mathbb{P}^{z_j}_{\text{true}}$ has continuous density on $\mathbb{R}$ bounded by $\mu$.

Let us verify the assumptions required by Theorem 1. First, $f_\theta$ has $2^d$ pieces determined by the sign of $z_j - \theta_j$, $j = 1, \ldots, d$ and each piece is linear, thus Assumption 1 is satisfied with $H = 0$. Second, Assumption 2 is satisfied with $M = \sum_{j=1}^d |h_j| + |b_j|$ and $L = 0$. Third, $\mathsf{d}(z, \mathcal{D}_{f_\theta}) = \min_{1\le j\le d} |z_j - \theta_j|$, thereby Assumption 3 holds since

$$\frac{1}{\delta}\mathbb{P}_{\text{true}}\Big\{\min_{1\le j\le d}|z_j - \theta_j| < \delta\Big\} \le \frac{1}{\delta}\sum_{j=1}^d \mathbb{P}^{z_j}_{\text{true}}\{|z_j - \theta_j| < \delta\} \le d\mu.$$

Fourth, Assumption 4 is verified in Lemma EC.15 in Appendix EC.3.1.

Let $p \in (1, \infty)$ and $t > 0$. Using Theorem 1 and Lemma EC.16 in Appendix EC.3.1, we have with probability at least $1 - e^{-t}$, for all $\theta \in \Theta$,

$$\begin{aligned}\Big|\mathcal{R}_{\mathbb{P}_n,p}(\rho_n; f_\theta) - \rho_n\Big(\mathbb{E}_{\mathbb{P}_n}\Big[\sum_{j=1}^d |h_j|^q 1\{z_j < \theta_j\} + |b_j|^q 1\{\theta_j > z_j\}\Big]\Big)^{\frac{1}{q}}\Big| \\ \le C\rho_n^{2\wedge p} + d\mu\rho_n\sqrt{\frac{t}{2n}} + \frac{C_2\rho_n}{\sqrt{n}}\big(\sqrt{d\log(B/\rho_n)} + \sqrt{d\pi}\big).\end{aligned} \quad \clubsuit$$

We remark that, in this example, the remainder is $\tilde{O}(d/n)$ for $p \ge 2$ when $\rho_n = O(1/\sqrt{n})$. Also note that the objective function has $2^d$ pieces while the remainder is only linear in $d$.

### 4.2. Portfolio Selection

Next, we consider the classic portfolio utility maximization problem with piecewise linear utility functions. Let $z \in \mathcal{Z} \subset (\mathbb{R}^d, \|\cdot\|_2)$ be the random loss of $d$ assets and let $\Theta \subset \{\theta \in \mathbb{R}^d :\ \theta^\top 1 = 1,\ \|\theta\|_2 \le B\}$ be the set of admissible portfolio vectors, where $B > 0$. The goal is to find a portfolio that maximizes the expected utility of the portfolio, or minimizes the negative utility.

Example 6 (Portfolio Selection with Piecewise Linear Utility). Consider a piecewise linear (negative) utility function

$$f_\theta(z) = \max_{1\le k\le K} a_k\theta^\top z + b_k,$$

where $a_k \le 0$, $k = 1, \ldots, K$. Let $\mathcal{U}$ be the (finite) set of non-differentiable points of the one-dimensional piecewise linear function $u \mapsto \max_{1\le k\le K} a_k u + b_k$. Assume $A := \max_{1\le k\le K}|a_k| > 0$. We denote by $k(u)$ so that $|a_{k(u)}| = \max_{1\le j\le K}\{|a_j| : a_j u + b_j = \max_{1\le k\le K} a_k u + b_k\}$. Assume $\mathbb{P}_{\text{true}}$ is continuous with bounded density and $\zeta := \inf_{\theta\in\Theta}\mathbb{E}_{\mathbb{P}_{\text{true}}}[a^2_{k(\theta^\top z)}]^{\frac{1}{2}} > 0$.

We verify the assumptions required by Theorem 4. First, for every $\theta \in \Theta$, $f_\theta$ is linear on each piece determined by $\mathcal{U}$, hence Assumption 1 is satisfied with $H = 0$. Second, for $z_1, z_2$ belonging to adjoint pieces, we have $\|\nabla f_\theta(z_1) - \nabla f_\theta(z_2)\| \le \|\theta\|_2 \max_{1\le j<k\le K}|a_j - a_k|$. Hence Assumption 2 is satisified with $M = 2AB$ and $L = 0$. Third, to verify Assumption 3, note that for all $\theta \ne 0$, $\mathsf{d}(z, \mathcal{D}_{f_\theta}) = \min_{u\in\mathcal{U}}|\theta^\top z - u|/\|\theta\|_2$, thus

$$\mathbb{E}_{\mathbb{P}_{\text{true}}}\big[1\{0 < \mathsf{d}(z, \mathcal{D}_{f_\theta}) < \delta\}\big] = \mathbb{E}_{\mathbb{P}_{\text{true}}}\big[1\big\{\min_{u\in\mathcal{U}}|\theta^\top z - u| < \delta\|\theta\|_2\big\}\big] \le \sum_{u\in\mathcal{U}}\mathbb{E}_{\mathbb{P}_{\text{true}}}\big[1\big\{|\theta^\top z - u| < \delta\|\theta\|_2\big\}\big].$$

Define a random variable $U = \theta^\top z$ and denote by $\mu_\theta$ its density induced from $z \sim \mathbb{P}_{\text{true}}$. For any $\theta \ne 0$,

$$\lim_{\delta\downarrow 0}\frac{1}{\delta}\mathbb{P}_{\text{true}}\big\{|\theta^\top z - u| \le \delta\|\theta\|_2\big\} = \|\theta\|_2\lim_{\delta\downarrow 0}\frac{1}{\delta}\mathbb{P}_{\text{true}}\big\{|\theta^\top z - u| \le \delta\big\} = \|\theta\|_2\mu_\theta(u).$$

By our assumption on $\mathbb{P}_{\text{true}}$, $\mu_\theta$ is continuous and has a finite upper bound on the compact set $\Theta$, hence Assumption 3 holds. Fourth, at differentiable points, we have

$$w_{f_\theta}(z) = \frac{\|\nabla f_\theta(z)\|_*}{\|\,\|\nabla f_\theta\|_*\,\|_{\mathbb{P}_{\text{true}},2}} = \frac{|a_{k(\theta^\top z)}|\|\theta\|_2}{\mathbb{E}_{\mathbb{P}_{\text{true}}}[a^2_{k(\theta^\top z)}]^{\frac{1}{2}}\|\theta\|_2} \le A/\zeta.$$

Hence Assumption 4' holds with $c_1 = A/\zeta$ and $c_2 = 0$. Finally, by Lemma EC.17 in Appendix EC.3.2, Assumption 6 holds for all $n \ge n_0$ for some $n_0 \ge 1$.

Using Theorem 4 in Appendix A and Lemma EC.18 in Appendix EC.3.2, there exists $n_0, t_0, C, C_1, C_2 > 0$ such that for $n \ge n_0$, with probability at least $1 - 2e^{-t} - e^{-nt_0}$, for every $\theta \in \Theta$,

$$\left|\mathcal{R}_{\mathbb{P}_n,2}(\rho_n; f_\theta) - \rho_n\|\theta\|_2\Big(\mathbb{E}_{\mathbb{P}_n}\Big[a^2_{k(\theta^\top z)}\Big]\Big)^{\frac{1}{2}}\right| \le C\rho_n^2 + C_1\rho_n\sqrt{\frac{d}{n}} + \rho_n\sqrt{\frac{t}{2n}}. \qquad ♣$$

## 4.3. Linear Prediction

In this subsection, we consider two examples on linear prediction, covering two particular cases of $p = 1$ (Corollaries 1 and 2). Let $z = (x, y)$, where $x \in \mathcal{X} \subset (\mathbb{R}^d, \|\cdot\|)$, and $y \in \mathbb{R}$ for regression whereas for classification, $y$ belongs to a probability simplex in $\mathbb{R}^K$, $K \in \mathbb{N}_{\ge 2}$. To ease the exposition, we assume $\mathrm{d}(z, \tilde{z}) = \|x - \tilde{x}\| + \infty \cdot 1\{y \ne \tilde{y}\}$, thereby we can omit the $y$-component when we compute $\|\nabla f(z)\|_*$.

Example 7 (Lipschitz loss on a bounded domain, $p = 1$). Suppose the loss function has a form

$$f_\theta(z) := l(\theta^\top x, y) := \begin{cases} \ell(\theta^\top x - y), & \text{regression},\\ y\ell(\theta^\top x), & \text{binary classification}, \end{cases} \tag{5}$$

where $\theta \in \Theta \subset \{\theta \in \mathbb{R}^d : \|\theta\|_* \le B\}$ for some $B > 0$ and $\ell : \mathbb{R} \to \mathbb{R}$ is $L_\ell$-Lipschitz. Denote by $l'(\cdot, y)$ the derivative of $l$ with respect to its first argument. Assume $\ell$ has $\hbar_\ell$-Lipschitz gradient. Assume there exists $\eta \in (0, 1]$ such that

$$0 < \alpha := \begin{cases} \inf_{\theta\in\Theta} \mathbb{P}_{\text{true}}\big\{(x, y) : |\ell'(\theta^\top x - y)| \ge \eta L_\ell\big\}, & \text{regression},\\ \inf_{\theta\in\Theta} \mathbb{P}_{\text{true}}\big\{(x, y) : |\ell'(\theta^\top x)| \ge \eta L_\ell\big\}, & \text{binary classification}. \end{cases} \tag{6}$$

Let us verify the assumptions in Corollary 1. We have $\|\nabla f_\theta(z)\|_* = \|\theta\|_* l'(\theta^\top x, y)$, thus $\|f_\theta\|_{\text{Lip}} = \|\theta\|_* \sup_{(x,y)\in\mathcal{Z}} l'(\theta^\top x, y) \le L_\ell\|\theta\|_*$ and $f_\theta$ is $\hbar_\ell B^2$-semi-convex. Moreover, the constraints in (6) is equivalent to $\|\nabla f(z)\|_* \ge \eta\|f\|_{\text{Lip}}$, thereby $\alpha$ in Corollary 1 is well-defined. By Appendix EC.3.3, $\mathcal{N}(\epsilon; \mathcal{F}, \mathrm{d}_{\mathcal{F}}) \le \left(\frac{L_\ell \mathrm{diam}(\mathcal{X}) \vee (L_\ell + B\hbar_\ell \mathrm{diam}(\mathcal{X}))}{\epsilon}\right)^d$. Let $c < \frac{\alpha}{1-\alpha} \wedge \frac{1-\alpha}{\alpha}$. Then using Corollary 1, with probability at least

$$1 - d\exp\Big(-nH\big(c\|(\tfrac{\alpha}{1-\alpha} \wedge \tfrac{1-\alpha}{\alpha})\big) + \log\big(L_\ell\mathrm{diam}(\mathcal{X}) \vee (L_\ell + B\hbar_\ell\mathrm{diam}(\mathcal{X}))\big) + \log n\Big),$$

it holds for all $\theta \in \Theta$ that

$$\eta\rho_n\|\theta\|_* \max_{1\le i\le n} |l'(\theta^\top x_i^n, y_i^n)| - c\hbar_\ell\rho_n^2 - (1 \vee \rho_n)/n \le \mathcal{R}_{\mathbb{P}_n,1}(\rho_n; f_\theta) \le \rho_n\|\theta\|_* \max_{1\le i\le n} |l'(\theta^\top x_i^n, y_i^n)|. \qquad ♣$$

Example 8 (Lipschitz loss on an unbounded domain). Consider the loss function defined in (5). Suppose $\mathcal{X} = \mathbb{R}^d$. Assume additionally $\limsup_{|t|\to\infty} \frac{\ell(t)}{|t|} = L_\ell$. Examples of $\ell(t)$ include convex losses such as hinge loss $(1 - t)_+$, softplus (logistic) loss $\log(1 + e^t)$, as well as non-convex losses such as inverse S-shaped curve $\mathrm{sgn}(t)\log(\frac{1}{2}(1 + e^t))$. For classification, assume further that there exists $(x_0, y_0) \in \operatorname{supp}\mathbb{P}_n$ with $y_0 = 1$. Then $f_\theta$ is Lipschitz continuous with constant bounded by $L_\ell B$ and $\limsup_{\|x\|\to\infty} l(\theta^\top x, y) = L_\ell\|\theta\|_* = \|f_\theta\|_{\text{Lip}}$, thus (**L**) in Corollary 2 is satisfied, and we have

$$\mathcal{R}_{\mathbb{P}_n,1}(\rho; f_\theta) = \rho \cdot L_\ell\|\theta\|_*, \quad \forall\theta \in \Theta. \qquad ♣$$

We remark that this result relaxes the convexity assumption in the equivalence results derived in [24, 47].

### 4.4. Neural Networks

Following the previous section, we study supervised learning with nonlinear class in this subsection. In particular, we consider a two-layer network with leaky ReLU activations $\sigma(z) = z$ if $z \geq 0$ and $\sigma(z) = az$ if $z < 0$, where $a > 0$. As before, we consider a $K$-class classification. Let $z = (x, y) \in \mathcal{Z} = \mathcal{X} \times \mathcal{Y}$, where $\mathcal{X} \subset \mathbb{R}^d$ and $\mathcal{Y}$ is a subset of the probability simplex in $\mathbb{R}^K$. Suppose $\mathsf{d}(\tilde{z}, z) = \|\tilde{x} - x\|_2 + \infty 1\{\tilde{y} = y\}$.

EXAMPLE 9 (LEAKY RELU NETWORK). Let $\theta = (W_1, W_2)$ where $W_1 \in \mathbb{R}^{d_1 \times d}$ and $W_2 \in \mathbb{R}^{K \times d_1}$ are weight matrices. Define a two-layer ReLU network with cross-entropy loss

$$f_\theta(z) := \ell(W_2\sigma(W_1x), y) = -\log \frac{\sum_{k=1}^K y_k \exp(W_{2,k}\sigma(W_1x))}{\sum_{k=1}^K \exp(W_{2,k}\sigma(W_1x))},$$

where $W_{2,k}$ is the $k$-th row of $W_2$. Denote by $\sigma'(x)$ the diagonal matrix whose $j$-th diagonal equals $1\{x_j \geq 0\}$. Using the chain rule, at differentiable point we have

$$\|\nabla f_\theta(z)\|_2 = \|\nabla \ell(W_2\sigma(W_1x), y) W_2 \sigma'(W_1x) W_1\|_2,$$

where we have adopted the convention that the gradient is a row vector. Assume $\mathcal{X}$ is compact; $\|W_2\|_{op}\|W_1\|_{op} \leq 1$, where $\|\cdot\|_{op}$ denotes the matrix operator norm; and the marginal distribution of $\mathbb{P}^x_{\text{true}}$ on $\mathcal{X}$ is continuous.

Now we verify the assumptions required by Theorem 1. Assumption 1 is satisfied because of the Lipschitz continuity of $\ell$ and piecewise linearity of the ReLU activation function. Assumption 2 is satisfied due to the Lipschitz continuity of $\ell$ and $\sigma$. Since $\mathbb{P}_{\text{true}}$ is continuous, $\mathcal{D}_{f_\theta}$ is bounded and $\Theta$ is compact, the conditional density $d\mathbb{P}_{\text{true}}(\mathcal{D}_{f_\theta})$ is uniformly bounded over $\theta \in \Theta$, hence Assumption 3 is satisfied. Finally, Assumption 4 is verified in Appendix EC.3.4.

Let $t > 0$. Using Theorem 1 and Lemma EC.21 in Appendix EC.3.4, there exists $\bar{\rho}, C > 0$ such that for all $\rho < \bar{\rho}$, with probability at least $1 - e^{-t}$, for every $\theta \in \Theta$,

$$\left|\mathcal{R}_{\mathbb{P}_n,2}(\rho_n; f) - \rho_n \|\,\|\nabla f\|_*\,\|_{\mathbb{P}_n,2}\right| \leq C_1\rho_n^2 + C_2 d_1\sqrt{\frac{d}{n}} + \rho_n\sqrt{\frac{t}{2n}}. \qquad \clubsuit$$

### 4.5. Non-Euclidean Spaces

In this subsection we consider the case where $\mathcal{Z}$ is a metric space different from the Euclidean space using the theory developed in Section 3.4. For the ease of exposition, we consider smooth loss functions.

EXAMPLE 10 (MANIFOLD REGULARIZATION). Suppose $\mathcal{Z} \subset \mathbb{R}^d$ is a Riemannian manifold and $f : \mathbb{R}^d \to \mathbb{R}$ is twice differentiable with bounded Hessian. Then $|\partial f|(z) = \operatorname{Exp}(\nabla f)$, where Exp denotes the exponential map [16]. For example, when $\mathcal{Z}$ is the unit sphere $\{z \in \mathbb{R}^d : \|z\|_2 = 1\}$, then $\|\nabla f\|_*(z) = \operatorname{Exp}(\nabla f) = \|(I_d - zz^\top)\nabla f(z)\|_2$, where $I_d$ denotes the $d$-dimensional identity matrix. By Theorem 1 in Appendix B, there exists $a, C > 0$ such that for all $\rho_0 < a$, $n \in \mathbb{N}_{\geq 1}$ and $f \in \mathcal{F}$,

$$\left|\mathcal{R}_{\mathbb{P}_n,2}(\rho_n; f) - \rho_n \mathcal{V}_{\mathbb{P}_n,2}(f)\right| \leq C\rho_n^2,$$

where $\mathcal{V}_{\mathbb{P}_n,2}(f) = \|\,\|\operatorname{Exp}(\nabla f)\|_2\,\|_{\mathbb{P}_n,2}$. This establishes a connection between Wasserstein DRO and Laplacian regularization in manifold optimization [7]. $\clubsuit$

As another example, we consider the case where the distance $\mathsf{d}(\tilde{z}, z)$ is defined through another Wasserstein distance, in which each sample point $z$ is viewed as a measure. This setup occurs in various applications. For instance, let $\mathcal{Z}$ be the space of inhomogeneous Poisson processes on $\Xi = [0, T]$. Then each $z \in \mathcal{Z}$ can be viewed as a distribution of sample paths on $\Xi$. Each sample path is identified with a counting measure on $\Xi$, and the distance $\mathsf{d}(\tilde{z}, z)$ between two sample paths $\tilde{z}$ and $z$ is measured by some Wasserstein distance between counting measures on $\Xi$. This is called *nested distance* in [26, Section 4.2]. As another example, let $\mathcal{Z}$ be the space of black-and-white images with fixed resolution $r\times r$. Then each image $z\in\mathcal{Z}$ can be viewed as a two-dimensional histogram on the space of pixels $\Xi=\{1,\dots,r\}^2$, with each pixel representing a bin. The distance $\mathsf{d}(\tilde z,z)$ between two images $\tilde z$ and $z$ is measured by some Wasserstein distance between 2-dimensional histograms on $\Xi$. This is called *Wasserstein of Wasserstein loss* in [22].

More formally, suppose each element $z\in\mathcal{Z}$ itself is a Borel measure on a metric space $(\Xi,\mathsf{d}_\Xi)$. We define the metric $\mathsf{d}$ on $\mathcal{Z}$ as a 2-Wasserstein metric

$$\mathsf{d}(\tilde z,z)=W_\Xi(\tilde z,z):=\inf_{\gamma\in\Gamma(\tilde z,z)}\|\mathsf{d}_\Xi\|_{\gamma,2},$$

where $\Gamma(\tilde z,z)$ represents the set of Borel measures on $\Xi^2$ with marginal measures $\tilde z$ and $z$. Then the 2-Wasserstein distance between two probability distributions $\mathbb{P},\mathbb{Q}$ on $\mathcal{Z}$ becomes

$$W_2(\mathbb{P},\mathbb{Q})=\inf_{\pi\in\Gamma(\mathbb{P},\mathbb{Q})}\Big(\mathbb{E}_{(\tilde z,z)\sim\pi}\big[W_\Xi(\tilde z,z)\big]\Big)^{\frac12}.$$

Example 11 (Intensity Estimation for Point Processes). Let $\Xi\subset(\mathbb{R}^d,\|\cdot\|)$. Consider the problem of estimating the intensity function $f:\Xi\to\mathbb{R}$ of a point process, namely, $f$ satisfies $f(E)=\mathbb{E}_{z\sim\mathbb{P}}[z(E)]$ for every Borel set $E\subset\Xi$. Suppose the negative log-likelihood of a sample path $z_i^n=\sum_{m=1}^{M_i}\boldsymbol{\delta}_{\xi_{i,m}}$ has the form

$$\int_\Xi f(\xi)d\xi-\sum_{m=1}^{M_i}\log f(\xi_{i,m})=\int_\Xi f(\xi)d\xi-\mathbb{E}_{\xi\sim z_i^n}\big[\log f(\xi)\big],$$

which holds, for example, the inhomogeneous Poisson process. Then the distributionally robust negative log-likelihood function is

$$\mathcal{L}_n^{\mathsf{rob}}(f;\rho_n)=\sup_{\mathbb{P}:\,W_2(\mathbb{P},\mathbb{P}_n)\le\rho_n}\left\{\int_\Xi f(\xi)d\xi-\mathbb{E}_{z\sim\mathbb{P}}\Big[\mathbb{E}_{\xi\sim z}\big[\log f(\xi)\big]\Big]\right\}.$$

Assume $\log f$ has Lipschitz gradient bounded by $\hbar>0$. Then in Appendix EC.3.5 we show that for every $f$,

$$|\mathcal{L}_n^{\mathsf{rob}}(f;\rho_n)-\mathcal{L}_n^{\mathsf{reg}}(f;\rho_n)|\le\frac{C}{n},$$

where

$$\mathcal{L}_n^{\mathsf{reg}}(f;\rho_n):=\int_\Xi f(\xi)d\xi-\frac1n\sum_{i=1}^n\mathbb{E}_{\xi\sim z_i^n}\big[\log f(\xi)\big]+\rho_n\Big(\frac1n\sum_{i=1}^n\sum_{m=1}^{M_i}\|\nabla_\xi\log f(\xi_{i,m})\|_2^2\Big)^{\frac12},$$

here $\nabla_\xi\log f(\xi_{i,m})$ is also known as the *score function* in statistics. Therefore, this example demonstrates that 2-Wasserstein DRO penalizes the norm of the score function. ♣

## 5. Generalization Guarantees for Adversarial Robust Learning

In this section we study *adversarial robust learning*, as an application of our developed theory for $\infty$-Wasserstein DRO.

Recent studies (e.g., [54, 29]) have shown that machine learning models are vulnerable to adversarial attacks. For example, by adding a small perturbation adversarially to an image, a well-trained classification model may make a wrong prediction, even when such perturbations are imperceptible to human eyes. To improve the robustness and generalization of machine learning models, one popular approach is the following adversarial robust learning framework, which considers the following *empirical adversarial risk minimization* problem

$$\min_{f\in\mathcal{F}} \left\{ \mathcal{A}_n(\rho; f) := \frac{1}{n}\sum_{i=1}^{n} \sup_{x\in\mathcal{X}:\|x-x_i^n\|\le\rho} \ell(f(x), y_i^n) \right\}, \tag{7}$$

where $\ell : \mathbb{R}\times\{1,\ldots,K\} \to [0,1]$ is a $K$-class classification loss function such as cross-entropy, $\mathcal{F}$ is the hypothesis family on $\mathcal{X}$, and $\rho > 0$ is a small real number. Note that (7) is the dual formulation (D) of $\infty$-Wasserstein DRO when $\mathbb{Q} = \mathbb{P}_n$. The *population adversarial risk minimization* corresponding to (7) is

$$\min_{f\in\mathcal{F}} \left\{ \mathcal{A}(\rho; f) := \mathbb{E}_{\mathbb{P}_{\text{true}}}\left[ \sup_{\tilde{x}\in\mathcal{X}:\|\tilde{x}-x\|\le\rho} \ell(f(x), y) \right] \right\},$$

which is the dual formulation (D) of $\infty$-Wasserstein DRO when $\mathbb{Q} = \mathbb{P}_{\text{true}}$. One of the most important questions that this minimax formulation raises is to characterize the *generalization* property of the adversarial risk, i.e., the gap between the empirical adversarial risk and the population adversarial risk [63, 3, 4].

A immediate consequence of Theorem 1 is the following.

Example 12 (adversarial robust learning and total-variation regularization). Assume $\ell$ is smooth and every $f \in \mathcal{F}$ is piecewise smooth, which is satisfied by cross-entropy loss and ReLU network. Assume $\mathbb{P}_{\text{true}}$ is a continuous distribution on a compact set $\mathcal{X}\times\{1,\ldots,K\}$. Then Assumptions 1, 2, 3 are satisfied imeditately. Thereby Theorem 1 shows that with probability at least $1-e^{-t}$, problem (7) is equivalent to an empirical total variation regularization problem

$$\min_{f\in\mathcal{F}} \left\{ \frac{1}{n}\sum_{i=1}^{n} \ell(f(x_i^n), y_i^n) + \rho\cdot\mathbb{E}_{\mathbb{P}_n}\left[\ell'(f(x), y)\|\nabla f(x)\|_*\right] \right\} + \epsilon_n,$$

where the remainder $\epsilon_n = \rho^2(C + \|H\|_{\mathbb{P}_n,1}) + 2\rho\mathbb{E}_\otimes[\mathfrak{R}_n(\mathcal{J}_\rho)] + \rho\sqrt{\frac{t}{2n}}$ with terms defined in Theorem 1 and its assumptions, and $\mathcal{J}_\rho = \left\{x \mapsto 1\{\mathsf{d}(x, \mathcal{D}_f) < \rho\} : f\in\mathcal{F}, \mathcal{D}_f \ne \varnothing\right\}$. ♣

We develop an upper bound on the generalization error $\mathcal{A}(\rho; f) - \mathcal{A}_n(\rho; f)$, whose proof is given in Appendix EC.4. Define

$$|\partial\mathcal{F}| = \{|\partial f| : f\in\mathcal{F}\},$$

recalling $|\partial f|$ is the local slope of $f$ defined in Definition 2 that generalizes the gradient norm.

Theorem 3. *Under the setting of Example 12, assume $\ell$ is $L_\ell$-Lipschitz, and each piece of $f\in\mathcal{F}$ has gradient bounded by $L > 0$. Let $t > 0$. Then there exists $\bar{\rho}, C > 0$ such that for all $\rho < \bar{\rho}$, with probability at least $1 - 3e^{-t}$, for every $f \in \mathcal{F}$,*

$$\mathcal{A}(\rho; f) - \mathcal{A}_n(\rho; f) \le 2L_\ell(\mathbb{E}_\otimes[\mathfrak{R}_n(\mathcal{F})] + \rho\mathbb{E}_\otimes[\mathfrak{R}_n(|\partial\mathcal{F}|)] + \rho\mathbb{E}_\otimes[\mathfrak{R}_n(\mathcal{J}_\rho)]) + (2 + (L+1)L_\ell\rho)\sqrt{\frac{t}{2n}} + L_\ell C\rho^2.$$

Theorem 3 unveils that, apart from $\mathfrak{R}_n(\mathcal{F})$ appeared also in the empirical risk minimization, the Rademacher complexity of the local slope $\mathfrak{R}_n(|\partial\mathcal{F}|)$ plays a crucial role in controlling the generalization error of adversarial robust learning. Typically, $\rho$ is very small and therefore the last $O(\rho^2)$ term in the bound is likely to be negligible. When $\rho = 0$, our bound reduces to the usual generalization bound for empirical risk minimization. When $\mathcal{F}$ is a family of smooth losses, the bound in Theorem 3 becomes

$$\mathcal{A}(\rho; f) - \mathcal{A}_n(\rho; f) \le 2L_\ell(\mathbb{E}_\otimes[\mathfrak{R}_n(\mathcal{F})] + \rho\mathbb{E}_\otimes[\mathfrak{R}_n(|\partial\mathcal{F}|)]) + (2 + LL_\ell\rho)\sqrt{\frac{t}{2n}} + L_\ell C\rho^2.$$

When $\mathcal{F}$ is a family of linear losses $\mathcal{F}=\{f_\theta=\theta^\top x:\ \theta\in\Theta\}$, the bound becomes

$$\mathcal{A}(\rho;f_\theta)-\mathcal{A}_n(\rho;f_\theta)\le 2L_\ell(\mathbb{E}_\otimes[\mathfrak{R}_n(\mathcal{F})]+\rho\mathbb{E}_\otimes[\mathfrak{R}_n(\{\|\theta\|_*:\ \theta\in\Theta\})])+(2+LL_\ell\rho)\sqrt{\frac{t}{2n}},$$

which leads to the bounds developed in [63, 4], in which case the local slope is precisely $\|\theta\|_*$. The $\mathfrak{R}_n(|\partial\mathcal{F}|)$ factor appears to be new in the literature, but should make intuitive sense. Indeed, if the complexity of the variation is small, the model is more robust to the adversarial perturbations and thus generalizes better.

## 6. Conclusion

Regularization is at the core of many learning and decision-making tasks in the world of big data. In this paper, we introduce a new family of regularization schemes, termed as variation regularization, and develop a framework connecting Wasserstein DRO and variation regularization. The general theory fills the gap between the empirical success of Wasserstein DRO and the theoretical understanding of its regularization effect, and some of our results deepen our understanding of ad hoc regularization approaches used in practice. We exemplify our results under a variety of contexts in operations research and machine learning, which not only solidifies existing regularization techniques but also inspires new regularizers. Our theory also provides new aspects in understanding the generalization behavior of adversarial robust learning.

### Appendix A: A Variant of Theorem 1

In this section, we provide an alternative to Theorem 1(II) which does not require the lower bound $c_3$ in Assumption 4.

ASSUMPTION 4'. *For $z\in\mathcal{Z}\setminus\mathcal{D}_f$, define $w_f(z):=\frac{\|\nabla f(z)\|_*}{\|\,\|\nabla f\|_*\|_{\mathbb{P}_{\text{true}},q}}$. Assume there exists $c_1,c_2>0$ such that for all $f\in\mathcal{F}$ with $\mathcal{D}_f\neq\emptyset$ and all $z\in\mathcal{Z}\setminus\mathcal{D}_f$,*

$$w_f(z)\le c_1+c_2\mathsf{d}(z,\mathcal{D}_f)^{p-1}.$$

We impose the following condition, which is called the *lower isometry property* in the literature (e.g., Liang et al. [37]).

ASSUMPTION 6 **(Lower isometry property)**. *There exists $a_0,\eta,t_0,n_0>0$, such that for every $n\ge n_0$, with probability at least $1-2\exp(-nt_0)$, for every $f\in\mathcal{F}$ satisfying $\|\,|\partial f|\,\|_{\mathbb{P}_{\text{true}},q}>a_0/\sqrt{n}$, it holds that*

$$\|\,|\partial f|\,\|_{\mathbb{P}_n,q}>\eta\|\,|\partial f|\,\|_{\mathbb{P}_{\text{true}},q}.$$

Namely, we impose a positive lower bound on the ratio between the empirical variation and the true variation for loss functions whose variation is larger than $O(1/\sqrt{n})$. In Appendix EC.2.4.1, we will provide a sufficient condition ensuring Assumption 6 and prove the following theorem.

THEOREM 4. *Let $p\in(1,\infty)$. Assume Assumptions 4', 5, 6 are in force. Let $\rho_n=\rho_0/\sqrt{n}$ for some $\rho_0>0$ and let $t>0$. Then there exists $\bar\rho,C_1,C_2>0$ such that for all $\rho_0<\bar\rho$ and $n\ge n_0$, with probability at least $1-e^{-t}-e^{-nt_0}$, for every $f\in\mathcal{F}$,*

$$\left|\mathcal{R}_{\mathbb{P}_n,p}(\rho_n;f)-\rho_n\mathcal{V}_{\mathbb{P}_n,q}(f)\right|\le\rho_n^{2\wedge p}(\|H\|_{\mathbb{P}_n,\frac{p}{p-2}}1\{p>2\}+C_1)+M\mathbb{E}_{\mathbb{P}_n}\left[\left(C_2\rho_n-\mathsf{d}(z,\mathcal{D}_f)\right)_+\right].$$

This result is illustrated in the context of portfolio selection (Example 6 in Section 4).

## Appendix B: Results for General Metric Space

We first restate Assumption 4 for general metric spaces.

ASSUMPTION 4. *Let $p \in (1,\infty)$. For $z \in \mathcal{Z} \setminus \mathcal{D}_f$, define $w_f(z) := \left(\frac{|\partial f|(z)}{\| |\partial f| \|_{\mathbb{P}_{\mathrm{true}},q}}\right)^{\frac{1}{p-1}}$. Assume there exists $c_1, c_2, c_3 > 0$ such that for all $f \in \mathcal{F}$ with $\mathcal{D}_f \neq \varnothing$ and all $z \in \mathcal{Z} \setminus \mathcal{D}_f$,*

$$c_3 \le w_f(z) \le c_1 + c_2 \mathsf{d}(z, \mathcal{D}_f)^{p-1}.$$

THEOREM 1.

(I) *Let $p = \infty$. Assume Assumptions 1, 2 are in force. Then there exists $\bar{\rho} > 0$ such that for all $\rho < \bar{\rho}$ and $f \in \mathcal{F}$,*

$$\left|\mathcal{R}_{\mathbb{P}_n,\infty}(\rho; f) - \rho \| |\partial f| \|_{\mathbb{P}_n,1}\right| \le \rho^2 \|H\|_{\mathbb{P}_n,1} + M \mathbb{E}_{\mathbb{P}_n}[(\rho - \mathsf{d}(z, \mathcal{D}_f))_+].$$

(II) *Let $p \in (1,\infty)$ and $\rho_n = \rho_0/\sqrt{n}$. Assume Assumptions 1, 2, 4 are in force. Then there exists $\bar{\rho}, C_1, C_2 > 0$ such that for all $\rho_0 < \bar{\rho}$ and $f \in \mathcal{F}$,*

$$\left|\mathcal{R}_{\mathbb{P}_n,p}(\rho_n; f) - \rho_n \| |\partial f| \|_{\mathbb{P}_n,q}\right| \le \rho_n^{2\wedge p}(\|H\|_{\mathbb{P}_n,\frac{p}{p-2}} 1\{p > 2\} + C_1) + M \mathbb{E}_{\mathbb{P}_n}\left[\left(C_2 \rho_n - \mathsf{d}(z, \mathcal{D}_f)\right)_+\right].$$

Theorem 1(III) remains the same format.

THEOREM 2. *Let $p = 1$. Assume every $f \in \mathcal{F}$ is Lipschitz continuous. Assume further that there exists $\varepsilon > 0$, $\delta, \eta \in (0,1]$ such that for every $f \in \mathcal{F}$, there exists $\mathcal{Z}_f \subset \mathcal{Z}$ and $\mathfrak{T}_f : \mathcal{Z}_f \to \mathcal{Z}$ such that with probability at least $1 - \delta_n$,*

$$\begin{aligned} &f(\mathfrak{T}_f(z)) - f(z) \ge \eta(\|f\|_{\mathrm{Lip}} - \varepsilon)\mathsf{d}(\mathfrak{T}_f(z), z), \quad \forall z \in \mathcal{Z}_f, \\ &\mathbb{E}_{\mathbb{P}_n}\left[\mathsf{d}(\mathfrak{T}_f(z), z) 1\{z \in \mathcal{Z}_f\}\right] > 0. \end{aligned} \tag{T}$$

*Suppose $\rho_n \le \min_{f\in\mathcal{F}} \mathbb{E}_{\mathbb{P}_n}[\mathsf{d}(\mathfrak{T}_f(z), z)1\{z \in \mathcal{Z}_f\}]$. Then with probability at least $1 - \mathcal{N}(\frac{1}{n}; \mathcal{F}, \mathsf{d}_{\mathcal{F}}) \cdot \delta_n$,*

$$\eta \rho_n \mathcal{V}_{\mathbb{P}_n,q}(f) - \rho_n \varepsilon - (1 \vee \rho_n)/n \le \mathcal{R}_{\mathbb{P}_n,p}(\rho_n; f) \le \rho_n \mathcal{V}_{\mathbb{P}_n,q}(f).$$

# Proofs of Statements

## Appendix EC.1: Proofs for Section 2

Let $v_P$ and $v_D$ be respectively the optimal value of (P) and its dual problem (D). The following duality result is subtracted from [26].

Lemma EC.1 **(Strong duality, $p \in [1,\infty)$).** *Let $p \in [1,\infty)$ and $\mathbb{Q} \in \mathcal{P}_p(\mathcal{Z})$. Assume $f$ is upper semi-continuous and ((I)) holds. Then*

(I) $v_P \le v_D < \infty$.

(II) *Suppose either $\mathbb{Q}$ has finite support or $\mathcal{Z}$ is Polish. Then $v_P = v_D < \infty$, and the dual minimizer exists.*

We prove the following duality result for $\infty$-Wasserstein DRO below which suffices our need.

Lemma EC.2 **(Strong duality, $p=\infty$).** *Let $p=\infty$. Then $v_P \le v_D < \infty$. Suppose further that $\mathbb{Q}$ has finite support. Then $v_P = v_D$.*

*Proof.* Let $\mathbb{P}$ be such that $W_\infty(\mathbb{P},\mathbb{Q}) \le \rho$. Since $\mathbb{Q}$ has finite support, there exists a joint distribution $(\tilde z, z) \sim \pi$ with marginal distributions $\tilde z \sim \mathbb{P}$ and $z \sim \mathbb{Q}$ satisfying $\|\mathrm{d}\|_{\pi,\infty} \le W_\infty(\mathbb{P},\mathbb{Q})$. Denote by $\pi_z$ the conditional distribution of $\tilde z$ given $z$. We have that

$$\mathbb{E}_{\mathbb{P}}[f] = \mathbb{E}_{(\tilde z,z)\sim\pi}[f(\tilde z)] = \mathbb{E}_{z\sim\mathbb{Q}}\Big[\mathbb{E}_{\tilde z\sim\pi_z}[f(\tilde z)|z]\Big] \le \mathbb{E}_{z\sim\mathbb{Q}}\Big[\sup_{\tilde z}\{f(\tilde z) : \mathrm{d}(\tilde z, z) \le \rho\}\Big].$$

Taking the supremum over $\mathbb{P}$, we obtain the first part.

For the second part, suppose $\mathbb{Q} = \sum_{j=1}^m q_j \boldsymbol{\delta}_{z_j}$. Restricting on distributions of the form $\sum_{j=1}^m q_j \boldsymbol{\delta}_{\tilde z_j}$, we have that

$$v_P \ge \sup_{\{\tilde z_j\}_j}\left\{\sum_{j=1}^m q_j \boldsymbol{\delta}_{\tilde z_j} : \mathrm{d}(\tilde z_j, z_j) \le \rho\right\} = v_D.$$

Hence the proof is completed. □

Lemma EC.3. *Let $p \in (1,\infty)$. Assume $f$ is upper semi-continuous and ((I)) holds. For $\tau > 0$, define*

$$f_\tau(z) := \sup_{\tilde z \in \mathcal{Z}}\left\{f(\tilde z) - \frac{1}{p\tau^{p-1}}\mathrm{d}(\tilde z, z)^p\right\}. \tag{EC.1}$$

(I) *For any $\epsilon > 0$, it holds that*

$$\lim_{\tau\downarrow 0}\sup_{\tilde z\in\mathcal{Z}}\left\{\mathrm{d}(\tilde z, z) : f(\tilde z) - \frac{1}{p\tau^{p-1}}\mathrm{d}(\tilde z, z)^p \ge f_\tau(z) - \epsilon\right\} = 0.$$

*If the set of maximizers*

$$\mathcal{Z}_o(\tau, z) := \arg\max_{\tilde z\in\mathcal{Z}}\left\{f(\tilde z) - \frac{1}{p\tau^{p-1}}\mathrm{d}(\tilde z, z)^p\right\}$$

*is non-empty, then*

$$\lim_{\tau\downarrow 0}\ \sup_{z_\tau\in\mathcal{Z}_o(\tau,z)} \mathrm{d}(z_\tau, z) = 0.$$

(II) *For all $z \in \mathcal{Z}$,*

$$\lim_{\tau\downarrow 0} f_\tau(z) = f(z).$$

(III) *For all* $z \in \mathcal{Z}$*,*

$$\lim_{\tau \downarrow 0} \frac{f_\tau(z) - f(z)}{\tau} \le \frac{1}{q} |\partial f|(z)^q.$$

*Proof.* To prove (I), from Assumption ((I)) we have that there exists $M, L > 0$ such that for all $\tilde{z}, z \in \mathcal{Z}$,

$$f(\tilde{z}) \le M + L \mathsf{d}(\tilde{z}, z)^p.$$

Therefore, for all $\tilde{z}, z \in \mathcal{Z}$,

$$f(\tilde{z}) - f(z) \le M - f(z) + L\mathsf{d}(\tilde{z}, z_0)^p \le M - f(z) + L2^{p-1}(\mathsf{d}(\tilde{z}, z)^p + \mathsf{d}(z, z_0)^p) =: \tilde{M} + \tilde{L}\mathsf{d}(\tilde{z}, z)^p,$$

where we have used the elementary inequality $(a+b)^p \le 2^{p-1}(a^p + b^p)$ for $a, b \ge 0$. It follows that

$$f_\tau(z) - f(z) \le \sup_{\tilde{z}} \left\{ \tilde{M} - (\frac{1}{p\tau^{p-1}} - \tilde{L})\mathsf{d}(\tilde{z}, z)^p - \epsilon \right\}.$$

Since $f_\tau - f \ge 0$, the inequality above implies that for any $\tau < (\frac{1}{p\tilde{L}})^{\frac{1}{p-1}}$, any $\tilde{z}$ satisfying

$$\mathsf{d}(\tilde{z}, z)^p > \frac{\tilde{M} - \epsilon}{\frac{1}{p\tau^{p-1}} - \tilde{L}}$$

cannot be a maximizer. Letting $\tau \to 0$ yields the first part. When $\mathcal{Z}_o(\tau, z) \neq \emptyset$, setting $\epsilon = 0$ in the reasoning above gives the result.

(II) is obtained using (I) and the upper semi-continuity of $f$.

To show (III), assume that for sufficiently small $\tau$, the supremum in (EC.1) is attained at some point, denoted as $z_\tau$ (otherwise we argue by approximation). As a result, we have

$$\begin{aligned}
\limsup_{\tau \downarrow 0} \frac{f_\tau(z) - f(z)}{\tau} &= \limsup_{\tau \downarrow 0} \left\{ \frac{f(z_\tau) - f(z)}{\tau} - \frac{\mathsf{d}(z_\tau, z)^p}{p\tau^p} \right\} \\
&= \limsup_{\tau \downarrow 0} \left\{ \frac{f(z_\tau) - f(z)}{\mathsf{d}(z_\tau, z)} \frac{\mathsf{d}(z_\tau, z)}{\tau} - \frac{\mathsf{d}(z_\tau, z)^p}{p\tau^p} \right\} \\
&\le \limsup_{\tau \downarrow 0} \left\{ \frac{1}{q} \Big( \frac{f(z_\tau) - f(z)}{\mathsf{d}(z_\tau, z)} \Big)^q \right\} \\
&\le \frac{1}{q} |\partial f|^q(z),
\end{aligned}$$

where the first inequality follows from Young's inequality, and the second inequality is due to the definition of $|\partial f|(z)$. □

*Proof of Proposition 1.* Since $\mathcal{R}_{\mathbb{Q},p}(\rho; f)$ is monotone in $\rho$ and has a lower bound zero, the limit exists due to monotone convergence.

To compute the limit, we first consider $p \in [1, \infty)$. By ((I)) Lemma EC.1(I),

$$\mathcal{R}_{\mathbb{Q},p}(\rho; f) \le \min_{\lambda \ge 0} \left\{ \lambda\rho^p + \mathbb{E}_{z \sim \mathbb{Q}} \Big[ \sup_{\tilde{z} \in \mathcal{Z}} \{ f(\tilde{z}) - \lambda \mathsf{d}^p(\tilde{z}, z) \} - f(z) \Big] \right\}.$$

Using the elementary inequality $(a+b)^p \le 2^{p-1}(a^p + b^p)$ and ((I)), there exists $M, L > 0$ such that

$$f(\tilde{z}) \le L(\mathsf{d}(\tilde{z}, z) + \mathsf{d}(z, z_0))^p + M \le 2^{p-1}L(\mathsf{d}(\tilde{z}, z)^p + \mathsf{d}(z, z_0)^p) + M, \ \forall \tilde{z} \in \mathcal{Z}.$$

It follows that

$$\mathbb{E}_{z\sim\mathbb{Q}}\Big[\sup_{\tilde z\in\mathcal{Z}}\{f(\tilde z)-\lambda\mathsf{d}(\tilde z,z)^p\}\Big]\le\mathbb{E}_{z\sim\mathbb{Q}}\Big[\sup_{\tilde z\in\mathcal{Z}}\{2^{p-1}L\mathsf{d}(\tilde z,z)^p+2^{p-1}L\mathsf{d}(z,z_0)^p+M-\lambda\mathsf{d}(\tilde z,z)^p\}\Big]$$
$$\le\mathbb{E}_{z\sim\mathbb{Q}}\Big[2^{p-1}L\mathsf{d}(z,z_0)^p+M\Big],\quad\forall\lambda>2^{p-1}L.$$

Hence, we can apply the reverse Fatou's lemma and obtain

$$\lim_{\lambda\uparrow\infty}\mathbb{E}_{z\sim\mathbb{Q}}\Big[\sup_{\tilde z\in\mathcal{Z}}\{f(\tilde z)-\lambda\mathsf{d}(\tilde z,z)^p\}\Big]\le\mathbb{E}_{z\sim\mathbb{Q}}\Big[\limsup_{\lambda\uparrow\infty}\sup_{\tilde z\in\mathcal{Z}}\{f(\tilde z)-\lambda\mathsf{d}(\tilde z,z)^p\}\Big]=\mathbb{E}_{z\sim\mathbb{Q}}[f(z)],$$

where the equality is due to Lemma EC.3(II). Let $\{\rho_n\}_{n=1}^\infty$ be any sequence of positive real numbers approaching to zero, and let $\{\lambda_n\}_{n=1}^\infty$ be a sequence of positive real numbers approaching to infinity and satisfying $\lambda_n\rho_n^p\to 0$ as $n\to\infty$. It follows that

$$\lim_{\rho\downarrow 0}\mathcal{R}_{\mathbb{Q},p}(\rho;f)=\lim_{n\to\infty}\mathcal{R}_{\mathbb{Q},p}(\rho_n;f)\le\lim_{n\to\infty}\lambda_n\rho_n^p+\mathbb{E}_{z\sim\mathbb{Q}}\Big[\sup_{\tilde z\in\mathcal{Z}}\{f(\tilde z)-\lambda_n\mathsf{d}(\tilde z,z)^p\}\Big]-\mathbb{E}_{\mathbb{Q}}[f]=0.$$

Next, consider $p=\infty$. By Lemma EC.2,

$$\lim_{\rho\to 0}\mathcal{R}_{\mathbb{P}_n,\infty}(\rho;f)\le\lim_{\rho\to 0}\mathbb{E}_{\mathbb{P}_n}\Big[\sup_{\tilde z\in\mathcal{Z}}\{f(\tilde z)-f(z):\ \mathsf{d}(\tilde z,z)\le\rho\}\Big]\le\mathbb{E}_{\mathbb{P}_n}\Big[\lim_{\rho\to 0}\sup_{\tilde z\in\mathcal{Z}}\{f(\tilde z)-f(z):\ \mathsf{d}(\tilde z,z)\le\rho\}\Big]=0,$$

where the last inequality follows from the upper semi-continuity on $f$. □

## Appendix EC.2: Proofs for Section 3

*In the sequel, we develop our results for general metric spaces (Appendix B).*

### EC.2.1. Auxiliary Results

We prepare several results that will be frequently used. Recall $G_f(\delta,z)$ is defined in (4).

LEMMA EC.4. *Let $\lambda,\delta,\rho\ge 0$. For all $z\in\mathcal{Z}$,*

$$\sup_{\tilde z\in\mathcal{Z}}\{f(\tilde z)-f(z)-\lambda\mathsf{d}(\tilde z,z)^p\}=\sup_{\delta\ge 0}\{G_f(\delta,z)-\lambda\delta^p\}.$$

*For all $\{z_i^n\}_i\subset\mathcal{Z}$,*

$$\sup_{\{\tilde z_i^n\}_i\subset\mathcal{Z}}\left\{\frac{1}{n}\sum_{i=1}^n f(\tilde z_i^n)-f(z_i^n):\ \frac{1}{n}\sum_{i=1}^n\mathsf{d}(\tilde z_i^n,z_i^n)^p\le\rho^p\right\}=\sup_{\delta_i\ge 0,\,1\le i\le n}\left\{\frac{1}{n}\sum_{i=1}^n G_f(\delta_i,z_i^n):\ \frac{1}{n}\sum_{i=1}^n\delta_i^p\le\rho^p\right\}.$$

*Proof.* Introducing an auxiliary variable $\delta\ge 0$, we have

$$\sup_{\tilde z\in\mathcal{Z}}\{f(\tilde z)-f(z)-\lambda\mathsf{d}(\tilde z,z)^p\}=\sup_{\tilde z\in\mathcal{Z},\,\delta\ge 0}\Big\{f(\tilde z)-f(z)-\lambda\delta^p:\ \mathsf{d}(\tilde z,z)\le\delta\Big\}$$
$$=\sup_{\delta\ge 0}\Big\{\sup_{\tilde z\in\mathcal{Z}:\,\mathsf{d}(\tilde z,z)\le\delta}f(\tilde z)-f(z)-\lambda\delta^p\Big\}$$
$$=\sup_{\delta\ge 0}\{G_f(\delta,z)-\lambda\delta^p\}.$$

For the second part, introducing auxiliary variables $\delta_i = \mathsf{d}(\tilde{z}_i^n, z_i^n)$, we have

$$\begin{aligned}
&\sup_{\{\tilde{z}_i^n\}_i \subset \mathcal{Z}} \left\{ \frac{1}{n}\sum_{i=1}^n f(\tilde{z}_i^n) - f(z_i^n) : \ \frac{1}{n}\sum_{i=1}^n \mathsf{d}(\tilde{z}_i^n, z_i^n)^p \le \rho^p \right\} \\
&= \sup_{\{\tilde{z}_i^n\}_i \subset \mathcal{Z}, \{\delta_i\}_i \ge 0} \left\{ \frac{1}{n}\sum_{i=1}^n f(\tilde{z}_i^n) - f(z_i^n) : \ \delta_i \ge \mathsf{d}(\tilde{z}_i^n, z_i^n),\ 1 \le i \le n,\ \frac{1}{n}\sum_{i=1}^n \delta_i^p \le \rho^p \right\} \\
&= \sup_{\{\delta_i\}_i \ge 0} \left\{ \frac{1}{n}\sum_{i=1}^n \sup_{\delta_i \ge 0} \left\{ f(\tilde{z}_i^n) - f(z_i^n) : \ \delta_i \ge \mathsf{d}(\tilde{z}_i^n, z_i^n) \right\} : \ \frac{1}{n}\sum_{i=1}^n \delta_i^p \le \rho^p \right\} \\
&= \sup_{\{\delta_i\}_i \ge 0} \left\{ \frac{1}{n}\sum_{i=1}^n G_f(\delta, z_i^n) : \ \frac{1}{n}\sum_{i=1}^n \delta_i^p \le \rho^p \right\}.
\end{aligned}$$

□

The following lemma is adapted from [49, Theorem 26.5].

Lemma EC.5. *Let $\mathcal{H}$ be a set of functions on $\mathcal{Z}$. Assume $h \in [0, M]$ for every $h \in \mathcal{H}$. Let $t > 0$. Then with probability at least $1 - e^{-t}$,*

$$\mathbb{E}_{\mathbb{P}_n}[h] \le \mathbb{E}_{\mathbb{P}_{\text{true}}}[h] + 2\mathfrak{R}_n(\mathcal{H}) + M\sqrt{\frac{2t}{n}}.$$

*Similarly, with probability at least $1 - e^{-t}$,*

$$\mathbb{E}_{\mathbb{P}_{\text{true}}}[h] \le \mathbb{E}_{\mathbb{P}_n}[h] + 2\mathfrak{R}_n(\mathcal{H}) + M\sqrt{\frac{2t}{n}}.$$

*Proof.* Using McDiarmid's inequality, with probability at least $1 - e^{-t}$, for every $f \in \mathcal{F}$,

$$\sup_{f \in \mathcal{F}} \left\{ \mathbb{E}_{\mathbb{P}_n}[h] - \mathbb{E}_{\mathbb{P}_{\text{true}}}[h] \right\} - \mathbb{E}_{\otimes}\left[ \sup_{f \in \mathcal{F}} \left\{ \mathbb{E}_{\mathbb{P}_n}[h] - \mathbb{E}_{\mathbb{P}_{\text{true}}}[h] \right\} \right] \le M\sqrt{\frac{t}{2n}}.$$

Using the standard symmetrization argument of the Rademacher complexity,

$$\mathbb{E}_{\otimes}\left[ \sup_{f \in \mathcal{F}} \left\{ \mathbb{E}_{\mathbb{P}_n}[h] - \mathbb{E}_{\mathbb{P}_{\text{true}}}[h] \right\} \right] \le 2\mathbb{E}_{\otimes}[\mathfrak{R}_n(\mathcal{H})],$$

which completes the proof. □

Rademacher complexity has the following contraction property due to Ledoux and Talagrand [34].

Lemma EC.6. *Let $\ell \circ \mathcal{H} = \{\ell \circ h : h \in \mathcal{H}\}$, where $\ell$ is $L_\ell$-Lipschitz. Then*

$$\mathbb{E}_{\otimes}[\mathfrak{R}_n(\ell \circ \mathcal{H})] \le L_\ell \mathbb{E}_{\otimes}[\mathfrak{R}_n(\mathcal{H})].$$

### EC.2.2. Proofs for Section 3.2.1

For smooth functions we have $K_f = 1$. Recall $G_f(\delta, z)$ defined in (4). Consider the following condition: there exists $\hbar, L, \delta_0 > 0$, $H \in \mathcal{L}^{\frac{p}{p-2}}(\mathbb{P}_{\text{true}})$ when $p \in (2, \infty)$ and $H \in \mathcal{L}^\infty(\mathcal{Z})$ when $p \in (1, 2]$, such that for all $f \in \mathcal{F}$ and all $z \in \mathcal{Z}$,

$$\begin{aligned}
&\text{When } p \in (1,2]: && -\hbar\delta^2 \le G_f(\delta, z) - |\partial f|(z)\delta \le L\delta^p, && \forall \delta \ge 0; \\
&\text{When } p \in (2,\infty): && G_f(\delta, z) - |\partial f|(z)\delta \le H(z)\delta^2 + L\delta^p, && \forall \delta \ge 0, \\
& && G_f(\delta, z) - |\partial f|(z)\delta \ge -H(z)\delta^2, && \forall \delta < \delta_0; \\
&\text{When } p = \infty: && \left| G_f(\delta, z) - |\partial f|(z)\delta \right| \le H(z)\delta^2, && \forall \delta < \delta_0.
\end{aligned} \tag{GS}$$

This condition is stated for general metric spaces. We first show that it is implied by the smoothness condition (**S**) and the growth condition Assumption 2(I).

Lemma EC.7. *The smoothness condition* (**S**) *and the growth condition Assumption* 2(I) *imply* (**GS**).

*Proof.* By the smoothness condition in (**S**), for every $\epsilon > 0$, there exists $\delta_0 > 0$ such that for all $f \in \mathcal{F}$, $\tilde{z}, z \in \mathcal{Z}$ with $\|\tilde{z} - z\| < \delta_0$,

$$\big|\|\nabla f(\tilde{z})\|_* - \|\nabla f(z)\|_*\big| \leq h_1(z)\|\tilde{z} - z\|, \text{ where } h_1(z) := \begin{cases} (H(z)+\epsilon), & p \in (2,\infty], \\ \|H\|_\infty, & p \in (1,2]. \end{cases}$$

and particularly, $\delta_0$ can be chosen to be $+\infty$ when $p \in (1,2]$. Using the mean value theorem, for every $\delta < \delta_0$ and $\tilde{z} \in \mathcal{Z}$ satisfying $\|\tilde{z} - z\| \leq \delta$, there exists $\bar{z}$ on the line segment connecting $z$ and $\tilde{z}$ such that

$$f(\tilde{z}) - f(z) = \langle \nabla f(\bar{z}), \tilde{z} - z\rangle \Rightarrow |f(\tilde{z}) - f(z) - \langle \nabla f(z), \tilde{z} - z\rangle| \leq h_1(z)\|\tilde{z} - z\|.$$

It follows that for all $\delta < \delta_0$ and $f \in \mathcal{F}$,

$$G_f(\delta, z) \geq \sup_{\|\tilde{z} - z\| \leq \delta} \langle \nabla f(z), \tilde{z} - z\rangle - h_1(z)\|\tilde{z} - z\|^2 \geq \|\nabla f(z)\|_*\delta - h_1(z)\delta^2,$$

which proves the lower bound for all cases, and that

$$G_f(\delta, z) \leq \sup_{\|\tilde{z} - z\| \leq \delta} \langle \nabla f(z), \tilde{z} - z\rangle + h_1(z)\|\tilde{z} - z\|^2 \leq \|\nabla f(z)\|_*\delta + h_1(z)\delta^2, \tag{EC.2}$$

which proves the upper bound for $p = \infty$.

When $p \in (2,\infty)$, using the growth condition in Assumption 2(I), there exists $M, L > 0$ such that for all $\tilde{z}, z \in \mathcal{Z}$ with $\|\tilde{z} - z\| \geq \delta_0$ and $f \in \mathcal{F}$,

$$\big|\|\nabla f(\tilde{z})\|_* - \|\nabla f(z)\|_*\big| \leq M + L\|\tilde{z} - z\|^{p-1} \leq \|\tilde{z} - z\|\frac{M}{\delta_0} + L\|\tilde{z} - z\|^{p-1}.$$

Combining this inequality with (EC.2) yields that there exists $h_2 := \frac{M}{\delta_0} \vee h_1 \in \mathcal{L}^{\frac{p}{p-2}}(\mathbb{P}_{\text{true}})$ such that for all $\tilde{z}, z \in \mathcal{Z}$ and $f \in \mathcal{F}$,

$$\big|\|\nabla f(\tilde{z})\|_* - \|\nabla f(z)\|_*\big| \leq h_2(z)\|\tilde{z} - z\| + L\|\tilde{z} - z\|^{p-1}.$$

Consequently, for all $\delta \geq 0$ and $f \in \mathcal{F}$,

$$G_f(\delta, z) \leq \sup_{\|\tilde{z} - z\| \leq \delta} \langle \nabla f(z), \tilde{z} - z\rangle + h_2(z)\|\tilde{z} - z\|^2 + L\|\tilde{z} - z\|^{p-1} \leq \|\nabla f(z)\|_*\delta + h_2(z)\delta^2 + L\delta^p.$$

Thus we have shown the upper bound for $p \in (2,\infty)$.

When $p \in (1,2]$, the growth condition in in Assumption 2(I) implies that for any $\tilde{z}, z$ with $\|\tilde{z} - z\| > \delta_0$ and $f \in \mathcal{F}$,

$$\big|\|\nabla f(\tilde{z})\|_* - \|\nabla f(z)\|_*\big| \leq \frac{M}{\delta_0^{p-1}}\|\tilde{z} - z\|^{p-1} + L\|\tilde{z} - z\|^{p-1}.$$

Combining this inequality with (EC.2) and observing that there exists $C > 0$ such that for all $\delta \leq \delta_0$, $h_1(z)\delta^2 = \|H\|_\infty\delta^2 \leq C\delta^p$, we obtain that for all $\delta \geq 0$ and $f \in \mathcal{F}$,

$$G_f(\delta, z) - \|\nabla f(z)\|_*\delta \leq (C + \tfrac{M}{\delta_0^{p-1}} + L)\delta^p,$$

which completes the proof. □

We prove Lemma 1 under the condition (**GS**) for general metric spaces.

*Proof of Lemma 1.* We separate the cases $p=\infty$, $p\in(2,\infty)$ and $p\in(1,2]$.
(I) We first consider $p=\infty$.
By Lemma EC.2 and Lemma EC.4 we have

$$\mathcal{R}_{\mathbb{P}_n,\infty}(\rho;f)-\rho\||\partial f|\|_{\mathbb{P}_n,1}=\frac{1}{n}\sum_{i=1}^n\sup_{\tilde z\in\mathcal{Z}}\left\{f(\tilde z)-f(z_i^n)-\rho|\partial f|(z_i^n):\ \mathsf{d}(\tilde z,z)\le\rho\right\}$$
$$=\sup_{\{\delta_i\}_i}\left\{\frac{1}{n}\sum_{i=1}^n G_f(\delta_i,z_i^n)-\rho|\partial f|(z_i^n):\ 0\le\delta_i\le\rho,\ 1\le i\le n\right\}.$$

Using (**GS**), for $\rho<\delta_0$,

$$\left|G_f(\delta_i,z_i^n)-\rho|\partial f|(z_i^n)\right|\le H(z_i^n)\rho^2,\quad i=1,\dots,n.$$

Thus we conclude the result.

(II) Next, we consider $p\in(2,\infty)$. We first prove the upper bound. By (**GS**), for all $\delta\ge 0$, $z\in\mathcal{Z}$ and $f\in\mathcal{F}$,

$$G_f(\delta,z)\le|\partial f|(z)\delta+H(z)\delta^2+L\delta^p.$$

Using Lemma EC.1 with auxiliary variables $\lambda_1+\lambda_2=\lambda$ and Lemma EC.4, for any $\rho\ge 0$ it holds that

$$\begin{aligned}
&\mathcal{R}_{\mathbb{P}_n,p}(\rho;f)\\
&=\min_{\lambda_1,\lambda_2\ge 0}\left\{(\lambda_1+\lambda_2)\rho^p+\mathbb{E}_{\mathbb{P}_n}\left[\sup_{\delta\ge 0}\left\{G_f(\delta,z)-(\lambda_1+\lambda_2)\delta^p\right\}\right]\right\}\\
&\le\min_{\lambda_1\ge 0}\left\{(\lambda_1+\lambda_2)\rho^p+\mathbb{E}_{\mathbb{P}_n}\left[\sup_{\delta\ge 0}\left\{|\partial f|(z)\delta+H(z)\delta^2+L\delta^p-(\lambda_1+\lambda_2)\delta^p\right\}\right]\right\}\\
&\le\min_{\lambda_1\ge 0}\left\{\lambda_1\rho^p+\mathbb{E}_{\mathbb{P}_n}\left[\sup_{\delta\ge 0}\left\{|\partial f|(z)\delta-\lambda_1\delta^p\right\}\right]\right\}+\min_{\lambda_2\ge 0}\left\{\lambda_2\rho^p+\mathbb{E}_{\mathbb{P}_n}\left[\sup_{\delta\ge 0}\left\{H(z)\delta^2-(\lambda_2-L)\delta^p\right\}\right]\right\}\\
&=\min_{\lambda_1\ge 0}\left\{\lambda_1\rho^p+\mathbb{E}_{\mathbb{P}_n}\left[\tfrac{1}{q}(\lambda_1 p)^{-\frac{q}{p}}|\partial f|^q\right]\right\}+\min_{\lambda_2\ge 0}\left\{\lambda_2\rho^p+\mathbb{E}_{\mathbb{P}_n}\left[\tfrac{p-2}{p}(\lambda_2\tfrac{p}{2})^{-\frac{2}{p-2}}H(z)^{\frac{p}{p-2}}\right]\right\}+L\rho^p\\
&=\rho\||\partial f|\|_{\mathbb{P}_n,p}+\rho^2\|H\|_{\mathbb{P}_n,\frac{p}{p-2}}+L\rho^p.
\end{aligned}$$

For any $\bar\rho>0$, there exists $C>0$ such that $L\rho^p\le C\rho^2$ forall $\rho\le\bar\rho$. Therefore for any $\rho<\bar\rho$ and $f\in\mathcal{F}$,

$$\mathcal{R}_{\mathbb{P}_n,p}(\rho;f)-\rho\||\partial f|\|_{\mathbb{P}_n,p}\le\rho^2(\|H\|_{\mathbb{P}_n,\frac{p}{p-2}}+C).$$

To prove for the lower bound, by restricting on distributions that are supported on $n$ points, we have

$$\mathcal{R}_{\mathbb{P}_n,p}(\rho_n;f)\ge\sup_{\{\tilde z_i^n\}_i\subset\mathcal{Z}}\left\{\frac{1}{n}\sum_{i=1}^n f(\tilde z_i^n)-f(z_i^n):\ \frac{1}{n}\sum_{i=1}^n\mathsf{d}(\tilde z_i^n,z_i^n)^p\le\rho_n^p\right\}$$
$$=\sup_{\delta_i\ge 0,\,1\le i\le n}\left\{\frac{1}{n}\sum_{i=1}^n G_f(\delta_i,z_i^n):\ \frac{1}{n}\sum_{i=1}^n\delta_i^p\le\rho_n^p\right\},$$

where the second inequality is due to Lemma EC.4. By (**GS**), there exists $\delta_0>0$ such that for all $\delta<\delta_0$ and $f\in\mathcal{F}$,

$$G_f(\delta,z)-|\partial f|(z)\cdot\delta\ge -H(z)\delta^2,\quad\forall z\in\mathcal{Z}.$$

Meanwhile, the constraint $\frac{1}{n}\sum_{i=1}^n\delta_i^p\le\rho_n^p$ implies that for all $1\le i\le n$,

$$\delta_i\le n^{1/p}\rho_0/\sqrt{n}=\rho_0 n^{1/p-1/2}.$$

Thus whenever $\rho_0 n^{1/p-1/2}<\delta_0$, for all $1\le i\le n$,

$$G_f(\delta_i,z_i^n)-|\partial f|(z_i^n)\cdot\delta_i\ge -H(z_i^n)\delta_i^2,\quad\forall f\in\mathcal{F}.$$

It follows that

$$\begin{aligned}\mathcal{R}_{\mathbb{P}_n,p}(\rho_n;f) &\geq \sup_{\delta_i\geq 0,\,1\leq i\leq n}\left\{\frac{1}{n}\sum_{i=1}^n|\partial f|(z_i^n)\delta_i - H(z_i^n)\delta_i^2:\ \frac{1}{n}\sum_{i=1}^n\delta_i^p\leq\rho_n^p\right\}\\ &\geq \sup_{\delta_i\geq 0,\,1\leq i\leq n}\left\{\frac{1}{n}\sum_{i=1}^n|\partial f|(z_i^n)\delta_i:\ \frac{1}{n}\sum_{i=1}^n\delta_i^p\leq\rho_n^p\right\} - \sup_{\delta_i\geq 0,\,1\leq i\leq n}\left\{\frac{1}{n}\sum_{i=1}^n H(z_i^n)\delta_i^2:\ \frac{1}{n}\sum_{i=1}^n\delta_i^p\leq\rho_n^p\right\}\\ &= \rho_n\||\partial f|\|_{\mathbb{P}_n,p} - \rho_n^2\|H\|_{\mathbb{P}_n,\frac{p}{p-2}},\end{aligned}$$

where we have used Hölder's inequality to obtain the last equality.

(III) Finally we consider $p\in(1,2]$. By (**GS**), there exists $L>0$ such that

$$G_f(\delta,z)\leq|\partial f|(z)\delta+L\delta^p.$$

Using Lemma EC.1 and Lemma EC.4, for all $\alpha\geq 0$ it holds that

$$\begin{aligned}\mathcal{R}_{\mathbb{P}_n,p}(\rho;f) &= \min_{\lambda\geq 0}\left\{\lambda\rho^p+\mathbb{E}_{\mathbb{P}_n}\left[\sup_{\delta\geq 0}\left\{G_f(\delta,z)-\lambda\delta^p\right\}\right]\right\}\\ &\leq \min_{\lambda\geq 0}\left\{\lambda\rho^p+\mathbb{E}_{\mathbb{P}_n}\left[\sup_{\delta\geq 0}\left\{|\partial f|(z)\delta-(\lambda-L)\delta^p\right\}\right]\right\}\\ &= \min_{\lambda\geq 0}\left\{\lambda\rho^p+\mathbb{E}_{\mathbb{P}_n}\left[\tfrac{1}{q}(\lambda p)^{-\frac{q}{p}}|\partial f|^q\right]\right\}+L\rho^p\\ &= \rho\||\partial f|\|_{\mathbb{P}_n,p}+L\rho^p.\end{aligned}$$

On the other hand, by (**GS**) there exists $\hbar\geq 0$ such that for all $\delta>0$ and $f\in\mathcal{F}$,

$$G_f(\delta,z)\geq|\partial f|(z)\delta-\hbar\delta^2,\quad \forall z\in\mathcal{Z}.$$

It follows that for any $\rho\geq 0$ and $f\in\mathcal{F}$,

$$\begin{aligned}\mathcal{R}_{\mathbb{P}_n,p}(\rho;f) &\geq \sup_{\delta_i\geq 0,\,1\leq i\leq n}\left\{\frac{1}{n}\sum_{i=1}^n|\partial f|(z_i^n)\delta_i-\hbar\delta_i^2:\ \frac{1}{n}\sum_{i=1}^n\delta_i^p\leq\rho^p\right\}\\ &\geq \sup_{\delta_i\geq 0,\,1\leq i\leq n}\left\{\frac{1}{n}\sum_{i=1}^n|\partial f|(z_i^n)\delta_i:\ \frac{1}{n}\sum_{i=1}^n\delta_i^p\leq\rho^p\right\} - \sup_{\delta_i\geq 0,\,1\leq i\leq n}\left\{\frac{1}{n}\sum_{i=1}^n\hbar\delta_i^2:\ \frac{1}{n}\sum_{i=1}^n\delta_i^p\leq\rho^p\right\}\\ &= \rho\||\partial f|\|_{\mathbb{P}_n,p}-\hbar\rho^p,\end{aligned}$$

where the last equality follows from Hölder's inequality. □

### EC.2.3. Proofs for Section 3.2.2

We first show that Assumption 5 is implied by Assumptions 1 and 2.

Lemma EC.8. *Assumptions 1 and 2 imply Assumption 5. If, in addition, Assumption 4' holds, then $M$ in Assumption 5 can be chosen as $\Delta\||\partial f|\|_{\mathbb{P}_{\text{true}},q}$ for some $\Delta>0$.*

*Proof.* Let $z,\tilde z\in\mathcal{Z}$ with $\|\tilde z-z\|\leq\delta$. We denote by $\bar z:=z+t_f(\tilde z-z)$ first intersection point of $\mathcal{D}_f$ and the line segment connecting from $z$ to $\tilde z$, and use the convention that $t_f=1$ if there is no intersection. It follows from the mean value theorem that

$$\begin{aligned}f(\tilde z)-f(z) &= \int_0^1\langle\nabla f(z+t(\tilde z-z)),\tilde z-z\rangle dt\\ &= \int_0^{t_f}\langle\nabla f(z+t(\tilde z-z)),\tilde z-z\rangle dt+\int_{t_f}^1\langle\nabla f(z+t(\tilde z-z)),\tilde z-z\rangle dt.\end{aligned}$$

For the first integral, by Lemma EC.7 there exists $\delta_0 > 0$ and $H \in \mathcal{L}^{\frac{p}{p-2}}(\mathbb{P}_{\text{true}})$ when $p > 2$ and $H \in \mathcal{L}^{\infty}(\mathcal{Z})$ when $p \in (1,2]$ such that for every $f \in \mathcal{F}$,

$$\int_0^{t_f} \langle \nabla f(z + t(\tilde{z} - z)), \tilde{z} - z\rangle dt - t_f \langle \nabla f(z), \tilde{z} - z\rangle \le \begin{cases} L t_f^p \delta^p, \ \ \forall \delta \ge 0, & p \in (1,2], \\ H(z) t_f^2 \delta^2 + L t_f^p \delta^p, \ \ \forall \delta \ge 0, & p \in (2,\infty), \\ H(z) t_f^2 \delta^2, \ \ \forall \delta < \delta_0, & p = \infty, \end{cases}$$

and

$$\int_0^{t_f} \langle \nabla f(z + t(\tilde{z} - z)), \tilde{z} - z\rangle dt - t_f \langle \nabla f(z), \tilde{z} - z\rangle \ge \begin{cases} \hbar \delta^2, \ \ \forall \delta \ge 0, & p \in (1,2], \\ -H(z)\delta^2, \ \ \forall \delta < \delta_0, & p \in (2,\infty). \end{cases}$$

For the second integral, from the growth condition in Assumption 2 we have that

$$\left| \int_{t_f}^1 \langle \nabla f(z + t(\tilde{z} - z)), \tilde{z} - z\rangle dt - (1 - t_f)\langle \nabla f(z), \tilde{z} - z\rangle \right| \le \begin{cases} (M + L\delta^{p-1})(1 - t_f)\delta, \ \forall \delta \ge 0, & p \in (1,\infty), \\ M(1 - t_f)\delta, \ \forall \delta < \delta_0, & p = \infty. \end{cases}$$

Note that when Assumption 4' holds, the growth condition of $w_f$ suggests that $M$ in Assumption 2 can be replaced by $\Delta \| \|\nabla f\|_* \|_{\mathbb{P}_{\text{true}},q}$ for some $\Delta > 0$. Since $M(1 - t_f)\delta \le M(\delta - \mathsf{d}(z, \mathcal{D}_f))_+$, combining the two integrals $\int_0^{t_f}$ and $\int_{t_f}^1$ above and redefining $H$ in Assumption 5 completes the proof. □

Next, we derive a deterministic bound on the gap between the Wasserstein regularizer and the variation, which covers Theorem 1(I) and part of (II).

Lemma EC.9. *Assume Assumption 5 holds. When $p = \infty$, there exists $\bar{\rho} > 0$ such that for all $\rho < \bar{\rho}$ and $f \in \mathcal{F}$,*

$$\left| \mathcal{R}_{\mathbb{P}_n,\infty}(\rho; f) - \rho \| |\partial f| \|_{\mathbb{P}_n,1} \right| \le \rho_n^2 \|H\|_{\mathbb{P}_n,1} + M \mathbb{E}_{\mathbb{P}_n}[(\rho - \mathsf{d}(z, \mathcal{D}_f))_+],$$

*and when $p \in (1,\infty)$, there exists $\bar{\rho} > 0$ such that for all $\rho_0 < \bar{\rho}$ and $f \in \mathcal{F}$,*

$$\begin{aligned} &\left| \mathcal{R}_{\mathbb{P}_n,p}(\rho_n; f) - \rho_n \| |\partial f| \|_{\mathbb{P}_n,q} \right| \\ &\quad \le \rho_n^{2\wedge p} (\|H\|_{\mathbb{P}_n,\frac{p}{p-2}} 1\{p > 2\} + C) + M \mathbb{E}_{\mathbb{P}_n}\left[ \left( \rho_n \left(p \frac{|\partial f|(z) + M}{\| |\partial f| \|_{\mathbb{P}_n,q}}\right)^{\frac{1}{p-1}} - \mathsf{d}(z, \mathcal{D}_f) \right)_+ \right]. \end{aligned}$$

*Proof of Lemma EC.9.* We separate the cases $p = \infty$, $p \in [2,\infty)$ and $p \in (1,2)$.

(I) We first consider $p = \infty$. By Assumption 5 there exists $\delta_0 > 0$ and $H \in \mathcal{L}^1(\mathbb{P}_{\text{true}})$ such that for all $\delta < \delta_0$ and $z \in \mathcal{Z}$,

$$\left| G_f(\delta, z) - |\partial f|(z)\delta \right| \le H(z)\delta^2 + M(\delta - \mathsf{d}(z, \mathcal{D}_f))_+.$$

Thus, similar to the reasoning of Lemma 1, by Lemma EC.2 and Lemma EC.4 we have

$$\mathcal{R}_{\mathbb{P}_n,\infty}(\rho; f) - \rho \| |\partial f| \|_{\mathbb{P}_n,1} = \sup_{\{\delta_i\}_i} \left\{ \frac{1}{n} \sum_{i=1}^n G_f(\delta_i, z_i^n) - \rho |\partial f|(z_i^n) : \ 0 \le \delta_i \le \rho, \ 1 \le i \le n \right\},$$

and it follows that

$$\left| \mathcal{R}_{\mathbb{P}_n,\infty}(\rho; f) - \rho \| |\partial f| \|_{\mathbb{P}_n,1} \right| \le \rho^2 \|H\|_{\mathbb{P}_n,1} + M \mathbb{E}_{\mathbb{P}_n}[(\rho_n - \mathsf{d}(z, \mathcal{D}_f))_+].$$

(II) Next we consider $p \in (2,\infty)$.

For the lower bound, by Assumption 5, there exists $\delta_0, M > 0$ and $H \in \mathcal{L}^{\frac{p}{p-2}}(\mathbb{P}_{\text{true}})$ such that for all $\delta < \delta_0$ and $z \in \mathcal{Z}$,

$$G_f(\delta, z) \ge |\partial f|(z)\delta - H(z)\delta^2 - M(\delta - \mathsf{d}(z, \mathcal{D}_f))_+.$$

Similar to the reasoning in the proof of Lemma 1, whenever $\rho_0 n^{1/p-1/2} < \delta_0$ we have

$$\begin{aligned}\mathcal{R}_{\mathbb{P}_n,p}(\rho_n;f) &\geq \sup_{\delta_i\geq 0,\,1\leq i\leq n}\left\{\frac{1}{n}\sum_{i=1}^n |\partial f|(z_i^n)\delta_i - H(z_i^n)\delta_i^2 - M(\delta_i - \mathsf{d}(z_i^n,\mathcal{D}_f))_+ :\ \frac{1}{n}\sum_{i=1}^n \delta_i^p \leq \rho_n^p\right\}\\ &\geq \sup_{\delta_i\geq 0,\,1\leq i\leq n}\left\{\frac{1}{n}\sum_{i=1}^n |\partial f|(z_i^n)\delta_i - M(\delta_i - \mathsf{d}(z_i^n,\mathcal{D}_f))_+ :\ \frac{1}{n}\sum_{i=1}^n \delta_i^p \leq \rho_n^p\right\} - \rho_n^2\|H\|_{\mathbb{P}_n,\frac{p}{p-2}}.\end{aligned}$$

If $\||\partial f|\|_{\mathbb{P}_n,q} = 0$, then $\mathcal{R}_{\mathbb{P}_n,p}(\rho_n;f) \geq \rho_n\||\partial f|\|_{\mathbb{P}_n,q} = 0$ holds trivially. Otherwise, define

$$\hat{w}_f(z) := \left(\frac{|\partial f|(z)}{\||\partial f|\|_{\mathbb{P}_n,q}}\right)^{\frac{1}{p-1}}. \tag{EC.3}$$

By Hölder's inequality, it holds that $\|\hat{w}|\partial f|\|_{\mathbb{P}_n,1} = \|\hat{w}\|_{\mathbb{P}_n,p}\||\partial f|\|_{\mathbb{P}_n,q}$. Set

$$\delta_i = \rho_n \hat{w}_f(z_i^n), \quad i = 1,\dots,n.$$

Then we have

$$\frac{1}{n}\sum_{i=1}^n \delta_i^p = \rho_n^p \mathbb{E}_{\mathbb{P}_n}[\hat{w}_f(z)^p] = \rho_n^p,$$

and

$$\mathcal{R}_{\mathbb{P}_n,p}(\rho_n;f) \geq \rho_n\||\partial f|\|_{\mathbb{P}_n,q} - \rho_n^2\|H\|_{\mathbb{P}_n,\frac{p}{p-2}} - \frac{M}{n}\sum_{i=1}^n(\rho_n\hat{w}_f(z_i^n) - \mathsf{d}(z_i^n,\mathcal{D}_f))_+. \tag{EC.4}$$

Hence we have proven for the lower bound.

For the upper bound, let $\lambda_1,\lambda_2 \geq 0$, by Assumption 5, there exists $M > 0$ such that for all $z \in \mathcal{Z}$ and $f \in \mathcal{F}$,

$$\begin{aligned}&\sup_{\delta\geq 0}\left\{G_f(\delta,z) - (\lambda_1+\lambda_2)\delta^p\right\}\\ &\leq \sup_{\delta\geq 0}\left\{|\partial f|(z)\delta + H(z)\delta^2 + L\delta^p + M(\delta - \mathsf{d}(z,\mathcal{D}_f))_+ - (\lambda_1+\lambda_2)\delta^p\right\}\\ &\leq \sup_{\delta\geq 0}\left\{|\partial f|(z)\delta + M(\delta - \mathsf{d}(z,\mathcal{D}_f))_+ - \lambda_1\delta^p\right\} + \sup_{\delta\geq 0}\left\{H(z)\delta^2 + L\delta^p - \lambda_2\delta^p\right\}.\end{aligned}$$

Using Lemma EC.1 with auxiliary variables $\lambda_1+\lambda_2 = \lambda$ and Lemma EC.4, we have for all $\rho \geq 0$, it holds that

$$\begin{aligned}\mathcal{R}_{\mathbb{P}_n,p}(\rho;f) \leq \min_{\lambda_1\geq 0}&\left\{\lambda_1\rho^p + \mathbb{E}_{\mathbb{P}_n}\left[\sup_{\delta\geq 0}\left\{|\partial f|(z)\delta + M(\delta - \mathsf{d}(z,\mathcal{D}_f))_+ - \lambda_1\delta^p\right\}\right]\right\}\\ &+ \min_{\lambda_2\geq 0}\left\{\lambda_2\rho^p + \mathbb{E}_{\mathbb{P}_n}\left[\sup_{\delta\geq 0}\left\{H(z)\delta^2 - (\lambda_2 - L)\delta^p\right\}\right]\right\}.\end{aligned}$$

Similar to the reasoning of Lemma 1, the second term is upper bounded by $\rho^2\|H\|_{\mathbb{P}_n,\frac{p}{p-2}} + L\rho^p$. For the first term, observe that

$$\begin{aligned}&\sup_{\delta\geq 0}\left\{|\partial f|(z)\delta + M(\delta - \mathsf{d}(z,\mathcal{D}_f))_+ - \lambda_1\delta^p\right\}\\ &\leq \sup_{\delta\geq 0}\left\{|\partial f|(z)\delta - \lambda_1\delta^p\right\} + M\Big(\lambda_1^{-\frac{1}{p-1}}\big(|\partial f|(z) + M\big)^{\frac{1}{p-1}} - \mathsf{d}(z,\mathcal{D}_f)\Big)_+\\ &= \frac{|\partial f|(z)^{\frac{p}{p-1}}}{\lambda_1^{\frac{1}{p-1}}}\Big(\frac{1}{p^{\frac{1}{p-1}}} - \frac{1}{p^{\frac{p}{p-1}}}\Big) + M\Big(\lambda_1^{-\frac{1}{p-1}}\big(|\partial f|(z) + M\big)^{\frac{1}{p-1}} - \mathsf{d}(z,\mathcal{D}_f)\Big)_+,\end{aligned}$$

where the inequality holds because the maximizer cannot be larger than $\lambda_1^{-\frac{1}{p-1}}\big(|\partial f|(z)+M\big)^{\frac{1}{p-1}}$. As a result, picking $\lambda_1 = \frac{1}{p\rho^{p-1}}\||\partial f|\|_{\mathbb{P}_n,q}$, the first term is upper bounded by

$$\rho\||\partial f|\|_{\mathbb{P}_n,q} + M\mathbb{E}_{\mathbb{P}_n}\Big[\Big(\rho\big(p\frac{|\partial f|(z)+M}{\||\partial f|\|_{\mathbb{P}_n,q}}\big)^{\frac{1}{p-1}} - \mathsf{d}(z,\mathcal{D}_f)\Big)_+\Big].$$

Therefore we conclude the desired result.

(III) Finally, we consider $p\in(1,2]$.

For the lower bound, by Assumption 5, there exists $\hbar, M>0$ such that for all $\delta\geq 0$ and $z\in\mathcal{Z}$,

$$G_f(\delta,z)\geq |\partial f|(z)\delta - \hbar\delta^2 - M(\delta-\mathsf{d}(z,\mathcal{D}_f))_+.$$

Similar to the reasoning in the proof of Lemma 1,

$$\begin{aligned}
&\mathcal{R}_{\mathbb{P}_n,p}(\rho;f)\\
&\geq \sup_{\delta_i\geq 0,\,1\leq i\leq n}\Big\{\frac{1}{n}\sum_{i=1}^n |\partial f|(z_i^n)\delta_i - \hbar\delta_i^2 - M(\delta_i-\mathsf{d}(z,\mathcal{D}_f))_+ :\ \frac{1}{n}\sum_{i=1}^n\delta_i^p\leq\rho^p\Big\}\\
&\geq \sup_{\delta_i\geq 0,\,1\leq i\leq n}\Big\{\frac{1}{n}\sum_{i=1}^n |\partial f|(z_i^n)\delta_i - M(\delta_i-\mathsf{d}(z,\mathcal{D}_f))_+ :\ \frac{1}{n}\sum_{i=1}^n\delta_i^p\leq\rho^p\Big\} - \hbar\rho^p.
\end{aligned}$$

The rest of the proof for the lower bound is identical to that for the case $p\in(2,\infty)$.

For the upper bound, let $\lambda\geq 0$. By Assumption 5, there exists $M\geq 0$ such that for all $z\in\mathcal{Z}$,

$$\sup_{\delta\geq 0}\big\{G_f(\delta,z)-\lambda\delta^p\big\}\leq \sup_{\delta\geq 0}\big\{|\partial f|(z)\delta + L\delta^p + M(\delta-\mathsf{d}(z,\mathcal{D}_f))_+ - \lambda\delta^p\big\}.$$

Using Lemma EC.1 and Lemma EC.4, similar to the reasoning of Lemma 1, we have for all $\rho\geq 0$,

$$\mathcal{R}_{\mathbb{P}_n,p}(\rho;f)\leq \min_{\lambda\geq 0}\Big\{\lambda\rho^p + \mathbb{E}_{\mathbb{P}_n}\Big[\sup_{\delta\geq 0}\big\{|\partial f|(z)\delta + M(\delta-\mathsf{d}(z,\mathcal{D}_f))_+ - \lambda\delta^p\big\}\Big]\Big\} + L\rho^p.$$

The rest of the proof for the upper bound is identical to that for the case $p\in(2,\infty)$. □

The next lemma completes the proof of Theorem 1(II).

Lemma EC.10. *Assume Assumption 4 hold. Then there exists $C>0$ such that*

$$\mathbb{E}_{\mathbb{P}_n}\big[\big(p^{\frac{1}{p-1}}\rho s_f(w_f(z)+\Delta)-\mathsf{d}(z,\mathcal{D}_f)\big)_+\big]\leq \mathbb{E}_{\mathbb{P}_n}\big[\big(C\rho-\mathsf{d}(z,\mathcal{D}_f)\big)_+\big].$$

*Proof.* Define

$$s_f := \Big(\frac{\mathbb{E}_{\mathbb{P}_{\text{true}}}[|\partial f|^q]}{\mathbb{E}_{\mathbb{P}_n}[|\partial f|^q]}\Big)^{\frac{1}{p}}. \tag{EC.5}$$

By Assumption 4', $M$ from Lemma EC.8 can be chosen as $\Delta\||\partial f|\|_{\mathbb{P}_{\text{true}},q}$. Thus in Lemma EC.9 we have

$$\begin{aligned}
\mathbb{E}_{\mathbb{P}_n}\Big[\Big(\rho\big(p\tfrac{|\partial f|(z)+M}{\||\partial f|\|_{\mathbb{P}_n,q}}\big)^{\frac{1}{p-1}}-\mathsf{d}(z,\mathcal{D}_f)\Big)_+\Big] &= \mathbb{E}_{\mathbb{P}_n}\Big[\Big(\rho s_f\big(p\tfrac{|\partial f|(z)+\Delta\||\partial f|\|_{\mathbb{P}_{\text{true}},q}}{\||\partial f|\|_{\mathbb{P}_{\text{true}},q}}\big)^{\frac{1}{p-1}}-\mathsf{d}(z,\mathcal{D}_f)\Big)_+\Big]\\
&= \mathbb{E}_{\mathbb{P}_n}\big[\big(p^{\frac{1}{p-1}}\rho s_f(w_f(z)^{p-1}+\Delta)^{\frac{1}{p-1}}-\mathsf{d}(z,\mathcal{D}_f)\big)_+\big].
\end{aligned}$$

Using Assumption 4', $\mathsf{d}(z,\mathcal{D}_f)\leq p^{\frac{1}{p-1}}\rho\big(\frac{|\partial f|(z)}{\||\partial f|\|_{\mathbb{P}_{\text{true}},q}}+\Delta\big)^{\frac{1}{p-1}}$ implies that

$$\begin{aligned}
&\mathsf{d}(z,\mathcal{D}_f)^{p-1}\leq \rho^{p-1}(w_f(z)^{p-1}+\Delta)\leq \rho^{p-1}(c_1+\Delta+c_2\mathsf{d}(z,\mathcal{D}_f)^{p-1})\\
&\Rightarrow \mathsf{d}(z,\mathcal{D}_f)^{p-1}\leq \frac{\rho^{p-1}(c_1+\Delta)}{1-c_2\rho^{p-1}}\leq c_1+\Delta,\quad \forall\rho<1/c_2^{\frac{1}{p-1}},\\
&\Rightarrow (w_f(z)^{p-1}+\Delta)^{\frac{1}{p-1}}\leq (c_1+c_2(c_1+\Delta)+\Delta)^{\frac{1}{p-1}},\quad \forall\rho<1/c_2^{\frac{1}{p-1}}.
\end{aligned}$$

Note that this holds for all $f \in \mathcal{F}$. Moreover, by the lower bound in Assumption 4, $s_f \leq c_3^{-q/p}$. Consequently,

$$\mathbb{E}_{\mathbb{P}_n}\left[\left(p^{\frac{1}{p-1}}\rho s_f(w_f(z)+\Delta) - \mathsf{d}(z,\mathcal{D}_f)\right)_+\right] \leq \mathbb{E}_{\mathbb{P}_n}\left[\left(C\rho - \mathsf{d}(z,\mathcal{D}_f)\right)_+\right].$$

□

In the following two results, we provide probabilistic upper bounds for Theorem 1(III). Note that

$$\mathbb{E}_{\mathbb{P}_n}[(\rho - \mathsf{d}(z,\mathcal{D}_f))_+] \leq \rho\mathbb{E}_{\mathbb{P}_n}[1\{\mathsf{d}(z,\mathcal{D}_f) < \rho\}].$$

Lemma EC.11. *Assume Assumption 3 holds. Let $t > 0$. Then there exists $\bar{\rho}, C > 0$ such that for all $\rho < \bar{\rho}$, with probability at least $1 - e^{-t}$, for every $f \in \mathcal{F}$,*

$$\mathbb{E}_{\mathbb{P}_n}[1\{\mathsf{d}(z,\mathcal{D}_f) < \rho\}] \leq C\rho + 2\mathbb{E}_{\otimes}[\mathfrak{R}_n(\mathcal{I}_\rho)] + \sqrt{\frac{t}{2n}}.$$

*Moreover, there exists $\bar{\rho}, C > 0$ such that for all $\rho < \bar{\rho}$, with probability at least $1 - e^{-t}$, for every $f \in \mathcal{F}$,*

$$\mathbb{E}_{\mathbb{P}_n}[1\{\mathsf{d}(z,\mathcal{D}_f) < \rho\}] \leq C\rho + \frac{48}{\sqrt{n}}\int_0^1 \sqrt{\log\mathcal{N}(\epsilon\rho;\mathcal{E},\|\cdot\|_{\mathbb{P}_n,2})}d\epsilon + \sqrt{\frac{t}{2n}}.$$

*Proof.* By Lemma EC.5, with probability at least $1 - e^{-t}$, for every $f \in \mathcal{F}$,

$$\mathbb{E}_{\mathbb{P}_n}[1\{\mathsf{d}(z,\mathcal{D}_f) < \rho\}] \leq \mathbb{E}_{\mathbb{P}_{\text{true}}}\left[1\{\mathsf{d}(z,\mathcal{D}_f) < \rho\}\right] + 2\mathbb{E}_{\otimes}[\mathfrak{R}_n(\mathcal{I}_\rho)] + \sqrt{\frac{t}{2n}}.$$

By Assumption 3, there exists $\bar{\rho} > 0$ such that for all $\rho < \bar{\rho}$, $\mathbb{E}_{\mathbb{P}_{\text{true}}}[1\{\mathsf{d}(z,\mathcal{D}_f) < \rho\}] \leq C\rho$.

Define

$$\phi_\rho(t) = \begin{cases} 1, & t \leq 0, \\ 1 - t/\rho, & t \in [0,\rho], \\ 0, & t > 1. \end{cases}$$

Then we have

$$\mathbb{E}[1\{\mathsf{d}(z,\mathcal{D}_f) < \rho\}] \leq \mathbb{E}\left[\phi_\rho(\mathsf{d}(z,\mathcal{D}_f) - \rho)\right] \leq \mathbb{E}[1\{\mathsf{d}(z,\mathcal{D}_f) < 2\rho\}].$$

Define

$$\mathcal{H}_\rho := \left\{z \mapsto \phi_\rho\left(\mathsf{d}(z,\mathcal{D}_f) - \rho\right) : f \in \mathcal{F}\right\}.$$

By Lemma EC.5, it follows that with probability at least $1 - e^{-t}$,

$$\begin{aligned}
\mathbb{E}_{\mathbb{P}_n}\left[\phi_\rho\left(\mathsf{d}(z,\mathcal{D}_f) - \rho\right)\right] &\leq \mathbb{E}_{\mathbb{P}_{\text{true}}}\left[\phi_\rho\left(\mathsf{d}(z,\mathcal{D}_f) - \rho\right)\right] + 2\mathbb{E}_{\otimes}[\mathfrak{R}_n(\mathcal{H}_\rho)] + \sqrt{\frac{t}{2n}} \\
&\leq \mathbb{E}_{\mathbb{P}_{\text{true}}}\left[1\{\mathsf{d}(z,\mathcal{D}_f) < 2\rho\}\right] + 2\mathbb{E}_{\otimes}[\mathfrak{R}_n(\mathcal{H}_\rho)] + \sqrt{\frac{t}{2n}} \\
&\leq C\rho + 2\mathbb{E}_{\otimes}[\mathfrak{R}_n(\mathcal{H}_\rho)] + \sqrt{\frac{t}{2n}}.
\end{aligned}$$

It remains to bound $\mathfrak{R}_n(\mathcal{H}_\rho)$. Observe that $\sup_{h,\tilde{h}\in\mathcal{H}_\rho}\|h - \tilde{h}\|_{\mathbb{P}_n,2} \leq 1$. By Dudley's entropy integral (see, e.g., [56, (5.48)]), we have

$$\mathfrak{R}_n(\mathcal{H}_\rho) \leq \frac{48}{\sqrt{n}}\int_0^1 \sqrt{\log\mathcal{N}(\epsilon;\mathcal{H}_\rho,\|\cdot\|_{\mathbb{P}_n,2})}d\epsilon.$$

Observe that

$$\|\phi_\rho(\mathsf{d}(\cdot,\mathcal{D}_f)-\rho)-\phi_\rho(\mathsf{d}(\cdot,\mathcal{D}_{\tilde{f}})-\rho)\|_{\mathbb{P}_n,2}\leq\frac{1}{\rho}\|\mathsf{d}(\cdot,\mathcal{D}_f)-\mathsf{d}(\cdot,\mathcal{D}_{\tilde{f}})\|_{\mathbb{P}_n,2}.$$

Hence $\mathcal{N}(\epsilon;\mathcal{H}_\rho,\|\cdot\|_{\mathbb{P}_n,2})\leq\mathcal{N}(\epsilon\rho;\mathcal{E},\|\cdot\|_{\mathbb{P}_n,2})$ and

$$\mathfrak{R}_n(\mathcal{H}_\rho)\leq\frac{48}{\sqrt{n}}\int_0^1\sqrt{\log\mathcal{N}(\epsilon\rho;\mathcal{E},\|\cdot\|_{\mathbb{P}_n,2})}d\epsilon,$$

which completes the proof. □

Finally, Theorem 1 is proved by combining the previous lemmas.

### EC.2.4. Proofs for Appendix A

*Proof of Theorem 4.* In view of the proof of Theorem 1, it suffices to provide a probabilistic upper bound on $s_f$. Assumption 6 implies that with probability at least $1-\exp(-nt_0)$, for every $f\in\mathcal{F}$ satisfying $\||\partial f|\|_{\mathbb{P}_{\text{true}},q}>a_0/\sqrt{n}$, it holds that $s_f\leq\eta^{\frac{1}{p}}$. Hence the reasoning in Theorem 1 applies for the set of losses $\{f\in\mathcal{F}:\||\partial f|\|_{\mathbb{P}_{\text{true}},2}>a_0/\sqrt{n}\}$. For $f\in\mathcal{F}$ with $\||\partial f|\|_{\mathbb{P}_{\text{true}},q}<a_0/\sqrt{n}$, in Lemma EC.9 we have $M=\Delta\||\partial f|\|_{\mathbb{P}_{\text{true}},q}$ and

$$M\mathbb{E}_{\mathbb{P}_n}[(\rho_n-\mathsf{d}(z,\mathcal{D}_f))_+]\leq\rho_n a_0/\sqrt{n},$$

and by the reasoning in Lemma EC.10,

$$M\mathbb{E}_{\mathbb{P}_n}\Big[\Big(\rho_n\big(p\frac{|\partial f|(z)+M}{\||\partial f|\|_{\mathbb{P}_n,q}}\big)^{\frac{1}{p-1}}-\mathsf{d}(z,\mathcal{D}_f)\Big)_+\Big]\leq C\rho_n a_0/\sqrt{n}.$$

Thus the remainder is of $O(n)$ and we complete the proof. □

#### EC.2.4.1. A sufficient condition on the lower isometry property

A sufficient condition to ensure Assumption 6 is the so-called *small-ball condition* adapted from Mendelson [40].

Let $\text{star}(\mathcal{H})$ be the *star hull* of a set of functions $\mathcal{H}$

$$\text{star}(\mathcal{H}):=\{\alpha h:\ h\in\mathcal{H},\ 0\leq\alpha\leq 1\},$$

and let $\mathcal{S}_{r,q}$ be a sphere of $\mathcal{L}^q(\mathbb{P}_{\text{true}})$ with radius $r$

$$\mathcal{S}_{r,q}:=\{h\in\mathcal{L}^q(\mathbb{P}_{\text{true}}):\ \|h\|_{\mathbb{P}_{\text{true}},q}=r\}.$$

Define for $\tau>0$ that

$$\beta_{\mathcal{H},n}(\tau):=\inf_{r>0}\Big\{r:\ \mathbb{E}_\otimes\big[\mathfrak{R}_n\big(\text{star}(\mathcal{H})\cap\mathcal{S}_{r,q}\big)\big]\leq\tau r\Big\}.$$

For a function class $\mathcal{H}$, we define

$$Q_{\mathcal{H}}(\tau):=\inf_{h\in\mathcal{H}}\mathbb{P}_{\text{true}}\{|h|\geq\tau\|h\|_{\mathbb{P}_{\text{true}},q}\}.$$

Define

$$|\partial\mathcal{F}|:=\big\{|\partial f|:\ f\in\mathcal{F}\big\}.$$

ASSUMPTION EC.1 **(Small-ball condition)**. *Assume there exists $\tau>0$ such that*

$$Q_{|\partial\mathcal{F}|}(\tau)=\inf_{f\in\mathcal{F}}\mathbb{P}_{\text{true}}\big\{z:\ |\partial f|(z)\geq\tau\||\partial f|\|_{\mathbb{P}_{\text{true}},q}\big\}>0.$$

*Assume further there exist $a_0>0$ such that*

$$\beta_{|\partial\mathcal{F}|,n}(\tau Q_{|\partial\mathcal{F}|}(2\tau)/16)=\inf_{r>0}\Big\{r:\ \mathbb{E}_\otimes\big[\mathfrak{R}_n\big(\text{star}(|\partial\mathcal{F}|)\cap\mathcal{S}_{r,q}\big)\big]\leq r\tau Q_{|\partial\mathcal{F}|}(2\tau)/16\Big\}\leq a_0/\sqrt{n}.$$

The first condition basically means that for sufficiently many points, their local slopes achieves at least a fraction of the variation (that is, the weight average of local slopes). The second condition depends on the complexity of the loss function class, and is verified case-by-case.

Below let us briefly comment on how Assumption EC.1 leads to a guarantee of Assumption 6. By Lemma EC.12 below, whenever Assumption EC.1 holds, for any $r > \beta_{|\partial\mathcal{F}|,n}(\tau Q_{|\partial\mathcal{F}|}(2\tau)/16)$, with probability at least $1-2\exp(-nQ^2_{|\partial\mathcal{F}|}(2\tau)/2)$, for all $f \in \mathcal{F}$ satisfying $\| |\partial f| \|_{\mathbb{P}_{\text{true}},q} \geq r$,

$$\| |\partial f| \|^q_{\mathbb{P}_n,q} \geq \frac{\tau^q Q_{|\partial\mathcal{F}|}(2\tau)}{4} \| |\partial f| \|^q_{\mathbb{P}_{\text{true}},q}.$$

Therefore, Assumption 6 holds.

Lemma EC.12. *Assume $Q_{\mathcal{H}}(2\tau) > 0$ for some $\tau > 0$. For any $r > \beta_{\mathcal{H},n}(\tau Q_{\mathcal{H}}(2\tau)/16)$, with probability at least $1-2\exp(-nQ^2_{\mathcal{H}}(2\tau)/2)$, for all $h \in \mathcal{H}$ satisfying $\|h\|_{\mathbb{P}_{\text{true}},q} \geq r$,*

$$\text{card}\{i : \ |h(z_i^n)| \geq \tau \|h\|_{\mathbb{P}_{\text{true}},q}\} \geq \frac{nQ_{\mathcal{H}}(2\tau)}{4},$$

*whence*

$$\frac{1}{n}\sum_{i=1}^n h^q(z_i^n) \geq \frac{\tau^q Q_{\mathcal{H}}(2\tau)}{4}\|h\|^q_{\mathbb{P}_{\text{true}},q}.$$

Lemma EC.12 is a straightforward extension of Corollary 5.5 in Mendelson [40]. For the readers' convenience, we provide its proof, which is a consequence of the two lemmas below.

Lemma EC.13. *Let $\mathcal{H} \subset \mathcal{S}_q$. Assume there exists $\tau > 0$ such that $Q_{\mathcal{H}}(2\tau) > 0$. If*

$$\mathbb{E}_{\otimes,\sigma}\left[\sup_{h\in\mathcal{H}} \frac{1}{n}\sum_{i=1}^n \sigma_i h(z_i^n)\right] \leq \frac{\tau Q_{\mathcal{H}}(2\tau)}{16},$$

*then with probability at least $1-2\exp(-nQ^2_{\mathcal{H}}(2\tau)/2)$,*

$$\inf_{h\in\mathcal{H}} |\{i : h(z_i^n) \geq \tau\}| \geq nQ_{\mathcal{H}}(2\tau)/4.$$

*Proof.* Note first that $\text{card}\{i : |h(z_i^n)| \geq \tau\} = n\mathbb{E}_{\mathbb{P}_n}[1\{|h| \geq \tau\}]$. Define $\phi_\tau : \mathbb{R}_+ \to [0,1]$ as

$$\phi_\tau(t) := \begin{cases} 1, & t \geq 2\tau, \\ t/\tau - 1, & \tau \leq t < 2\tau, \\ 0, & t < \tau. \end{cases}$$

It follows that $1\{t \geq \tau\} \geq \phi_\tau(t) \geq 1\{t \geq 2\tau\}$. We have that

$$\begin{aligned} \mathbb{E}_{\mathbb{P}_n}[1\{|h| \geq \tau\}] &= \mathbb{E}_{\mathbb{P}_{\text{true}}}[1\{|h| \geq 2\tau\}] + \mathbb{E}_{\mathbb{P}_n}[1\{|h| \geq \tau\}] - \mathbb{E}_{\mathbb{P}_{\text{true}}}[1\{|h| \geq 2\tau\}] \\ &\geq \inf_{h\in\mathcal{H}} \mathbb{E}_{\mathbb{P}_{\text{true}}}[1\{|h| \geq 2\tau\}] - \sup_{h\in\mathcal{H}} |\mathbb{E}_{\mathbb{P}_n}[\phi_\tau(|h|)] - \mathbb{E}_{\mathbb{P}_{\text{true}}}[\phi_\tau(|h|)]|. \end{aligned}$$

Applying McDiarmid's inequality, Lipschitz composition property and symmetrization lemma of Rademacher complexity, we obtain that with probability at least $1-2\exp(-2t^2)$,

$$\begin{aligned} \sup_{h\in\mathcal{H}} |\mathbb{E}_{\mathbb{P}_n}[\phi_\tau(|h|)] - \mathbb{E}_{\mathbb{P}_{\text{true}}}[\phi_\tau(|h|)]| &\leq \mathbb{E}_\otimes\left[\sup_{h\in\mathcal{H}} \left|\mathbb{E}_{\mathbb{P}_n}[\phi_\tau(|h|)] - \mathbb{E}_{\mathbb{P}_{\text{true}}}[\phi_\tau(|h|)]\right|\right] + \frac{t}{\sqrt{n}} \\ &\leq \frac{4}{\tau}\mathbb{E}_{\otimes,\sigma}\left[\sup_{h\in\mathcal{H}}\left|\frac{1}{n}\sum_{i=1}^n \sigma_i h(z_i^n)\right|\right]. \end{aligned}$$

Hence, with probability at least $1-2\exp(-2t^2)$, for every $h \in \mathcal{H}$,

$$\mathbb{E}_{\mathbb{P}_n}[1\{|h| \geq \tau\}] \geq \inf_{h\in\mathcal{H}} \mathbb{P}_{\text{true}}\{|h| \geq 2\tau\} - \frac{4}{\tau}\mathbb{E}_{\otimes,\sigma}\left[\sup_{h\in\mathcal{H}}\left|\frac{1}{n}\sum_{i=1}^n \sigma_i h(z_i^n)\right|\right] - \frac{t}{\sqrt{n}}.$$

Setting $t = \sqrt{n}Q_{\mathcal{H}}(2\tau)/2$ and using the condition in the lemma yields the result. □

LEMMA EC.14. *Let $\mathcal{H}$ be star-shaped around o and assume $Q_{\mathcal{H}}(2\tau) > 0$ for some $\tau > 0$. Then for every $r > \beta_{\mathcal{H},n}(\tau Q_{\mathcal{H}}(2\tau)/16)$, with probability at least $1 - 2\exp(-NQ^2_{\mathcal{H}}(2\tau)/2)$, for every $h \in \mathcal{H}$ satisfying $\|h\|_{\mathbb{P}_{\text{true}},q} \geq r$,*

$$\text{card}\{i : |h(z_i^n)| \geq \tau\|h\|_{\mathbb{P}_{\text{true}},q}\} \geq nQ_{\mathcal{H}}(2\tau)/4.$$

*Proof.* Define $V = \{h/r : h \in \mathcal{H} \cup \mathcal{S}_{r,q}\}$. Then $Q_V(2\tau) \geq Q_{\mathcal{H}}(2\tau)$ since $rV \subset \mathcal{H}$. Since $\mathcal{H}$ is star-shaped around o, by definition of $\beta_{\mathcal{H},n}$,

$$\mathbb{E}_{\otimes,\sigma}\left[\sup_{h\in\mathcal{H}\cap\mathcal{S}_{r,q}}\left|\frac{1}{n}\sum_{i=1}^n \sigma_i h(z_i^n)\right|\right] \leq \frac{\tau Q_{\mathcal{H}}(2\tau)}{16} \leq \frac{\tau Q_V(2\tau)}{16}.$$

We have

$$\mathbb{E}_{\otimes,\sigma}\left[\sup_{v\in V}\left|\frac{1}{n}\sum_{i=1}^n \sigma_i v(z_i^n)\right|\right] = \mathbb{E}_{\otimes,\sigma}\left[\sup_{h\in\mathcal{H}\cap\mathcal{S}_{r,q}}\left|\frac{1}{n}\sum_{i=1}^n \sigma_i h(z_i^n)/r\right|\right] \leq \frac{\tau Q_{\mathcal{H}}(2\tau)}{16} r.$$

Applying Lemma EC.13 to $V$ yields that with probability at least $1 - 2\exp(-nQ^2_{\mathcal{H}}(2\tau)/2)$, for every $v \in V$,

$$|\{i : v(z_i^n) \geq \tau\}| \geq nQ_{\mathcal{H}}(2\tau)/4.$$

For any $h \in \mathcal{H}$ satisfying $\|h\|_{\mathbb{P}_{\text{true}},q} \geq r$, since $\mathcal{H}$ is star-shaped, $h/\|h\|_{\mathbb{P}_{\text{true}},q} \in V$, thus the result follows. □

### EC.2.5. Proofs for Section 3.3

*Proof of Theorem 2.* For the upper bound, Lemma EC.1 implies that $\mathcal{R}_{\mathbb{Q},1}(\rho; f)$ is concave and $\mathcal{R}_{\mathbb{Q},1}(\rho; f) = 0$. Therefore,

$$\mathcal{R}_{\mathbb{Q},1}(\rho; f) \leq \rho \lim_{\rho\to 0}\frac{\mathcal{R}_{\mathbb{Q},p}(\rho; f)}{\rho} \leq \rho\mathcal{V}_{\mathbb{Q},\infty}(f).$$

For the lower bound, let us temporarily fix $f \in \mathcal{F}$ and restrict on the event that (T) holds. Let $a_f = \mathbb{E}_{\mathbb{Q}}[\mathsf{d}(\mathfrak{T}_f(z), z)1\{z \in \mathcal{Z}_f\}]$. Let $\mathbb{Q}_1$ be the distribution obtained by transporting $\mathbb{Q}$ via the map $\mathfrak{T}_f$. Define

$$\mathbb{P} = \frac{\rho}{a_f}\mathbb{Q}_1 + (1 - \frac{\rho}{a_f})\mathbb{Q}.$$

The second condition in (T) implies that $W_1(\mathbb{P},\mathbb{Q}) \leq \frac{\rho}{a_f}W_1(\mathbb{Q}_1,\mathbb{Q}) \leq \frac{\rho}{a_f}a_f = \rho$. Moreover, the first condition in (T) implies that

$$\mathcal{R}_{\mathbb{Q},1}(\rho; f) \geq \mathbb{E}_{\mathbb{P}}[f] - \mathbb{E}_{\mathbb{Q}}[f] \geq \frac{\rho}{a_f}\mathbb{E}_{\mathbb{Q}}\left[\eta(\|f\|_{\text{Lip}} - \varepsilon)\mathsf{d}(\mathfrak{T}_f(z), z)1\{z \in \mathcal{Z}_f\}\right] \geq \rho(\eta\|f\|_{\text{Lip}} - \varepsilon).$$

To bound the probability for a family of losses, using the definition of the covering number, for any $\epsilon > 0$ there exists a finite subset $\mathcal{F}_\epsilon$ of $\mathcal{F}$ with cardinality smaller than or equal to $\mathcal{N}(\epsilon;\mathcal{F},\mathsf{d}_{\mathcal{F}})$, such that for any $f \in \mathcal{F}$ there exists $\tilde{f} \in \mathcal{F}_\epsilon$ with $\max(\|f - \tilde{f}\|_{\mathbb{P}_n}, |\|f\|_{\text{Lip}} - \|\tilde{f}\|_{\text{Lip}}|) \leq \epsilon$. Note that by Lemma EC.1,

$$\left|\mathcal{R}_{\mathbb{P}_n,1}(\rho; f) - \mathcal{R}_{\mathbb{P}_n,1}(\rho;\tilde{f})\right| \leq \|f - \tilde{f}\|_\infty, \quad \forall\rho \geq 0.$$

It follows that

$$\begin{aligned}
&\mathbb{P}_\otimes\left\{\sup_{f\in\mathcal{F}}\{\mathcal{R}_{\mathbb{P}_n,1}(\rho; f) - \eta\rho\|\mathfrak{l}_f\|_{\mathbb{P}_n,\infty} + \rho\varepsilon\} < -\epsilon(1\vee\rho)\right\}\\
&\leq \mathbb{P}_\otimes\left\{\sup_{f\in\mathcal{F}}\{\mathcal{R}_{\mathbb{P}_n,1}(\rho; f) - \eta\rho\|f\|_{\text{Lip}} + \rho\varepsilon\} < -\epsilon(1\vee\rho)\right\}\\
&= \mathbb{P}_\otimes\left\{\exists f \in \mathcal{F}, s.t.\ \mathcal{R}_{\mathbb{P}_n,1}(\rho; f) - \eta\rho\|f\|_{\text{Lip}} + \rho\varepsilon < -\epsilon(1\vee\rho)\right\}\\
&\leq \mathbb{P}_\otimes\left\{\exists\tilde{f} \in \mathcal{F}_\epsilon, s.t.\ \mathcal{R}_{\mathbb{P}_n,1}(\rho;\tilde{f}) - \eta\rho\|\tilde{f}\|_{\text{Lip}} + \rho\varepsilon < 0\right\}\\
&\leq \sum_{\tilde{f}\in\mathcal{F}_{\rho^2}}\mathbb{P}_\otimes\left\{\mathcal{R}_{\mathbb{P}_n,1}(\rho;\tilde{f}) - \eta\rho\|\tilde{f}\|_{\text{Lip}} + \rho\varepsilon < 0\right\}\\
&\leq \mathcal{N}(\epsilon(1\vee\rho),\mathcal{F},\mathbb{d})\cdot\delta.
\end{aligned}$$

Hence the proof is completed by letting $\epsilon = 1/n$. □

*Proof of Corollary 1.* Note that by definition, $|\partial f|(z) = \sup_{g\in\partial f(z)}\|g\|_*$. Using assumptions in the corollary statement, for any $\epsilon \in (0,\alpha)$, there exists $\mathcal{Z}_f \subset \mathcal{Z}$ with $\mathbb{P}_{\text{true}}(\mathcal{Z}_f) > \alpha - \epsilon =: \alpha'$ and a mapping $\mathfrak{T}_f : \mathcal{Z}_f \to \mathcal{Z}$ such that $\|\mathfrak{T}_f(z) - z\| = \delta = c\rho_n$ and

$$f(\mathfrak{T}_f(z)) - f(z) > |\partial f|(z)\delta - \hbar\delta^2, \quad \forall z \in \mathcal{Z}_f,$$

where we have used the fact that $\mathcal{Z}$ is a Banach space to ensure the existence of $\mathfrak{T}_f$. We order $z_i^n$'s in $\operatorname{supp}\mathbb{P}_n$ as $|\partial f|(z^n_{(1)}) \ge |\partial f|(z^n_{(2)}) \ge \cdots \ge |\partial f|(z^n_{(n)})$. Set $j = \lceil n/c\rceil$ and $s = n/c - j + 1$. Then

$$\frac{1}{n}\sum_{i=1}^{j-1}\|T(z^n_{(i)}) - z^n_{(i)}\| + \frac{s}{n}\|T(z^n_{(j)}) - z^n_{(j)}\| = \frac{j-1}{n}\delta + \frac{s}{n}\delta = \frac{n/c}{n}\delta = \rho_n.$$

Using the tail bound of the binomial expansion, we have

$$\mathbb{P}\Big\{\eta|\partial f|(z^n_{(j)}) < \|f\|_{\text{Lip}}\Big\} = \sum_{i=0}^{j-1}\binom{n}{i}(\alpha')^i(1-\alpha')^{n-i} \le \exp\Big(-nH\big(c\|(\tfrac{\alpha'}{1-\alpha'} \wedge \tfrac{1-\alpha'}{\alpha'})\big)\Big).$$

Thus, letting $\epsilon \to 0$ verifies (T). □

*Proof of Corollary 2.* We first show that $\|f\|_{\text{Lip}} = \mathcal{V}_{\mathbb{Q},\infty}(f)$. By definition, $\mathfrak{l}_f(z) \le \|f\|_{\text{Lip}}$ for all $z \in \mathcal{Z}$, thus by definition $\mathcal{V}_{\mathbb{Q},\infty}(f) \le \|f\|_{\text{Lip}}$. On the other hand, the definition of $\|f\|_{\text{Lip}}$ implies that for any $\epsilon > 0$ there exists $z_0, \tilde z \in \mathcal{Z}$ such that $f(\tilde z) - f(z_0) > (\|f\|_{\text{Lip}} - \epsilon)\mathrm{d}(\tilde z, z_0)$. Hence $\|\mathfrak{l}_f\|_\infty \ge \mathfrak{l}_f(z_0) > \|f\|_{\text{Lip}} - \epsilon$. Therefore $\|\mathfrak{l}_f\|_\infty = \|f\|_{\text{Lip}}$. To show $\|\mathfrak{l}_f\|_\infty = \|\mathfrak{l}_f\|_{\mathbb{Q},\infty}$, observe from [26, Lemma 6] that the condition $\lim_{\mathrm{d}(\tilde z,z_0)\to\infty}\frac{f(\tilde z)}{\mathrm{d}(\tilde z,z_0)} = \|\mathfrak{l}_f\|_\infty$ is independent of the choice of $z_0$. As a result of our assumption, $\mathfrak{l}_f(z) = \|\mathfrak{l}_f\|_\infty$ for all $z$ and thus $\|\mathfrak{l}_f\|_\infty = \|\mathfrak{l}_f\|_{\mathbb{Q},\infty}$.

To verify (T), using (L), for any $\varepsilon > 0$ and $a > 0$, there exists $\tilde z_1^n \in \mathcal{Z}$ such that $\mathrm{d}(\tilde z_1^n, z_1^n) > na$, and

$$f(\tilde z_1^n) - f(z_1^n) \ge (\|\mathfrak{l}_f\|_\infty - \varepsilon)\mathrm{d}(\tilde z_1^n, z_1^n).$$

Then (T) is satisfied by setting $\mathcal{Z}_f = \{z_1^n\}$, $\mathfrak{T}_f(z_1^n) = \tilde z_1^n$ and arbitrarily small $\varepsilon$. □

## Appendix EC.3: Proofs for Section 4

### EC.3.1. Proofs for Section 4.1

LEMMA EC.15. *Under the setting in Example 5,*

$$\frac{\||h| \vee |b|\|_q}{\||h| \wedge |b|\|_q} \le w_{f_\theta}(z) \le \frac{\||h| \vee |b|\|_q}{\||h| \wedge |b|\|_q},$$

*where* $|h| \vee |b| = (|h_1| \vee |b_1|, \ldots, |h_d| \vee |b_d|)$, $|h| \wedge |b| = (|h_1| \wedge |b_1|, \ldots, |h_d| \wedge |b_d|)$.

*Proof.* We have $\|\nabla f_\theta(z)\|_*^q = \sum_{j=1}^d |h_j|^q 1\{z_j < \theta_j\} + |b_j|^q 1\{\theta_j > z_j\}$ and

$$\|\,\|\nabla f_\theta\|_*\,\|^q_{\mathbb{P}_{\text{true}},q} = \mathbb{E}_{\mathbb{P}_{\text{true}}}\Big[\sum_{j=1}^d |h_j|^q 1\{z_j < \theta_j\} + |b_j|^q 1\{\theta_j > z_j\}\Big] \le \sum_{j=1}^d \max(|h_j|, |b_j|)^q.$$

It follows that

$$w_{f_\theta}(z)^q = \frac{\|\nabla f_\theta(z)\|_*^q}{\|\,\|\nabla f_\theta\|_*\,\|^q_{\mathbb{P}_{\text{true}},q}} \ge \frac{\sum_{j=1}^d \min(|h_j|, |b_j|)^q}{\sum_{j=1}^d \max(|h_j|, |b_j|)^q}.$$

Similarly,

$$w_{f_\theta}(z)^q \le \frac{\sum_{j=1}^d \max(|h_j|, |b_j|)^q}{\sum_{j=1}^d \min(|h_j|, |b_j|)^q}.$$

Hence the proof is completed. □

Lemma EC.16. *Under the setting in Example 5,*

$$\int_0^1 \sqrt{\log \mathcal{N}(\epsilon\rho_n; \mathcal{E}, \|\cdot\|_{\mathbb{P}_n,2})}d\epsilon \le \sqrt{d\log(B/\rho_n)} + \sqrt{d\pi}.$$

*Proof.* Since

$$|\mathsf{d}(z, \mathcal{D}_{f_{\tilde\theta}}) - \mathsf{d}(z, \mathcal{D}_{f_\theta})| \le \max_{1\le j\le d} |\tilde\theta_j - \theta_j| = \|\tilde\theta - \theta\|_\infty,$$

we have $\mathcal{N}(\epsilon; \mathcal{E}, \|\cdot\|_{\mathbb{P}_n,2}) \le (B/\epsilon)^d$. Therefore,

$$\begin{aligned}
\int_0^1 \sqrt{\log \mathcal{N}(\epsilon\rho_n; \mathcal{E}, \|\cdot\|_{\mathbb{P}_n,2})}d\epsilon &\le \int_0^1 \sqrt{d\log(\tfrac{B}{\epsilon\rho_n})}d\epsilon \\
&\le \sqrt{d\log(B/\rho_n)} + \sqrt{d}\int_0^1 \sqrt{\log(1/\epsilon)}d\epsilon \\
&\le \sqrt{d\log(B/\rho_n)} + \sqrt{d}\int_0^\infty u^{\frac{1}{2}} e^{-u} du \\
&= \sqrt{d\log(B/\rho_n)} + \sqrt{d\pi}.
\end{aligned}$$

where the second inequality follows from the fact $\sqrt{a+b} \le \sqrt{a} + \sqrt{b}$, and the last inequality is due to the Gamma function $\Gamma(1/2) = \sqrt{\pi}$. □

### EC.3.2. Proofs for Section 4.2

The following Lemma verifies Assumption 6.

Lemma EC.17. *Under the setting of Example 6, there exists $n_0(A, \zeta, m, d) > 0$ such that for all $n > n_0$, with probability at least $1 - \exp(-nQ^2_{|\partial\mathcal{F}|}(2\tau)/2)$, for all $\theta \in \Theta$,*

$$\| \|\nabla f_\theta\|_* \|_{\mathbb{P}_n,q} > \frac{\tau^2 Q_{|\partial\mathcal{F}|}(2\tau)}{4} \| \|\nabla f_\theta\|_* \|_{\mathbb{P}_{\text{true}},q}.$$

*Proof.* To verify Assumption EC.1, observe that for all $\theta \in \Theta$,

$$\frac{\mathbb{E}_{\mathbb{P}_{\text{true}}}[\|\nabla f_\theta\|_*^2]^2}{\mathbb{E}_{\mathbb{P}_{\text{true}}}[\|\nabla f_\theta\|_*^4]} = \frac{\mathbb{E}_{\mathbb{P}_{\text{true}}}\left[|a_{k(\theta^\top z)}|^2 \|\theta\|_2^2\right]^2}{\mathbb{E}_{\mathbb{P}_{\text{true}}}\left[(\|\theta\|_2 |a_{k(\theta^\top z)}|)^4\right]} \ge \frac{\zeta^4}{A^4} > 0.$$

Using Paley-Zygmund inequality, for any $\tau < 1/2$,

$$Q_{|\partial\mathcal{F}|}(2\tau) = \inf_{\theta\in\Theta} \mathbb{P}_{\text{true}}\{\|\nabla f_\theta(z)\|_* > 2\tau \| \|\nabla f_\theta\|_* \|_{\mathbb{P}_{\text{true}},2}\} \ge \inf_{\theta\in\Theta} (1 - 4\tau^2) \frac{\mathbb{E}_{\mathbb{P}_{\text{true}}}[\|\nabla f_\theta\|_*^2]^2}{\mathbb{E}_{\mathbb{P}_{\text{true}}}[\|\nabla f_\theta\|_*^4]} > 0.$$

To compute $\beta_{\mathcal{H},n}(\tau)$, observe that $\| \|\nabla f_\theta\|_* \|_{\mathbb{P}_{\text{true}},2} \le r$ implies $\|\theta\|_2 \le \frac{r}{2\zeta}$. For any $r > 0$, it holds that

$$\begin{aligned}
\text{star}(\{\|\nabla f_\theta\|_* : \theta \in \Theta\}) \cap \mathcal{S}_{r,2} &= \left\{\|\theta\|_2 |a_{k(\theta^\top z)}| : \theta \in \Theta\right\} \cap \mathcal{S}_{r,2} \\
&\subset \left\{\|\theta\|_2 |a_{k(\theta^\top z)}| : \|\theta\|_2 \le \tfrac{r}{2\zeta}\right\}.
\end{aligned}$$

Let $-\infty = u_0 < u_1 < \cdots < u_m < u_{m+1} = \infty$, where $u_j \in \mathcal{U}$, $j = 1, \ldots, m$. Thereby we have

$$\begin{aligned}
&\mathfrak{R}_n\big(\mathrm{star}(\{\|\nabla f_\theta\|_* : \theta \in \Theta\}) \cap \mathcal{S}_{r,2}\big) \\
&\leq \mathfrak{R}_n\big(\{\|\theta\|_2 |a_{k(\theta^\top z)}| : \|\theta\|_2 \leq \tfrac{r}{2\zeta}\}\big) \\
&\leq \frac{r}{2\zeta}\mathfrak{R}_n\Big(\Big\{\max_{1\leq j\leq m} |a_{k(\theta^\top z)}| 1\{u_j < \theta^\top z \leq u_{j+1}\} : \|\theta\|_2 \leq \tfrac{r}{2\zeta}\Big\}\Big) \\
&\leq \frac{r}{2\zeta}\mathfrak{R}_n\Big(\Big\{\max_{1\leq j\leq m} |a_{k(\theta^\top z)}| 1\{u_j < \theta^\top z \leq u_{j+1}\} : \|\theta\|_2 \leq \tfrac{r}{2\zeta}\Big\}\Big) \\
&\leq \frac{CAmr}{2\zeta}\sqrt{\frac{d}{n}},
\end{aligned}$$

where which the last inequality follows from Lemma EC.19. This implies $\beta_{|\partial\mathcal{F}|,n}(\tau) = 0$ for sufficiently large $n$ satisfying $\frac{CAmr}{2\zeta}\sqrt{\frac{d+1}{n}} \leq r\tau Q_{|\partial\mathcal{F}|}(2\tau)/16$. Therefore, Assumption 6 is satisfied with $t_0 = Q^2_{|\partial\mathcal{F}|}(2\tau)/2$ and $\eta = \tau^2 Q_{|\partial\mathcal{F}|}(2\tau)/4$. □

In the next lemma, we derive an upper bound on $\mathbb{E}_\otimes[\mathfrak{R}_n(\mathcal{I}_\rho)]$ in Theorem (III).

Lemma EC.18. *Under the setting of Example 6, there exists $C > 0$ such that with probability at least $1 - 2e^{-t}$, for every $\rho > 0$,*

$$\mathbb{E}_\otimes[\mathfrak{R}_n(\mathcal{I}_\rho)] \leq C\sqrt{\frac{d}{n}}.$$

*Proof.* For $\theta \neq 0$, we have $\mathsf{d}(z, \mathcal{D}_{f_\theta}) = \min_{u\in\mathcal{U}} \frac{|\theta^\top z - u|}{\|\theta\|_2}$. We have that

$$\begin{aligned}
\mathfrak{R}_n\Big(\Big\{z \mapsto 1\{\min_{u\in\mathcal{U}} |\theta^\top z - u| < \rho\|\theta\|_2\} : \theta \in \Theta\Big\}\Big) &\leq \mathfrak{R}_n\Big(\Big\{z \mapsto 1\{\min_{u\in\mathcal{U}} |\theta^\top z - u| < v\} : \theta \in \mathbb{R}^d, v \in \mathbb{R}\Big\}\Big) \\
&= \mathfrak{R}_n\Big(\Big\{z \mapsto \max_{u\in\mathcal{U}} 1\{|\theta^\top z - u| < v\} : \theta \in \mathbb{R}^d, v \in \mathbb{R}\Big\}\Big) \\
&\leq C\sqrt{\frac{d}{n}},
\end{aligned}$$

where the last inequality follows from Lemma EC.19. □

Lemma EC.19. *Let $\Theta \subset \mathbb{R}^d$, $\mathcal{U} \subset \mathbb{R}$ and $a_j \in \mathbb{R}$, $j = 1, \ldots, m$. Set $A = \max_{1\leq j\leq m} |a_j|$. Define*

$$\mathcal{H} = \Big\{z \mapsto \max_{1\leq j\leq m} |a_j| \vee |a_{j+1}| \cdot 1\{u_j < \theta^\top z \leq u_{j+1}\} : \theta \in \Theta, u_j \in \mathcal{U}\Big\}.$$

*Then there exists $C > 0$ such that*

$$\mathbb{E}_\otimes[\mathfrak{R}_n(\mathcal{H})] \leq ACm\sqrt{\frac{d}{n}}.$$

*Proof.* Let $\boldsymbol{\sigma}_i$ be i.i.d Rademacher random variables. We have

$$\begin{aligned}
\mathfrak{R}_n(\mathcal{H}) &= \frac{1}{n}\mathbb{E}_{\boldsymbol{\sigma}}\Big[\max_{\theta\in\Theta, u_j\in\mathcal{U}} \sum_{i=1}^n \boldsymbol{\sigma}_i \max_{1\leq j\leq m} |a_j| \vee |a_{j+1}| 1\{u_j < \theta^\top z_i^n \leq u_{j+1}\}\Big] \\
&\leq \frac{1}{n}\sum_{j=1}^m \mathbb{E}_{\boldsymbol{\sigma}}\Big[\max_{\theta\in\Theta, u_j\in\mathcal{U}} \sum_{i=1}^n \boldsymbol{\sigma}_i |a_j| \vee |a_{j+1}| 1\{u_j < \theta^\top z_i^n \leq u_{j+1}\}\Big] \\
&= \sum_{j=1}^m \mathfrak{R}_n\Big(\Big\{z \mapsto ||a_j| \vee |a_{j+1}| 1\{u_j < \theta^\top z \leq u_{j+1}\} : \theta \in \Theta, u_j \in \mathcal{U}\Big\}\Big).
\end{aligned}$$

By contraction inequality of Rademacher complexity,

$$\begin{aligned}\mathfrak{R}_n\Big(\Big\{z \mapsto |a_j| \vee |a_{j+1}| 1\{u_j < \theta^\top z \le u_{j+1}\} :\ \theta \in \Theta, u_j \in \mathcal{U}\Big\}\Big) \\ \le A\mathfrak{R}_n\Big(\Big\{1\{u_j < \theta^\top z \le u_{j+1}\} :\ \theta \in \mathbb{R}^d, u_j \in \mathbb{R}\Big\}\Big).\end{aligned}$$

Using the relationship between Rademacher complexity and VC dimension [56, Examples 5.24],

$$\begin{aligned}&\mathbb{E}_\otimes\Big[\mathfrak{R}_n\Big(\Big\{z \mapsto 1\{u_j < \theta^\top z \le u_{j+1}\} :\ \theta \in \mathbb{R}^d, u_j \in \mathbb{R}\Big\}\Big)\Big] \\ &\le C\sqrt{\frac{\mathrm{VCdim}\Big(\Big\{z \mapsto 1\{u_j < \theta^\top z \le u_{j+1}\} :\ \theta \in \mathbb{R}^d, u_j \in \mathbb{R}\Big\}\Big)}{n}}.\end{aligned}$$

By [14] we have

$$\mathrm{VCdim}\Big(\Big\{z \mapsto 1\{u_j < \theta^\top z \le u_{j+1}\} :\ \theta \in \mathbb{R}^d, u_j \in \mathbb{R}\Big\} \le Cd.$$

Combining the inequalities above we conclude the result. □

### EC.3.3. Proofs for Section 4.3

*Proof of Example 7.* We derive an upper bound on $\mathcal{N}(\epsilon; \mathcal{F}, \mathsf{d}_\mathcal{F})$. We have that

$$\begin{aligned}&\Big|\|f_\theta\|_{\mathrm{Lip}} - \|f_{\tilde\theta}\|_{\mathrm{Lip}}\Big| \\ &= \Big|\|\theta\|_* \sup_{x\in\mathcal{X}} \ell'(\theta^\top x) - \|\tilde\theta\|_* \sup_{x\in\mathcal{X}} \ell'(\tilde\theta^\top x)\Big| \\ &= \Big|\|\theta\|_* \sup_{x\in\mathcal{X}} \ell'(\theta^\top x) - \|\tilde\theta\|_* \sup_{x\in\mathcal{X}} \ell'(\theta^\top x) + \|\tilde\theta\|_* \sup_{x\in\mathcal{X}} \ell'(\theta^\top x) - \|\tilde\theta\|_* \sup_{x\in\mathcal{X}} \ell'(\tilde\theta^\top x)\Big| \\ &\le L_\ell \|\tilde\theta - \theta\|_* + \|\tilde\theta\|_* \sup_{x\in\mathcal{X}} |\ell'(\theta^\top x) - \ell'(\tilde\theta^\top x)| \\ &\le L_\ell \|\tilde\theta - \theta\|_* + B\hbar_\ell \mathrm{diam}(\mathcal{X}) \|\tilde\theta - \theta\|_*.\end{aligned}$$

Hence

$$\begin{aligned}\mathsf{d}_\mathcal{F}(f_\theta, f_{\tilde\theta}) &= \sup_{x\in\mathcal{X}} \Big|\ell(\theta^\top x) - \ell(\tilde\theta^\top x)\Big| \bigvee \Big|\|f_\theta\|_{\mathrm{Lip}} - \|f_{\tilde\theta}\|_{\mathrm{Lip}}\Big| \\ &\le L_\ell \mathrm{diam}(\mathcal{X}) \|\theta - \tilde\theta\|_* \bigvee \big(L_\ell + B\hbar_\ell \mathrm{diam}(\mathcal{X})\big) \|\tilde\theta - \theta\|_* \\ &= \|\theta - \tilde\theta\|_* \Big(L_\ell \mathrm{diam}(\mathcal{X}) \vee \big(L_\ell + B\hbar_\ell \mathrm{diam}(\mathcal{X})\big)\Big).\end{aligned}$$

As a result,

$$\mathcal{N}(\epsilon; \mathcal{F}, \mathsf{d}_\mathcal{F}) \le \left(\frac{L_\ell \mathrm{diam}(\mathcal{X}) \vee (L_\ell + B\hbar_\ell \mathrm{diam}(\mathcal{X}))}{\epsilon}\right)^d.$$

□

### EC.3.4. Proofs for Section 4.4

We first state some properties of leaky ReLU networks. Using the positive homogeneity of the leaky ReLU function, for any $c > 0$, it holds that $\sigma(W_1 x) = \sigma(cW_1 x)/c$. As a result, for any $\theta = (W_1, W_2)$, there exists $c > 0$ and $(\tilde W_1, \tilde W_2)$ such that $\|\tilde W_1\|_{op} = \|\tilde W_2\|_{op} = 1$ and

$$W_2 \sigma(W_1 x) = c\tilde W_2 \sigma(\tilde W_1 x). \tag{EC.6}$$

The operator norm assumption on $W_m$'s implies $c \in [0, 1]$.

LEMMA EC.20. *Under the setting of Example 9, Assumption 4 holds.*

*Proof.* Let us verify Assumption 4. Since the softmax function $\ell$ has nonzero gradient everywhere, from the property of the compactness of $\mathcal{X}$ we have that

$$\inf_{0\leq c\leq 1, \|W_2\|_{op}=\|W_1\|_{op}=1} \mathbb{E}_{\mathbb{P}_{\text{true}}}\left[\|\nabla\ell(cW_2\sigma(W_1x),y)W_2\sigma'(W_1x)W_1\|_2^2\right]^{\frac{1}{2}} > 0,$$

which shows the lower bound in Assumption 4. Hence for all $\theta = (W_1, W_2) \in \Theta$,

$$\begin{aligned} & w_f(x,y) \\ &= \frac{\|\nabla\ell(W_2\sigma(W_1x),y)W_2\sigma'(W_1x)W_1\|_2}{\|\,\|\nabla f_\theta\|_2\,\|_{\mathbb{P}_{\text{true}},2}} \\ &\geq \inf_{\substack{0\leq c\leq 1\\ \theta\in\Theta:\ \|\tilde{W}_1\|_{op}=\|\tilde{W}_2\|_{op}=1}} \frac{\|\nabla\ell(c\tilde{W}_2\sigma(\tilde{W}_1x),y)\tilde{W}_2\sigma'(\tilde{W}_1x)\tilde{W}_1\|_2}{\|\,\|\nabla f_\theta\|_2\,\|_{\mathbb{P}_{\text{true}},2}} \\ &> 0, \end{aligned}$$

where we have used the compactness of $\Theta$ in the last inequality. Suppose $(x_0, y)$ is a non-differentiable point, and let $x_1, x_2$ be two differentiable points close to $x_0$ but belong to two different pieces. Using (EC.6) we have

$$\begin{aligned} & \limsup_{x_1,x_2\to x_0} w_f(x_1,y) - w_f(x_2,y) \\ &= \limsup_{x_1,x_2\to x_0} \frac{\|\nabla f_\theta(x_1,y)\|_2 - \|\nabla f_\theta(x_2,y)\|_2}{\|\,\|\nabla f_\theta\|_2\,\|_{\mathbb{P}_{\text{true}},2}} \\ &= \limsup_{x_1,x_2\to x_0} \frac{\|\nabla\ell(W_2\sigma(W_1x_1),y)W_2\sigma'(W_1x_1)W_1 - \nabla\ell(W_2\sigma(W_1x_1),y)W_2\sigma'(W_1x_2)W_1\|_2}{\|\,\|\nabla f_\theta\|_2\,\|_{\mathbb{P}_{\text{true}},2}} \\ &= \limsup_{x_1,x_2\to x_0} \frac{\|\nabla\ell(W_2\sigma(W_1x_0),y)W_2(\sigma'(W_1x_1) - \sigma'(W_1x_2))W_1\|_2}{\|\,\|\nabla f_\theta\|_2\,\|_{\mathbb{P}_{\text{true}},2}} \\ &\leq \sup_{\substack{0\leq c\leq 1\\ \|\tilde{W}_1\|_{op}=\|\tilde{W}_2\|_{op}=1}} \limsup_{x_1,x_2\to x_0} \frac{\|\nabla\ell(c\tilde{W}_2\sigma(\tilde{W}_1x_1),y)\tilde{W}_2(\sigma'(\tilde{W}_1x_1) - \sigma'(\tilde{W}_1x_2))\tilde{W}_1\|_2}{\|\,\|\nabla f_\theta\|_2\,\|_{\mathbb{P}_{\text{true}},2}}. \end{aligned}$$

thus we have shown that $w_{f_\theta}$ has bounded jump. Moreover, since $f_\theta$ is Lipschitz,

$$\limsup_{\mathsf{d}(z,\mathcal{D}_{f_\theta})\to\infty} \frac{w_{f_\theta}(z)}{\mathsf{d}(z,\mathcal{D}_{f_\theta})} = 0.$$

Thereby we have verified the upper bound in Assumption 4. ◻

Lemma EC.21. *Under the setting in Example 9,*

$$\mathbb{E}_\otimes[\mathfrak{R}_n(\mathcal{I}_\rho)] \leq Cd_1\sqrt{\frac{d}{n}}.$$

*Proof.* We have

$$\mathsf{d}(z,\mathcal{D}_{f_\theta}) = \min_{1\leq j\leq d_1} \frac{|W_{1,j}x|}{\|W_{1,j}\|_2}$$

Similar to the reasoning of Lemma EC.19, we have

$$\begin{aligned}\mathbb{E}_{\otimes}[\mathfrak{R}_n(\mathcal{I}_\rho)] &\le \mathfrak{R}_n\Big(\Big\{x \mapsto 1\big\{\min_{1\le j\le d_1} |W_{1,j}x| \le \rho\|W_{1,j}\|_2\big\} : W_1 \in \mathbb{R}^{d_1\times d}\Big\}\Big)\\ &= \mathbb{E}_{\otimes}\Big[\mathfrak{R}_n\Big(\Big\{x \mapsto \max_{1\le j\le d_1} 1\big\{|W_{1,j}x| \le \rho\|W_{1,j}\|_2\big\} : W_1 \in \mathbb{R}^{d_1\times d}\Big\}\Big)\Big]\\ &\le d_1\mathbb{E}_{\otimes}\Big[\mathfrak{R}_n\Big(\Big\{x \mapsto 1\big\{|w^\top x| \le \rho\|w\|_2\big\} : w \in \mathbb{R}^{d}\Big\}\Big)\Big]\\ &\le Cd_1\sqrt{\frac{d}{n}}.\end{aligned}$$

□

### EC.3.5. Proofs for Section 4.5

*Proof of Example 11.* Using the strong duality result from [26, Section 4.2], the dual problem equals

$$\min_{\lambda\ge 0, f}\Big\{\int_{\Xi} f(\xi)d\xi + \lambda\rho_n^2 - \frac{1}{n}\sum_{i=1}^{n}\sum_{m=1}^{M_i}\sup_{\xi\in\Xi}\Big\{-\log f(\xi) - \lambda\|\xi-\xi_{i,m}\|^2\Big\}\Big\},$$

which is of the same form of the dual (D). It follows that the gradient norm penalty in (V) equals

$$\Big(\frac{1}{n}\sum_{i=1}^{n}\sum_{m=1}^{M_i}\|\nabla_\xi \log f(\xi_{i,m})\|_2^2\Big)^{\frac{1}{2}}.$$

Since we have assumed $\log f$ has Lipschitz gradient bounded by $\hbar$, invoking Lemma 1 (which has been proved for the general metric space under the condition (**GS**)) yields that $|\mathcal{L}_n^{\rm rob}(\rho_n; f) - \mathcal{L}_n^{\rm reg}(\rho; f)| = O(1/n)$. □

## Appendix EC.4: Proofs for Section 5

*Proof of Theorem 3.* Let us denote $\ell_f(z) = \ell(f(x), y)$. Note from the Lipschitzness of $L_\ell$, $\ell(f(\tilde{x}), y) - \ell(f(x), y) \le L_\ell(f(\tilde{x}) - f(x))$, we have

$$\sup_{\tilde{x}\in\mathcal{X}}\{\ell(f(\tilde{x}), y) - \ell(f(x), y) :\ \|\tilde{x}-x\| \le \rho\} \le L_\ell \sup_{\tilde{x}\in\mathcal{X}}\{f(\tilde{x}) - f(x) :\ \|\tilde{x}-x\| \le \rho\},$$

thus for any distribution $\mathbb{Q}$, it holds that

$$\mathcal{R}_{Q,\infty}(\rho; \ell_f) \le L_\ell \mathcal{R}_{\mathbb{Q},\infty}(\rho; f).$$

Hence, we have the decomposition

$$\begin{aligned}\mathcal{A}(\rho; f) - \mathcal{A}_n(\rho; f) &= (\mathbb{E}_{\mathbb{P}_{\rm true}}[\ell_f] - \mathbb{E}_{\mathbb{P}_n}[\ell_f]) + \Big(\mathcal{R}_{\mathbb{P}_{\rm true},\infty}(\rho; \ell_f) - \mathcal{R}_{\mathbb{P}_n,\infty}(\rho; \ell_f)\Big)\\ &\le (\mathbb{E}_{\mathbb{P}_{\rm true}}[\ell_f] - \mathbb{E}_{\mathbb{P}_n}[\ell_f]) + L_\ell\Big(\mathcal{R}_{\mathbb{P}_{\rm true},\infty}(\rho; f) - \mathcal{R}_{\mathbb{P}_n,\infty}(\rho; f)\Big)\\ &\le (\mathbb{E}_{\mathbb{P}_{\rm true}}[\ell_f] - \mathbb{E}_{\mathbb{P}_n}[\ell_f]) + L_\ell\rho\Big(\|\,|\partial f|\,\|_{\mathbb{P}_{\rm true},1} - \|\,|\partial f|\,\|_{\mathbb{P}_n,1}\Big)\\ &\quad + L_\ell\Big(\mathcal{R}_{\mathbb{P}_{\rm true},\infty}(\rho; f) - \rho\|\,|\partial f|\,\|_{\mathbb{P}_{\rm true},1}\Big) + L_\ell\Big(\rho\|\,|\partial f|\,\|_{\mathbb{P}_n,1} - \mathcal{R}_{\mathbb{P}_n,\infty}(\rho; f)\Big).\end{aligned}$$

By Lemmas EC.5 and EC.6, with probability at least $1 - e^{-t}$, for every $f \in \mathcal{F}$,

$$\mathbb{E}_{\mathbb{P}_{\rm true}}[\ell_f] - \mathbb{E}_{\mathbb{P}_n}[\ell_f] \le 2L_\ell\mathbb{E}_{\otimes}[\mathfrak{R}_n(\mathcal{F})] + 2\sqrt{\frac{t}{2n}},$$

and with probability at least $1-e^{-t}$, for every $f \in \mathcal{F}$,

$$\| |\partial f| \|_{\mathbb{P}_{\text{true}},1} - \| |\partial f| \|_{\mathbb{P}_n,1} \le 2\mathbb{E}_{\otimes}[\mathfrak{R}_n(|\partial\mathcal{F}|)] + L\sqrt{\frac{t}{2n}}.$$

Using an argument similar to [5, Remark 9], there exists $C_1 > 0$ such that

$$\mathcal{R}_{\mathbb{P}_{\text{true}},\infty}(\rho; f) - \rho\| |\partial f| \|_{\mathbb{P}_{\text{true}},1} \le C_1\rho^2.$$

By Theorem 1(I)(III), there exists $\bar{\rho}, C > 0$ such that for all $\rho < \bar{\rho}$, with probability at least $1-e^{-t}$, for every $f \in \mathcal{F}$,

$$\rho\| |\partial f| \|_{\mathbb{P}_n,1} - \mathcal{R}_{\mathbb{P}_n,\infty}(\rho; f) \le C_2\rho^2 + 2\rho\mathbb{E}_{\otimes}[\mathfrak{R}_n(\mathcal{J}_\rho)] + \rho\sqrt{\frac{t}{2n}}.$$

Consequently, with probability at least $1-3e^{-t}$, for every $f \in \mathcal{F}$,

$$\mathcal{A}(\rho; f) - \mathcal{A}_n(\rho; f) \le 2L_\ell(\mathbb{E}_{\otimes}[\mathfrak{R}_n(\mathcal{F})] + \rho\mathbb{E}_{\otimes}[\mathfrak{R}_n(|\partial\mathcal{F}|)] + \rho\mathbb{E}_{\otimes}[\mathfrak{R}_n(\mathcal{J}_\rho)]) + (2 + (L+1)L_\ell\rho)\sqrt{\frac{t}{2n}} + L_\ell C\rho^2.$$

In particular, when $\mathcal{F} = \{x \mapsto \theta^\top x : \theta \in \Theta \subset \mathbb{R}^d\}$, the above bound reduces to

$$\mathcal{A}(\rho; f) - \mathcal{A}_n(\rho; f) \le 2L_\ell(\mathbb{E}_{\otimes}[\mathfrak{R}_n(\mathcal{F})] + \rho\mathbb{E}_{\otimes}[\mathfrak{R}_n(|\partial\mathcal{F}|)]) + (2 + LL_\ell\rho)\sqrt{\frac{t}{2n}},$$

and

$$\mathbb{E}_{\otimes}[\mathfrak{R}_n(|\partial\mathcal{F}|)]) = \mathbb{E}_{\otimes}[\mathfrak{R}_n(\{\|\theta\| : \theta \in \Theta\})].$$

□